\documentclass[]{article}
\usepackage{authblk}
\usepackage{fullpage}

\usepackage{multirow}
\usepackage{caption}
\usepackage{Definitions}
\usepackage[english]{babel}
\usepackage[latin1]{inputenc}
\usepackage{amsfonts}
\usepackage{amsmath}
\usepackage{latexsym}
\usepackage{hyperref}
\usepackage{mathrsfs}
\usepackage{etoolbox}
\usepackage{enumitem}
\usepackage{soul}
\usepackage{stackengine}
\usepackage{graphicx}
  \usepackage{subfig}
  \usepackage[noend]{algorithmic}
\usepackage[vlined,ruled]{algorithm2e}
\PassOptionsToPackage{numbers}{natbib}
\newcommand{\enc}{\phi^\text{enc}}

\newcommand{\nn}{\nonumber}

\newcommand{\ngreen}[1]{{\color{green} #1}}

\newcommand{\specialcell}[2][c]{%
  \begin{tabular}[#1]{@{}l@{}}#2\end{tabular}}

    \newcommand{\ba}{Barab{\'a}si-Albert}
\newcommand{\UnnumberedFootnote}[1]{{\def\thefootnote{}\footnote{#1}
\addtocounter{footnote}{-1}}}

 \graphicspath{{./FIG/}}
 
 \newcommand{\xhdr}[1]{\vspace{1mm}\noindent{{\bf #1.}}}

\renewcommand{\tilde}{\widetilde}

    \newcommand{\ourmodel}{\textsc{NeVAE}}

	\title{\textsc{NeVAE}: A Deep Generative Model for Molecular Graphs$^*$}

\date{}
\author[1]{Bidisha Samanta}
\author[2]{Abir De}
\author[1]{Gourhari Jana}
\author[3]{Vicen\c{c} Gomez}
\author[1]{\\ Pratim Kumar Chattaraj}
\author[1]{Niloy Ganguly}
\author[2]{Manuel Gomez-Rodriguez}

\affil[1]{IIT Kharagpur,  bidisha@iitkgp.ac.in,  gour2015hari@iitkgp.ac.in, pkc@chem.iitkgp.ernet.in, niloy@cse.iitkgp.ernet.in}
\affil[2]{Max Planck Institute for Software Systems, ade@mpi-sws.org, manuelgr@mpi-sws.org}
\affil[3]{DTIC, Universitat Pompeu Fabra, Barcelona, Spain}
\begin{document} 
\maketitle

\UnnumberedFootnote{$^{*}$Preliminary version of this work appeared in Samanta et al. \cite{aaaiBS}.}

\begin{abstract}
Deep generative models have been praised for their abi\-li\-ty to learn smooth latent representation of images, text, and 
audio, which can then be used to generate new, plausible data.
However, current generative models are unable to work with molecular graphs due to their unique characteristics---their 
underlying structure is not Euclidean or grid-like, they remain isomorphic under permutation of the nodes labels, and they 
come with a different number of nodes and edges. 
In this paper, we first propose a novel variational autoencoder for molecular graphs, whose encoder and decoder are specially 
designed to account for the above properties by means of several technical innovations.
Moreover, in contrast with the state of the art, our decoder is able to provide the spatial coordinates of the atoms of the molecules 
it generates.
Then, we develop a gradient-based algorithm to optimize the decoder of our model so that it learns to generate molecules that 
maximize the value of certain property of interest and, given a mo\-le\-cule of interest, it is able to optimize the spatial configuration 
of its atoms for greater stability.
Experiments reveal that our variational autoencoder can discover plausible, diverse and novel molecules more effectively than 
several state of the art models. Moreover, for several properties of interest, our optimized decoder is able to identify molecules 
with property values $121$\% higher than those identified by several state of the art methods based on Bayesian optimization 
and reinforcement learning.

% our gradient-based algorithm is able to find optimized decoder
%
% \ad{Moreover, our optimized decoder achieves a superior performance at property oriented molecule discovery than several state-of-the-art methods
% including those which
% typically rely on Bayesian optimization over the continuous latent representation of molecules, by a significantly large margin ($\sim 121\%$).}

\end{abstract}

% \begin{keywords}
% Drug design, Molecule discovery, Deep generative models, Variational autoencoders, Geometric deep learning.
% \end{keywords}

\section{Introduction}
\label{sec:introduction}
Drug design aims to identify (new) molecules with a set of specified properties, which in turn results in a therapeutic benefit to a group of patients. 
However, drug design is still a lengthy, expensive, difficult, and inefficient process with a low rate of new therapeutic discovery~\cite{paul2010improve}, 
in which candidate molecules are produced through chemical synthesis or biological processes.
In the context of computer-aided drug design~\cite{merz2010drug}, there is a great interest in developing automated, machine learning techniques 
to discover sizeable numbers of plausible, diverse, and novel candidate molecules with various desirable properties in the vast ($10^{23}-10^{60}$) 
and unstructured molecular space~\cite{polishchuk2013estimation}. 

In recent years, there has been a flurry of work devoted to developing deep generative models for automatic molecule design~\cite{dai2018syntax,decao2018molgan,gomez2016automatic,organ,jin2018junction,kusner2017grammar,graphvae, gcpn}, which has predominantly 
followed two strategies.
The first strategy~\cite{dai2018syntax,gomez2016automatic,organ,kusner2017grammar} consists of representing molecules using a 
domain specific textual representation---SMILES strings---and then leveraging deep generative models for text generation for molecule design.
Unfortunately, SMILE strings do not capture the structural similarity between molecules and, moreover, a molecule can have multiple SMILES 
representations. As a consequence, the generated molecules lack in terms of diversity and validity, as shown in Tables~\ref{tab:unique}--\ref{tab:validity} and Figure~\ref{fig:BO_score1}.
The second strategy~\cite{decao2018molgan,jin2018junction,graphvae,gcpn} consists of representing molecules using molecular graphs, rather than SMILES representations, 
and then developing deep generative models for molecular graphs, in which atoms correspond to nodes and bonds correspond to edges. 
However, current generative models for molecular graphs share one or more of the following limitations, which preclude them from realizing all their 
potential:
\vspace{1mm}
\begin{itemize}[noitemsep,nolistsep,leftmargin=0.9cm]
\item[I.] They can only generate (and be trained on) molecules with the same number of atoms while, in practice, molecules having similar properties 
often come with a different number of atoms and bonds.
\item[II.] They are not invariant to permutations of their node labels, however, molecular graphs remain isomorphic under permutation of their node labels.
\item[III.] Their training procedure suffers from a quadratic complexity with respect to the number of nodes in the graph, which makes it difficult to leverage 
a sizeable number of large molecules during training.
\item[IV.] They generate molecular graphs by combining a small set of molecular \emph{graphlets} (or subgraphs), which constrain the diversity of the 
generated molecules, as shown in Table~\ref{tab:unique} and Figure~\ref{fig:BO_score1}.
\item[V.] They do not provide the spatial coordinates of the atoms they generate, whereas in practice, a molecule is a three-dimensional object in which the coordinates 
of its atoms significantly influence its chemical properties, as shown in Figure~\ref{fig:bestMollogP}.
\item[VI.] To identify molecules that maximize the value of certain property (\eg, solubility in water), they resort to either traditional Bayesian optimization or reinforcement
learning over the continuous latent representation of molecules they find. However, such procedures are unable to discover a sizeable set of candidate molecules with high 
property values, as shown in Table~\ref{tab:propval} and Figure~\ref{fig:BO_score1}. %\vspace{1mm}
\end{itemize}
To address the first five shortcomings (I-V), we develop \ourmodel, a deep generative model for molecular graphs based on variational autoencoders. 
Our model relies on several technical innovations, which distinguish us from previous work:
\vspace{1mm}
\begin{itemize}[noitemsep,nolistsep,leftmargin=0.9cm]
\item[(i)] Our probabilistic encoder learns to aggregate information (\eg, bond features, atoms and their coordinates) from a different number of hops away 
from a given atom and then map this aggregate information into a continuous latent space, as in inductive graph representation 
learning~\cite{hamilton2017inductive,lei2017}.
However, in contrast with inductive graph representation learning, the aggregator functions are learned via variational inference so that the 
resulting aggregator functions are especially well suited to enable the probabilistic decoder to generate new molecules rather than other machine 
learning tasks such as, \eg, link prediction.
% manuel: fixed
% VICEN: I think stating "supervised learning tasks" instead of "machine learning tasks" gives more generality to our method
%machine learning tasks such as, \eg, link prediction. 
%
Moreover, by using (symmetric) aggregator functions, it is invariant to permutations of the node labels and can encode graphs with a variable number of atoms, as 
opposed to existing graph generative models, with a few the notable exception of those based on GCNs~\cite{welling}.

\item[(ii)] {Our probabilistic decoder} jointly represents all edges as an unnormalized log probability vector (or `logit'), which then feeds a single multinomial edge distribution. 
Such scheme allows for an efficient inference algorithm with $O(l)$ complexity, where $l$ is the number of true edges in the molecules, which is also invariant to permutations of the node labels.
In contrast, previous work typically models the presence and absence of each potential edge using a Bernoulli distribution and this leads to inference algorithms with $O(n^2)$ complexity, where $n$ is the number of nodes, which are not permutation invariant. 

\item[(iii)] Our probabilistic decoder is able to guarantee a set of local structural and functional properties in the generated molecules by using a \emph{mask} in the
edge distribution definition, which can prevent the generation of certain \emph{undesirable} edges during the decoding process. 
While masking have been increasingly used to account for prior (expert) knowledge in generative models~\cite{gomez2016automatic,kusner2017grammar} based on SMILES, their use in generative models for molecular graphs has been lacking.  

\item[(iv)] Our probabilistic decoder is able to provide the spatial coordinates of the atoms of the molecules it generates. To do so, it models the position of each atom using a 
Gaussian distribution whose mean and variance depend on its latent representation as well as that of each of its neighbors. \vspace{1mm}
\end{itemize}

To address the last shortcoming (VI), we develop a gradient-based algorithm to optimize the decoder of our model for property oriented molecule generation, \ie, to optimize the decoder 
so that it learns to generate molecules that maximize the value of certain property (\eg, solubility in water). 
Note that, in contrast with recent reinforcement learning methods for property oriented molecule generation~\cite{decao2018molgan, organ, gcpn}, our gradient-based algorithm benefits
from the inductive bias provided by the original decoder, which in turns enable us to identify \emph{better} molecules, as shown in Table~\ref{tab:propval}.
Moreover, given a mo\-le\-cule of interest, our gradient-based algorithm can also be used to optimize the spatial configuration of its atoms for greater stability.
%
%We believe our algorithm is of independent interest since it may be adapted to optimize other types of deep generative models so that they learn to generate data (\eg, graphs, images, text or audio) that maximizes certain property value.
%
We believe our algorithm is of independent interest since it may be adapted to other deep generative models designed for other data types such as graphs, images, text or audio, 
that maximizes certain property value.

We experiment with molecules from two publicly available datasets, ZINC~\cite{irwin2012zinc} and QM9~\cite{ramakrishnan2014quantum}.
First, we show that \ourmodel{} beats the state of the art in terms of several relevant quality metrics, \ie, validity, novelty and uniqueness,
and the resulting latent space representation of molecules exhibits powerful semantics---we can smoothly interpolate 
between molecules---and generalization ability---we can generate (valid) molecules that are larger than any of the molecules in the datasets.
Then, we demonstrate that, for several properties of interest (\eg, solubility in water), our gradient-based algorithm is able to successfully optimize \ourmodel{}'s decoder for property oriented molecule generation. In particular, the optimized decoder is able to identify molecules with property values $121$\% higher than those identified
by several state of the art methods based on Bayesian optimization and reinforcement learning and, given a mo\-le\-cule of interest, it is able to optimize the spatial configuration of its atoms for greater stability, \ie, lower potential energy. 
To facilitate research in this area, we are releasing an open source implementation of our model in Tensorflow as well as synthetic and real-world data used in our 
experiments\footnote{\scriptsize \url{https://github.com/Networks-Learning/nevae}}.

\section{Background on Variational Autoencoders}
\label{sec:background}
Variational autoencoders~\cite{vaebasic1,vaebasic2} are characterized by a probabilistic ge\-ne\-ra\-tive model $p_\theta(\xb|\zb)$ of the observed variables $\xb \in \RR^{N}$ given the latent 
variables $\zb \in \RR^{M}$, a prior distribution over the latent variables $p(\zb)$ and an approximate probabilistic inference model $q_{\phi}(\zb | \xb)$. 
In this characterization, $p_{\theta}$ and $q_{\phi}$ are arbitrary distributions parametrized by two (deep) neural networks $\theta$ and $\phi$ and one can think of the 
generative model as a probabilistic \emph{decoder}, which \emph{decodes} latent variables into observed variables, and the inference model as a probabilistic \emph{encoder}, 
which \emph{encodes} observed variables into latent variables.

Ideally, if we use the maximum likelihood principle to train a variational autoencoder, we should optimize the marginal log-likelihood of the observed data, \ie, $\EE_{\Dcal} \left[\log p_\theta(\xb)\right]$, 
where $p_\Dcal$ is the data distribution. Unfortunately, computing $\log p_{\theta}(\xb)$ requires marginalization with respect to the latent variable $\zb$, which is typically intractable.
Therefore, one resorts to maximizing a variational lower bound or evidence lower bound (ELBO) of the log-likelihood of the observed data, \ie,

\vspace*{-0.5cm}
 \begin{align*}
  \max_{\theta} \max_{\phi} \EE_{\Dcal} \left[ -\text{KL}(q_\phi(\zb|\xb)||p(\zb))+\EE_{q_\phi(\zb|\xb)}{\log p_\theta (\xb|\zb)} \right].\nn 
 \vspace*{-1cm}
 \end{align*}
{Finally, note that} the quality of this variational lower bound depends on the expressive ability of the approximate inference model $q_{\phi}(\zb|\xb)$, which is typically 
assumed to be a normal distribution whose mean and variance are parametrized by a neural network $\phi$ with the observed data $\xb$ as input. %an input. 

\section{{\ourmodel:~A~Variational~Autoencoder for~Molecular~Graphs}}
\label{sec:model}
In this section, we first give a high-level overview of the design of \ourmodel, our variational autoencoder for molecular graphs, starting from the data 
it is designed for. 
Then, we describe more in-depth the key technical aspects of its individual components. 
Finally, we elaborate on the training procedure, scalability and implementation details.
 
\xhdr{High-level overview}
We observe a collection of $N$ molecular graphs $\{ \Gcal_i = (\Vcal_i, \Ecal_i) \}_{i \in [N]}$, where $\Vcal_i$ and $\Ecal_i$ denote the corresponding set of nodes (atoms) and 
edges (bonds), respectively, and this collection may contain graphs with a different number of nodes and edges.
Moreover, for each molecular graph $\Gcal = (\Vcal, \Ecal)$, we also observe a set of node features $\Fcal = \{ \tb_u, \xb_u \}_{u \in \Vcal}$ % , a set of node positions $\Xcal=\{\xb _u \}_{ u \in \Vcal }$
and edge weights $\Ycal = \{ y_{uv} \}_{(u,v) \in \Ecal}$.
% \footnote{For
% ease of exposition, we assume the node features and edge weights take discrete values, however, our model could be augmented to continuous values.} 
%
More specifically, $\tb_u$ are one-hot representations of the type of the atoms (\ie, $C$, $H$, $N$ or $O$), $\xb_u$ are the coordinates of the atoms in three dimensional space,
and $y_{uv}$ are the bond types (\ie, single, double, triple).
Our goal is then to design a variational autoencoder for molecular graphs that, once trained on this collection of graphs, has the ability of creating new plausible molecular 
graphs, including node features and edge weights. 
In doing so, it will also provide a latent representation of any graph in the collection (or elsewhere) with meaningful semantics.
\begin{figure*}[t]
	\centering
	 \includegraphics[width=0.98\textwidth]{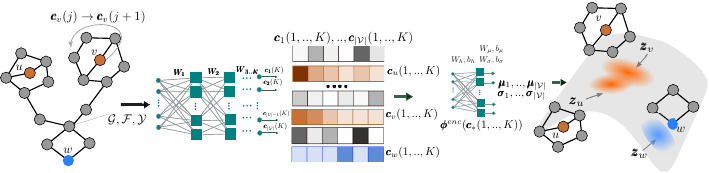}\hspace*{0.1 cm}
	\caption{The encoder of our variational autoencoder for molecular graphs. From left to right, given a molecular graph $\Gcal$ with a set of node features $\Fcal$ and edge 
	weights $\Ycal$, the encoder aggregates information from a different number of hops $j \leq K$ away for each node $v \in \Gcal$ into an embedding vector $\cb_v(j)$. 
% 	To do so, it uses a feedforward network to propagate information between different search depths, which is parametrized by a set of weight matrices 
% 	$\Wb_j$. 
	These embeddings are fed into a differentiable function $\phi^{enc}$ which parameterizes the posterior distribution $q_{\phi}$, from where the latent representation of each node 
	in the input graph are sampled from.}
	\label{fig:encoder}
\end{figure*}

% Following the above background on variational autoencoders, we characterize our variational autoencoder by means of:
Following the above background on variational autoencoders, we characterize \ourmodel\ by means of:
%
% \begin{align*}
% &\text{--- \emph{Prior}:} & p(\zb_1, \ldots, \zb_n), \text{ where } |\Vcal| = |\Fcal| =  n \sim \text{Poisson}(\lambda_n) \\
% &\text{--- \emph{Inference model (encoder)}:} & q_{\phi}(\zb_1, \ldots, \zb_n | \Vcal, \Ecal, \Fcal, \Ycal) \\
% &\text{--- \emph{Generative model (decoder)}:} & p_{\theta}(\Ecal, \Fcal, \Ycal | \zb_1, \ldots, \zb_n), \text{ where } |\Ecal| = |\Ycal| = l \sim \text{Poisson}(\lambda_l)
% \end{align*}
%
% \vspace*{-0.1cm}
\begin{align*}
&\text{--- \emph{Prior}:}   \ \ p_{z}(\zb_1, \ldots, \zb_n), \text{ where } |\Vcal| = |\Fcal| =  n \sim \text{Poisson}(\lambda_n) \\
&\text{--- \emph{Inference model (encoder)}:} \ \   q_{\phi}(\zb_1, \ldots, \zb_n | \Vcal, \Ecal, \Fcal, \Ycal) \\
&\text{--- \emph{Generative model (decoder)}:}\ \  p_{\theta}(\Ecal, \Fcal, \Ycal | \zb_1, \ldots, \zb_n)
\vspace*{-0.2cm}
% & \ \ \ \ \text{ where } l= |\Ecal| = |\Ycal|
\end{align*}
In the above characterization, note that 
we define one latent variable per node, \ie, we have a \emph{node-based} latent representation, and
%
% (ii) the decoder also generates the number of edges; and
%
the number of nodes is a random variable. As a consequence, both the latent representation as well as the graph
can vary in size.
%
% \ad{(iii) the number of nodes has a Poisson distribution with constant parameter $\lambda_n$; however, the number of edges depends
% on the latent representations of the nodes by means of $\lambda_{l,\psi}$, where $\psi$ is a neural network.}
%
% (iii) the generative model does not characterize the node features since our focus is on generating the edges and edge weights of the graph, however,
% it could be easily augmented to also model node features. 
%
Next, we formally define the functional form of the inference model, the generative model, and the prior.

\xhdr{Inference model (probabilistic encoder)}
% \xhdr{Inference model (encoder)}
%
Given a graph $\Gcal = (\Vcal, \Ecal)$ with node features $\Fcal$ and edge weights 
$\Ycal$,
our inference model $q_{\phi}$ defines a probabilistic encoding for each node in the graph by aggregating information
from different distances. More formally, for each node $u$, the inference model is defined as follows:
\begin{equation} \label{eq:prob-encoder}
q_{\phi}(\zb_u | \Vcal, \Ecal, \Fcal, \Ycal) \sim \Ncal(\mub_u, \diag(\sigmab_u))
\end{equation}
where $\zb_u$ is the latent variable associated to node $u$, $\left[\mub_u, \diag(\sigmab_u)\right] = \phi^{enc} \left( \cb_u(1), \ldots, \cb_u(K) \right)$, and 
$\cb_u(k)$ aggregates information from $k$ hops away from $u$, \ie,
 \begin{align} \label{eq:embedding}
\hspace{-0.2cm} \cb_u(k) \hspace{-0.1cm} = \hspace{-0.1cm}
\begin{cases}
\rb( \Wb^{\Tcal} _k \tb_u + \Wb^{\Xcal} _k  \xb_u) & \hspace{-0.3cm}\text{if} \ k = 1 \\
\rb \left((\Wb^{\Tcal} _k \tb_u +  \Wb^{\Xcal} _k  \xb_u ) \odot \Lambdab\big(  \cup_{{v\in\Ncal(u)} }  y_{uv}\, \gb(\cb_v(k-1)\big)\right) \ & \hspace{-0.3cm}\text{if} \ k > 1.
\end{cases}
\end{align}
In the above, $\Wb^{\Tcal} _k$ and $\Wb^{\Xcal} _k$ are trainable weight matrices, which propagate information between different search depths, $\Lambdab(.)$ is a (possibly nonlinear) symmetric aggregator function 
in its arguments, $\gb(\cdot)$ and $\rb(\cdot)$ are (possibly nonlinear) differentiable functions, $\phi^{enc}$ is a neural network, and $\odot$ denotes pairwise product. Figure~\ref{fig:encoder} 
describes our encoder architecture.

The above node embeddings, defined by Eq.~\ref{eq:embedding}, are very similar to the ones used in several graph representation learning algorithms such as 
GraphSAGE~\cite{hamilton2017inductive}, column networks~\cite{pham2017column}, and GCNs~\cite{kipf2016semi}, the main difference with our work is the way 
we will train the weight matrices $\Wb^{\bullet} _k$. 
Here, we will use variational inference so that the resulting embeddings are especially well suited to enable our probabilistic decoder to generate new, plausible molecular 
graphs. In contrast, the above algorithms use non variational approaches to compute general purpose embeddings to feed downstream machine learning tasks.
% \niloy{Which result support this claim}
% \manuel{Using variational inference means that the current embeddings maximize the likelihood of the decoded graphs. In that sense, they are specially well suited.
% we could try to encode graphs using for example GraphSAGE, and then given those fixed encodings, try to fit only the decoder and repeat all the experiments in synthetic
% and real data and incorporate that as a baseline, which would perform worse. That seems quite tedious though.}

The follo\-wing proposition highlights several desirable theoretical properties of our probabilistic encoder, % ~\ref{sec:proof-prop-1}), 
which distinguishes our design from most existing generative models of graphs~\cite{jin2018junction,graphvae}:
%
% \niloy{But is this unique to us - I dont think so}
% \manuel{inductive representation learning sometimes satisfy the properties below, however, generative models of graphs do not except the ones based on GCNs, that is why i added 
% a sentence above to clarify that. Also now in the intro we clarify that}
%
\begin{proposition}\label{prop-1}
The probabilistic encoder defined by Eqs.~\ref{eq:prob-encoder} and~\ref{eq:embedding} has the following properties:
\begin{itemize}[noitemsep,nolistsep]
\item[(i)] For each node $u$, its corresponding embedding $\cb_u(k)$ is invariant to permutations of the node labels of its neighbors.
% \niloy{But this is true for others}
% \manuel{Same comment as above, now clarify here and in the intro. This is true for node embeddings/representation learning, not for most generative models of graphs}
%
\item[(ii)] The weight matrices $\Wb^{\Tcal} _1, \ldots, \Wb^{\Tcal} _K$ and $\Wb^{\Xcal} _1, \ldots, \Wb^{\Xcal} _K$  do not depend on the number of nodes and edges in the graph and thus a single 
encoder allows for graphs with a variable number of nodes and edges. 
\end{itemize}
%
% \niloy{It seems one of the important contributions is independence of number of nodes, it is not clear why is this difficult and not done before and what innovation you have done to this end. Also I dont see any 
% experiment supporting this.
% For example, if I train with clustering coefficient }
% \manuel{I have clarified this when I talk about the aggregator functions in the introduction. Re. the difficulty, I don't think there is a single answer.}
%
\end{proposition}
\begin{proof}
\noindent First, we prove property (i). 
Assume the embedding $\cb_v(k-1)\in \RR^{D\times 1}$ for all $k> 1$ and $v\in \Vcal$. Moreover,
note that, in Eq.~\ref{eq:embedding}, all the functions $r(\cdot)$, $g(\cdot)$ and $\Lambdab(\cdot)$ are defined term-wise.

Consider $\pi$, a permutation of the node labels, \ie\ for each $u$, we have $\pi(u)\in\Vcal$; and the set of all shuffled 
labels $\{\pi(w)|w\in\Vcal\}=\Vcal$. Let us denote $\util:=\pi(u)$. Now we need to prove
\begin{align}\label{ToProof}
\cb_u(k)=\cb_{\util}(k) \ \forall k\ge 1,\  \forall u\in\Vcal 
\end{align}
We proof this by induction. Since the features $\tb_u$ and $\xb_u$ are independent of the node label of $u$, we have
that $\tb_u=\tb_{\util}$ and $\xb_u =  \xb_{\util}$, which proves Eq.~\ref{ToProof} for $k=1$,  $\forall u\in \Vcal$.
Now assume that Eq.~\ref{ToProof} is true for $k\le k'-1$, with $k'>1$. That is,  we have,
\begin{align}\label{tf2}
\cb_{\vtil}(k'-1)=\cb_v(k'-1)\ \forall v\in\Vcal
\end{align}
Also, since the edge-weight $y_{uv}$ between nodes does not depend on their labels, we have 
\begin{align} \label{yuv}
y_{uv}=y_{\util,\vtil}.
\end{align}
This, along with Eq.~\ref{tf2} gives $\{\cup_{v\in\Ncal(u)} y_{uv} \, \gb(\cb_v(k'-1) \}=\{\cup_{\vtil\in\Ncal(\util)} y_{\util \vtil} \, \gb(\cb_{\vtil}(k'-1)\}$  which, due to the symmetric property of $\Lambdab(.)$, implies 
\begin{align}
 \Lambdab\big(\cup_{v\in\Ncal(u)} y_{uv} \,& \gb(\cb_v(k'-1)\big) 
 =\Lambdab\big(\cup_{\vtil\in\Ncal(\util)} y_{\util \vtil} \, \gb(\cb_{\vtil}(k'-1)\big)
\end{align}
The above equation, together with the fact that $\tb_u=\tb_{\util}$ and $\xb_u=\xb_{\util}$ proves Eq.~\ref{ToProof} for $k=k'$.

Second, we prove property $(ii)$. Assume the embedding $\cb_v(k-1)\in \RR^{D\times 1}$ for all $k> 1$ and $v\in \Vcal$. Moreover,
note that, in Eq.~\ref{eq:embedding}, all the functions $r(\cdot)$, $g(\cdot)$ and $\Lambdab(\cdot)$ are defined term-wise.
To ensure that $\tb_u \odot \Lambdab\left(\cup_{v\in\Ncal(u)} y_{uv} \, \gb\left(\cb_v(k-1)\right)\right)$ and $\xb_u \odot \Lambdab\left(\cup_{v\in\Ncal(u)} y_{uv} \, \gb\left(\cb_v(k-1)\right)\right)$ are well defined, we need to have $\cb_v(k-1)\in \RR^{D\times 1}$ for all $k> 1$ and $v\in \Vcal$.
Then, by matching the dimension of vectors in both sides of Eq.~\ref{eq:embedding}, we have that $\Wb^{\Tcal}_k\in \RR^{D\times D}$ and $\Wb^{\Xcal}_k\in \RR^{D\times D}$.
% Manuel: D is just the dimension of the embedding, which is something to be chosen by design. I have define it at the beginning of this paragraph.
% VICEN: where is D defined??

%Second, we prove property (ii). In Eq.~\ref{eq:embedding}, all the functions $r(.)$, $g(.)$ and $\Lambdab(.)$ are defined term-wise.
%Note that, to make sure that $\tb_u \odot \Lambdab\big(\cup_{v\in\Ncal(u)} y_{uv} \, \gb(\cb_v(k-1)\big)$ and $\xb_u \odot \Lambdab\big(\cup_{v\in\Ncal(u)} y_{uv} \, \gb(\cb_v(k-1)\big)$ are well defined, 
%we need to have $\cb_v(k-1)\in \RR^{D\times 1}$ for all $k> 1$ and $v\in \Vcal$.
%Then, by matching the dimension of vectors in both sides of Eq.~\ref{eq:embedding}, we have that $\Wb^{\Tcal}_k\in \RR^{D\times D}$ and $\Wb^{\Xcal}_k\in \RR^{D\times D}$.
%
\end{proof}

\xhdr{Generative model (probabilistic decoder)}
Given a set of of $n$ nodes with latent va\-ria\-bles $\Zcal = \{ \zb_u \}_{u \in [n]}$, our generative model $p _\theta$ is defined as follows:
\begin{equation}
p _{\theta}(\Ecal, \Ycal,\Fcal | \Zcal) = p_{\theta}(\Fcal | \Ecal,\Ycal, \Zcal) \, p_{\theta}(\Ecal,\Ycal| \Zcal) \label{eq:prob-decoder}
\end{equation}
with
\begin{align*}
p_{\theta}(\Fcal | \Ecal,\Ycal, \Zcal) &=  \prod_{u \in \Vcal} p  _{\theta} (\tb_u | \Zcal) \, p_{\theta}(\xb_u | \Ecal, \Ycal, \Zcal), \\
p_{\theta}(\Ecal,\Ycal| \Zcal) &=  p_{\theta} (l \big|\Zcal) \, \prod_{k \in [l]} p_{\theta}(e_k | \Ecal_{k-1}, \Fcal, \Zcal) \, p_{\theta}(y_{u_k v_k} | \Ycal_{k-1}, \Fcal, \Zcal)
% p _{\theta}(\Ecal,\Ycal| \Zcal,l) &= \prod_{k \in [l]} p_{\theta}(e_k | \Ecal_{k-1}, \Fcal, \Zcal) p_{\theta}(y_{u_k v_k} | \Ycal_{k-1}, \Fcal, \Zcal),\\
%  p_{\theta}(\Xcal|\Ecal,\Ycal, \Zcal) &= \prod_{u\in \Vcal} p_{\theta}(\xb_u | \Ecal, \Fcal, \Zcal),
\end{align*} 
% the ordering for the edge and edge weights is arbitrary, 
%
where the ordering for the edge and edge weights is independent of node labels and hence permutation invariant, $e_k$ and $y_{u_k v_k}$ denote the $k$-th edge and edge weight under the chosen order, 
and $\Ecal_{k-1} = \{e_1, \ldots, e_{k-1}\}$ and $\Ycal_{k-1} = \{y_{u_1 v_1}, \ldots, y_{u_{k-1} v_{k-1}}\}$ denote the $k-1$ previously generated edges and edge 
weights respectively.
%
% where the ordering for the edge and edge weights is arbitrary, $e_k$ and $y_{u_k v_k}$ denote the $k$-th edge and edge weight under this arbitrary ordering, 
% $\Ecal_{k-1} = \{e_1, \ldots, e_{k-1}\}$ and $\Ycal_{k-1} = \{y_{u_1 v_1}, \ldots, y_{u_{k-1} v_{k-1}}\}$ denote the $k-1$ previously generated edges and edge 
% weights respectively.
%

Moreover, the model characterizes the conditional probabilities in the above formulation as follows.
For each node, it represents all potential values for the atom types $\tb_u = \qb$ as an unnormalized log probability vector (or `logits'), feeds this logit into a softmax distribution 
and samples the node features. 
Then, it represents the average number of edges as a logit, feeds this logit into a Poisson distribution and samples the number of edges.
Next, it represents all potential edges as logits and, for each edge, all potential edge weights as another logit, and it feeds the former vector into a single softmax
distribution and the latter vectors each into a different softmax distribution. 
Moreover, the edge distribution and the corresponding edge weight distributions depend on a set of binary \emph{masks}, which may depend on the sampled node features 
and also get updated every time a new edge and edge weight are sampled. 
By doing so, it prevents the generation of certain \emph{undesirable} edges and edges weights, allowing for the generated graph to fulfill a set of predefined local structural 
and functional properties. 
Finally, for each atom, it samples its coordinates $\xb_u$ from a multidimensional Gaussian distribution whose mean and variance depends on the latent vectors of the corresponding 
atom as well as its neighbors and the underlying chemical bonds.

%
% \manuel{we only have molecule design, skip this}
% For example, in molecule design, masking facilitates the generation of molecules with a valid structure, as shown in the experiments with molecule design.%Section~\ref{sec:experiments-real}.}
%%
% Moreover, the model characterizes the conditional probabilities on the right hand side of the above equation as follows.
% %
% For each node, it represents all potential node feature values as an unnormalized log probability vector (or `logits'), feed this logit into a softmax distribution 
% and sample the node features. 
% %
% Then, it represents all potential edges as a logit and, for each edge, all potential edge weights as another logit, and it feeds the former vector into a single softmax
% distribution and the latter vectors each into a different softmax distribution. 
% %
% The edge distribution and the corresponding edge weight distributions depend on a set of binary \emph{masks}, which may depend on the sampled node features and also get updated 
% every time a new edge and edge weight are sampled. By doing so, it prevents the generation of certain \emph{undesirable} edges and edges weights, allowing for the 
% generated graph to fulfill a set of predefined local structural and functional properties. 
% %
% For example, in molecule design, masking facilitates the generation of molecules with a valid structure, as shown in the experiments on molecule design. %Section~\ref{sec:experiments-real}.
% %
% More formally, the distributions of each node, edge and edge weight are given by:
%

\begin{figure*}[t]
	\centering
	 \includegraphics[width=0.9\textwidth]{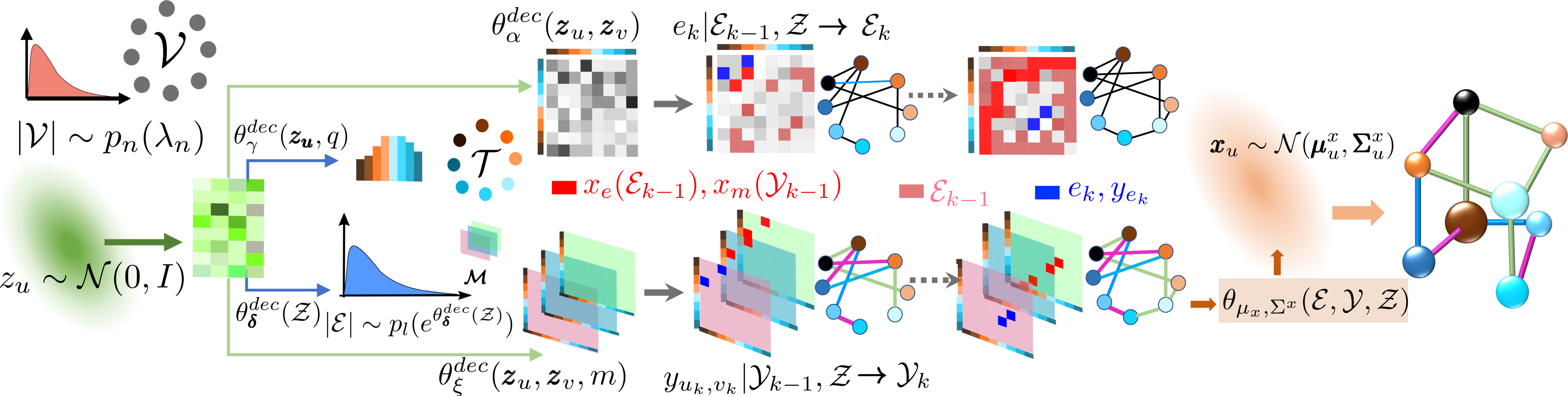}\hspace*{0.1 cm}
	 \vspace*{-1.5mm}
	\caption{The decoder of our variational autoencoder for molecular graphs. From left to right, the decoder first samples the number of nodes $n = |\Vcal|$
	from a Poisson distribution $p_n(\lambda_n)$ and it samples a latent vector $\zb_u$ per node $u \in \Vcal$ from $\Ncal(\mathbf{0}, \Ib)$.
	%
% 	Then, it feeds all latent vectors $\Zcal$ into a nonlinear log intensity function $\theta^{dec} _{\bm{\delta}}(\Zcal)$ which is used to sample the number of edges.
	%
	Then, for each node $u$, % it samples a feature $q$ from a distribution depending on $z_u$-s.
 	it represents all potential node feature values as an unnormalized log probability vector (or `logits'), where each entry is given by a nonlinearity $\theta^{dec}_{\gamma}$ of 
	the corresponding latent representation $\zb_u$, feeds this logit into a softmax distribution and samples the node features. 
        Next, it feeds all latent vectors $\Zcal$ into a nonlinear log intensity function $\theta^{dec} _{\bm{\delta}}(\Zcal)$ which is used to sample the number of edges.
	Thereafter, on the top row, it constructs a logit for all potential edges $(u, v)$, where each entry is given by a nonlinearity $\theta^{dec}_{\alpha}$ of the corresponding latent 
	representations $(\zb_u, \zb_v)$. Then, it samples the edges one by one from a soft max distribution depending on the logit and a mask $\beta_e(\Ecal_{k-1})$, which gets 
	updated every time it samples a new edge $e_k$. % and depends on any local structural or functional constraint the application may require on $\Ecal$.
	On the bottom row, it constructs a logit per edge $(u, v)$ for all potential edge weight values $m$, where each entry is given by a nonlinearity $\theta^{dec}_{\xi}$ 
	of the latent representations of the edge and edge weight value $(\zb_u, \zb_v, m)$. Then, every time it samples an edge, it samples the edge weight 
	value from a soft max distribution depending on the corresponding logit and mask $x_m(u, v)$, which gets updated every time it samples a new $y_{u_k v_k}$.
	Finally, for each atom $u$, it samples its coordinates $\xb_u$ from a multidimensional Gaussian distribution whose mean $\mu_{\xb}$ and variance $\Sigma_{\xb}$ 
	depends on the latent vectors of the corresponding atom and its neighbors and the underlying chemical bonds. 
	}
% 	\caption{The decoder of our variational autoencoder for graphs. From left to right, the decoder first samples the number of nodes $n = |\Vcal|$ and edges $l = |\Ecal|$ 
% 	from two Poisson distributions $p_n(\lambda_n)$ and $p_l(\lambda_l)$ respectively and it samples a latent vector $\zb_u$ per node $u \in \Vcal$ from a multidimensional 
% 	Gaussian distribution $\Ncal(\mathbf{0}, \Ib)$. 
% 	%
% 	Then, on the top row, it constructs an unnormalized log probability vector (or `logit') for all potential edges $(u, v)$, where each entry is given by a nonlinearity 
% 	$\theta^{dec}_{\alpha}$ of the corresponding latent representations $(\zb_u, \zb_v)$. Then, it samples the edges one by one from a soft max distribution depending 
% 	on the logit and a mask $\beta_e(\Ecal_{k-1})$, which gets updated every time it samples a new edge $e_k$ and depends on any local structural or functional constraint 
% 	the application may require on $\Ecal$.
% 	%
% 	On the bottom row, it constructs a logit per edge $(u, v)$ for all potential edge weight values $m$, where each entry is given by a nonlinearity $\theta^{dec}_{\xi}$ 
% 	of the latent representations of the edge and edge weight value $(\zb_u, \zb_v, m)$. Then, every time it samples an edge, it samples the edge weight 
% 	value from a soft max distribution depending on the corresponding logit and mask $x_m(u, v)$, which gets updated every time it samples a new  $y_{u_k v_k}$.
% 	%
% 	}
	\label{fig:decoder}
\end{figure*}

%%
More formally, the distributions of each node feature, the number of edges, each edge and edge weight are given by:
\vspace*{-0.2cm}
\begin{align}
p_{\theta} (\tb_u= \qb|\Zcal) &=  \frac{e^{ \theta^{dec} _{\gammab}(\zb_u, \qb) }}{\sum_{\qb'} e^{ \theta^{dec} _{\gammab}(\zb_{u},\qb')}}, \\
p_{\theta} (l \big|\Zcal) &= p_l(e^{\theta^{dec} _{\bm{\delta}}(\Zcal)}),\nonumber\\ % \label{eq:genmodel0} \\
p_{\theta}(e = (u, v) | \Ecal_{k-1}, \Zcal) &= \frac{\beta_{e} e^{  \theta^{dec}_{\alphab}(\zb_u, \zb_v)  }}{\sum_{e'=(u', v') \notin \Ecal_{k-1}} \beta_{e'} e^{  \theta^{dec}_{\alphab}(\zb_{u'}, \zb_{v'}) } },\nonumber \\%\label{eq:genmodel1} \\
% \end{align}
% \vspace*{-0.1cm}
% \begin{align*}
p_{\theta}(y_{uv} = m | \Ycal_{k-1}, \Zcal) &= \frac{\beta_{m}(u,v) e^ {\theta^{dec}_{\xib}(\zb_u, \zb_v, m) }}{\sum_{m'{} \neq m} \beta_{m'}(u,v) e^{\theta^{dec}_{\xib}(\zb_{u}, \zb_{v}, m'{})}},\nn\\% \label{eq:genmodel2}\\
p_{\theta} (\xb_u| \Ecal, \Ycal,\Zcal) &= \Ncal(\mub_x,\Sigmab_x), \quad \text{with, } [\mub^x _u,\Sigmab^x _u]= [\theta_{\mu^x}(\rb(u)), \theta_{\Sigma^x}(\rb(u))\theta^T _{\Sigma^x}(\rb(u)) ],\nn
\end{align}
where $p_l$ denotes a Poisson distribution, $\beta_e$ is the binary mask for edge $e$ and $\beta_{m}(u,v)$ is the binary mask for feature edge value $m$, 
$\theta^{dec}_{\bullet}$  are neural networks, $\theta_{\Sigma^x}\in \RR^{3\times 3}$, and $\rb(u) = \zb_u+\sum_{v\in\Ncal(u)} y_{uv}\zb_v$.
The parameters of the neural networks do not depend on the number of nodes or edges in the molecular graph and the dependency of the binary masks 
$\beta_e$ and $\beta_{m}(u,v)$ on the node features and the previously generated edges $\Ecal_{k-1}$ and edge weights $\Ycal_{k-1}$ is deterministic and domain 
dependent.
Figure~\ref{fig:decoder} summarizes our decoder architecture.\\
\indent Note that, by using a softmax distribution, it is only necessary to account for the presence of an edge, not its absence, and this, in combination 
with negative sampling, will allow for efficient training and decoding, as it will become clear later in this section. This is in contrast with previous generative models for 
graphs~\cite{welling,graphvae}, which need to model both the presence and absence of each potential edge.
Moreover, we would like to acknowledge that, while masking may be useful to account for prior (expert) knowledge, it may be costly to check for some local (or global) 
structural and functional properties on-the-fly.

\xhdr{Prior} Given a set of $n$ nodes with latent variables $\Zcal = \{ \zb_u \}_{u \in [n]}$, $p_z(\Zcal) \sim \Ncal(\mathbf{0}, \Ib)$.
%we consider a Gaussian prior 

\xhdr{Training}
Given a collection of $N$ molecular graphs $\{ \Gcal_i = (\Vcal_i, \Ecal_i) \}_{i \in [N]}$, each with $n_i$ nodes,  a set of node features $\Fcal_i$, set of node coordinates $\Xcal_i$, set of 
edge weights $\Ycal_i$, we train our variational autoencoder for graphs by maximizing the evidence lower bound (ELBO), as described in the previous section, 
plus the log-likelihood of the Poisson distribution $p_{\lambda_{n}}$ modeling the number of nodes in each graph. Hence we aim to solve:
%
% \vspace*{-0.1cm}
\begin{align}
\underset{\phi,\theta,\lambda_n}{\text{max}}\ \frac{1}{N} \sum_{i \in [N]}&\big(\EE_{q_{\phi}(\Zcal_i | \Vcal_i, \Ecal_i, \Fcal_i, \Ycal_i)} \log p _{\theta}(\Ecal_i, \Ycal_i, \Fcal_i | \Zcal_i) 
 -\text{KL}(q_{\phi} || p_z)  + \log p_{\lambda_{n}}(n_i)\big)\label{eq:obj-old}
\end{align}
%
% \ad{In the above objective, the second term ($\EE_{q_\phi} \log p_{\theta}(\Ecal_i,\Ycal_i,\Fcal_i||Zcal_i)$) ``as it is'' only considers one arbitrary pre-selected order of present edges   }
% To compute the second term in the above objective, we need to specify the order of edges in the training graphs. However, 
%
{Note that, in the above objective, computation of $\EE_{q_{\phi}} \log p _{\theta}(\Ecal_i, \Ycal_i, \Fcal_i | \Zcal_i)$ requires to specify an order of edges present in the graph $\Gcal_i$. 
To determine this order, we use breadth-first-traversals (BFS) with randomized tie breaking during the child-selection step. 
Such a tie breaking method makes the edge order independent of all node labels except for the source node label.
%
% However, this order still depends on the label of source node. 
Therefore, to make it completely permutation invariant, for each graph, we sample the source nodes from an arbitrary distribution. % and marginalize the log-likelihood 
% over all possible source nodes. 
%
More formally, we replace $\log p _{\theta}(\Ecal_i, \Ycal_i, \Fcal_i  | \Zcal_i)$ with $\log \EE_{s\sim\zeta(\Vcal_i) }p _{\theta}(\Ecal_i, \Ycal_i, \Fcal_i  | \Zcal_i)$ for each 
graph $\Gcal_i$, where $s$ is the randomly sampled source node for the BFS, and $\zeta$ is the sampling distribution for $s$.
Note that, the logarithm of a marginalized likelihood is difficult to compute. Fortunately, by using Jensen inequality, we can have a lower bound of the actual likelihood:
% \vspace*{-0.1cm}
\begin{align*}
 \log \EE_{s\sim\zeta(\Vcal_i) }p _{\theta}(\Ecal_i, \Ycal_i, \Fcal_i | \Zcal_i)\ge
  \EE_{s\sim\zeta(\Vcal_i) }\log p _{\theta}(\Ecal_i, \Ycal_i, \Fcal_i  | \Zcal_i)
\end{align*}
Therefore, to train our model, we maximize 
\begin{align}
\hspace*{-0.3cm}\frac{1}{N} \sum_{i \in [N]}& \big(\EE_{q_{\phi}(\Zcal_i | \Vcal_i, \Ecal_i, \Fcal_i, \Ycal_i),  s\sim\zeta(\Vcal_i) }\log p _{\theta}(\Ecal_i, \Ycal_i, \Fcal_i | \Zcal_i) 
-\text{KL}(q_{\phi} || p_z) + \log p_{\lambda_{n}}(n_i)\big), \label{eq:obj}
\end{align}
% where the variables are the neural network $\theta, \phi$ and the parameter $\lambda_n$. 
% Note that, computing the exact value of $\EE_{s\sim\zeta(\Vcal_i) }\log p _{\theta}(\Ecal_i, \Ycal_i, \Fcal_i | \Zcal_i)$
% is not feasible, and hence in practice, we compute the Monte-Carlo estimate for a set of samples. 
% More details about the training procedure,
% are provided in Appendix. 
The following theorem points out the key property of our objective function.
\begin{theorem}\label{prop-2}
If the source distribution $\zeta$ does not depend on the node labels, then the parameters learned by maximizing the objective in Eq.~\ref{eq:obj} are invariant to the permutations of the node labels.
\end{theorem}

\begin{proof}
For each training graph $\Gcal_i=(\Vcal_i,\Ecal_i)$, we denote each corresponding component in the objective function as
\begin{align*}
 \Lcal_{\Gcal_i}(\Theta)= &\EE_{q_{\phi}(\Zcal_i | \Vcal_i, \Ecal_i, \Fcal_i, \Ycal_i),s\sim\zeta(\Vcal_i)} \log p_{\theta}(\Ecal_i, \Ycal_i, \Fcal_i | \Zcal_i) 
  -\text{KL}(q_{\phi} || p_z) + \log p_{\lambda_{n}}(n_i),
\end{align*}
where $\Theta$ is the set of trainable parameters.
%
% \manuel{trivial}
% So the actual objective function is $\Lcal(\Theta)=\frac{1}{N}\sum_{i=1} ^N \Lcal_{\Gcal_i}(\Theta)$.
%

Then, to prove that the parameters $\hat{\Theta}$ estimated by maximizing $\Lcal(\Theta)$ are invariant to the permutations of the labels of all $\Vcal_i$'s, it is enough to prove that
$\Lcal_{\Gcal_i}(\Theta)$ is invariant to the  permutation of $\Vcal_i$ for any $i\in[N]$, and for any $\Theta$.
Moreover, note that $\log p_{\lambda_{n}}(n_i)$ depends on the total number of nodes and edges, and therefore is node permutation invariant.
Therefore, it is enough to prove the permutation invariance property of the first two components, \ie,
$\text{KL}(q_{\phi} || p_z)$ and $\EE_{q_{\phi}(\Zcal_i | \Vcal_i, \Ecal_i, \Fcal_i, \Ycal_i),s\sim\zeta(\Vcal_i)} \log p_{\theta}(\Ecal_i, \Ycal_i, \Fcal_i | \Zcal_i)$. Since $q_{\phi}$ and $p_z$ are both
normal distribution, we have:
\begin{align}
\text{KL}(q_{\phi} || p_z)= \frac{1}{2}&\Big( \tr (\Sigmab_p ^{-1} \Sigmab_q)+(\mub_p-\mub_q)^T \Sigmab^{-1} _p (\mub_p-\mub_q)  
  -kn_i+ \log   {\text{det} \ \Sigmab_p }-\log {\text{det} \ \Sigmab_q }\Big) 
\end{align}
which, in our case, reduces to:
\begin{align}
\frac{1}{2}\Big(\sum_{u\in\Vcal_i}(\one^T\sigmab_u^2+\one^T\mub_u^2)-kn_i -\one^T \log \sigmab^2 _{u}\Big).
\end{align}
Note that, from Proposition~\ref{prop-1}, we know that the values of $\cb_u(k)$ are invariant to the permutation of node labels. Now, since $[\mub_u, \diag(\sigmab_u)]=\enc(\cb_u(1),\hdots,\cb_u(K))$, 
$\text{KL}(q_{\phi} || p_z)$ is also invariant to the permutation of node labels.
Now, to prove that $\EE_{q_{\phi}(\Zcal_i | \Vcal_i, \Ecal_i, \Fcal_i, \Ycal_i)} \log p_{\theta}(\Ecal_i, \Ycal_i, \Fcal_i | \Zcal_i)$, we rely on a reparameterization trick for the normal distribution.
\begin{align*}
 &\EE_{q_{\phi}(\Zcal_i | \Vcal_i, \Ecal_i, \Fcal_i, \Ycal_i),s\sim\zeta(\Vcal_i)} \log p_{\theta}(\Ecal_i, \Ycal_i, \Fcal_i | \Zcal_i)=\EE_{\bm{\epsilon}_{u\in\Vcal_i}\sim \Ncal(\bm{0},\Ib),s\sim\zeta(\Vcal_i)}\log p_{\theta}\Big(\Ecal_i, \Ycal_i, \Fcal_i |(\mub_u+\diag(\sigmab_u)\bm{\epsilon}_u)_{u\in\Vcal_i}\Big)
\end{align*}
Note that, 
\begin{enumerate}
 \item $\zeta(\Vcal_i)$ does not depend on node labels (e.g. uniform sampling, degree based sampling, etc.);
 \item the edge sequence of $\Ecal_i$ is determined by BFS with randomized tie breaking;
 \item $\epsilon_u$ does not depend on $u$ since it is sampled from $ \Ncal(\bm{0},\Ib)$.
\end{enumerate}
% 
%  This, together with the fact that the edge sequence is determined by BFS with randomized tie breaking,
% makes $\EE_{s\sim\zeta(\Vcal_i)}\log p_{\theta}\Big(\Ecal_i, \Ycal_i, \Fcal_i |(\mub_u+\diag(\sigmab_u)\bm{\epsilon}_u)_{u\in\Vcal_i}\Big)$ invariant of node permutations. 
% %
% $\epsilon_u$ does not depend on $u$ since it is sampled from $ \Ncal(\bm{0},\Ib)$. 
% 
The facts, $(1)-(3)$, along with the permutation invariance property of $\mub_u$ and $\diag(\sigmab_u)$, conclude the proof.
\end{proof}
% Note that the above object
% manuel: We talked about this earlier already.
%
% \ad{\begin{proposition}\label{prop-1}
% The parameters learned by our models \ie\ $\Wb_{1},..,\Wb_K$ and those of the neural networks $\theta^{dec}_{\gammab}$, $\theta^{dec}_{\alphab}$ and $\theta^{dec}_{\xib}$  do not depend on the size of the input 
%graphs.
% \end{proposition}
% The proof is given in Appendix (TBD.)}
%
\xhdr{Scalability and implementation details} 
In terms of scalability, the major bottleneck is computing the gradient of the first term in Eq.~\ref{eq:obj} during training, rather than encoding and decoding graphs once the model is 
trained. More specifically, given a source node for a network without masks, an exact computation of the per edge partition function of the log-likelihood of the edges, \ie,
$\sum_{e'=(u', v') \notin \Ecal_{k-1}} \exp ( \theta^{dec}_{\alphab}(\zb_{u'}, \zb_{v'}) )$,
requires $O(|\Vcal|^2)$ computations, similarly as in most inference algorithms for existing generative models of graphs, and hence is costly to compute even for medium networks.
Fortunately, in practice,  we can approximate such partition function using negative sampling~\cite{mikolov2013distributed} 
which reduces the likelihood computation to $O(l)$, where $l = |\Ecal|$ is the number of (true) edges in the graph. Therefore, for $S$ samples
of source nodes, the complexity becomes $O(Sl)$. Here, note that most real-world graphs are sparse and thus $l  \ll |\Vcal|^2$.

\section{Property Oriented Molecule Generation}
\label{sec:prop}
In this section, we aim to optimize the probabilistic decoder of our variational autoencoder, described in Section~\ref{sec:model}, so that it learns to generate
molecules that maximize certain molecular property (\eg, solubility in water).
%
% To that aim, we augment the decoder of the above model to build an alternative decoder $p_{\theta}( \cdot)$ that can generate molecules specifically optimized for the given property.
%
To this aim, we approach the problem from the perspective of variational inference and show that the optimal property-oriented decoder can be expressed in terms of the 
original decoder and the value of the molecular property.
%
% , where the optimized decoder achieves an optimal trade off between the fidelity to the 
% original decoder and the maximization of the molecular property, and show that  
%
This result means that we can obtain molecules from the optimal property-oriented decoder just by applying rejection sampling on the molecules samples from the original decoder. 
However, in practice, such a naive sampling strategy will be inefficient and impractical, specially given the high-dimensional nature of the data. 
Therefore, we design a practical method for approximating the optimal property-oriented decoder, which iteratively adapts the parameters of a (parameterized) property-oriented 
decoder using a stochastic gradient-based algorithm.
%
% first formally state the propertuy problem cast this task 
% as an optimization problem which trades off the fidelity of the generative process 
% to the previous model $p_{\theta}$ and 
%
% , and then show that, the optimal decoder $p^* _{\theta}$
% can be expressed in terms of the orginal decoder model and the value of the given property. Finally, we  
% provide a stochastic gradient-based method that estimates the augmented decoder $p_{\theta}$
% which approximates the optimal decoder $p^* _{\theta}$, within a class of parameterized models $\Pcal(\theta)$. 

\xhdr{Property-oriented decoder design using variational inference} 
Let $p_{\theta}$ be our original generative model (decoder), which has been trained using a given collection of molecular graphs and $\ell(\cdot)$ be a loss function, 
which penalizes low values of the molecular property value of interest. 
Then, we construct the optimal property-oriented decoder $p^{*}$ by solving the following optimization problem:
% to guide the generative process for property specific molecule design
% 
% that optimally trades off the fidelity to the trained model $p_{\theta}$ and the loss $V( \cdot)$, by solving the following optimization problem:
%
\begin{equation}
 \underset{p(\cdot | \Zcal)} {\text{minimize}} \quad \EE_{\Zcal\sim p_z( \cdot)} \left[ \EE_{\Ecal, \Ycal,\Fcal \sim p(\cdot | \Zcal)} \left[ S(\Ecal, \Ycal,\Fcal | \Zcal) \right] \right], \label{eq:kl}
 \end{equation}
with
\begin{equation}
S(\Ecal, \Ycal,\Fcal | \Zcal) = \ell(\Ecal, \Ycal,\Fcal) + \rho \log \frac{p( \Ecal, \Ycal,\Fcal |\Zcal) }{ p_{\theta} ( \Ecal, \Ycal,\Fcal |\Zcal)}, \label{eq:S-phi}
\end{equation}
where the inner expectation is taken over all molecules generated using the property-oriented decoder $p(\cdot | \Zcal)$ given the latent vectors $\Zcal$, the outer expectation is 
taken over all possible latent vectors under the prior distribution\footnote{If a molecule is given, instead of the prior distribution, one may also consider using the posterior $q_\phi( \cdot | \Vcal, \Ecal, \Fcal, \Ycal)$.}
$p_{z}(\zb_1, \ldots, \zb_n)$ with $n \sim \text{Poisson}(\lambda_n)$, and we do not assume any specific parametric form for the property-oriented decoder $p$.
In Eq.~\ref{eq:S-phi}, the first term penalizes molecules with a low value of the property of interest, the second term penalizes property-oriented decoders whose generated molecules
differ more from those that the original decoder would generate and the parameter $\rho$ controls the trade off between both terms.
Here, note that the second term provides an inductive bias that ensures that the molecules generated by the property-oriented decoder are plausible.
Moreover, we can rewrite the inner expectation of the second term in terms of Kullback-Leibler (KL) divergence~\cite{kullback1951information}, \ie, 
\begin{equation*}
\text{KL}\left[ p( \cdot | \Zcal) || p_{\theta}( \cdot | \Zcal) \right] = \EE_{\Ecal, \Ycal,\Fcal \sim p(\cdot | \Zcal)} \left[ \log \frac{p( \Ecal, \Ycal,\Fcal |\Zcal) }{ p_{\theta} ( \Ecal, \Ycal,\Fcal |\Zcal)} \right]
\end{equation*}
which is commonly used as a \emph{distance} measure between distributions.

Then, it is straightforward to show that the above optimization problem is equivalent to the following problem:
\begin{equation}
 \underset{p(\cdot | \Zcal)} {\text{minimize}} \quad \EE_{\Zcal\sim p_z( \cdot)} \left[ \text{KL} \left[ {p(\cdot |\Zcal) } || { g_{\theta} (\cdot|\Zcal)  } \right] \right] \label{eq:klreduce}
\end{equation}
where 
\begin{equation*}
g_\theta(\Ecal, \Ycal,\Fcal |\Zcal)=\frac{p_{\theta} ( \Ecal, \Ycal,\Fcal | \Zcal) \exp( -\frac{\ell(\Ecal, \Ycal,\Fcal)}{\rho}) }{ \EE_{\Ecal', \Ycal', \Fcal' \sim p_{\theta}( \cdot |\Zcal)}\left[\exp( -\frac{\ell(\Ecal', \Ycal', \Fcal')}{\rho})\right]}.  
\end{equation*}
The above objective function achieves its global minimum of zero if the numerator and the denominator are equal. Thus, the optimal
property-oriented decoder is just given by:
\begin{equation}
 p^*(\Ecal, \Ycal, \Fcal |\Zcal)= \frac{p_{\theta} (\Ecal, \Ycal, \Fcal | \Zcal) \exp( -\frac{\ell(\Ecal, \Ycal, \Fcal)}{\rho}) } { \EE_{\Ecal', \Ycal', \Fcal' \sim p_{\theta}(\cdot |\Zcal)}\left[\exp( -\frac{\ell(\Ecal', \Ycal', \Fcal')}{\rho})\right]} \label{eq:optimal-property-oriented-prob-decoder}
\end{equation}
The above result has an important implication. It means that we can use sampling methods to obtain (unbiased) samples from the optimal property-oriented decoder. For example,
we can apply rejection sampling on the molecules generated by the original decoder $p_{\theta}$, where we accept or reject them according to the (exponentiated) property value of 
interest. 
%
% implicates that we can utilize rejection sampling methods, where we first sample the molecules using the trained decoder $p_{\theta}$, and then accept or reject them according to the (exponentiated) loss 
% function.
%
However, in practice, these sampling methods may be inefficient if the generated molecules under the original decoder $p_{\theta}$ have low probability under the optimal property-oriented 
decoder model. Given that molecules are high dimensional objects, this is specially problematic due to the curse of dimensionality.
Next, we will design a practical method for approximating $p^*(\Ecal, \Ycal, \Fcal | \Zcal)$, which iteratively adapts the parameters of a (parameterized) property-oriented model using a stochastic 
gradient-based algorithm.

\xhdr{A stochastic gradient-based algorithm}
In this section, we aim to find a property-oriented decoder $p_{\theta'}$ within the class of parameterized probabilistic decoders defined by Eq.~\ref{eq:prob-decoder} that 
approximates well the optimal property-oriented decoder $p^{*}$ that minimizes the objective function in Eq.~\ref{eq:kl}, \ie,
\begin{equation*}
\EE_{\Zcal\sim p_z( \cdot)} \left[ \EE_{\Ecal, \Ycal,\Fcal \sim p(\cdot | \Zcal)} \left[ S(\Ecal, \Ycal,\Fcal | \Zcal) \right] \right]
\end{equation*}
To this aim, we introduce a general gradient-based algorithm, which iteratively update the parameters $\theta'$ of the parameterized property-oriented decoder $p_{\theta'}$ using 
stochastic gradient descent (SGD)~\cite{kiefer1952stochastic}, \ie,
\begin{align*}
 \theta'_{j+1} &= \theta'_j + \alpha_j \, \nabla_{\theta'} \EE_{\Zcal\sim p_z( \cdot)} \left[ \EE_{\Ecal, \Ycal, \Fcal \sim p_{\theta'}( \cdot | \Zcal)} \left[ S_{\theta'}(\Ecal, \Ycal, \Fcal | \Zcal) \right] \right] |_{\theta' = \theta'_j} \\
 &=\theta'_j + \alpha_j \, \EE_{\Zcal\sim p_z( \cdot)} \left[ \nabla_{\theta'} \EE_{\Ecal, \Ycal, \Fcal \sim p_{\theta'}( \cdot | \Zcal)} \left[ S_{\theta'}(\Ecal, \Ycal, \Fcal | \Zcal) \right] \right] |_{\theta' = \theta'_j},
\end{align*}
where 
\begin{equation*}
S_{\theta'}(\Ecal, \Ycal,\Fcal | \Zcal) = \ell(\Ecal, \Ycal,\Fcal) + \rho \log \frac{p_{\theta'}( \Ecal, \Ycal,\Fcal |\Zcal) }{ p_{\theta} ( \Ecal, \Ycal,\Fcal |\Zcal)}, 
\end{equation*}
$\alpha_j > 0$ is the learning rate at step $j$, and $\theta'_0 = \theta$.

In the above, it may seem challenging to compute a finite sample estimate of the gradient of the function $\EE_{\Ecal, \Ycal, \Fcal \sim p_{\theta'}(\cdot | \Zcal)} \left[ S_{\theta'}(\Ecal, \Ycal, \Fcal |\Zcal) \right]$ since 
the derivate is taken with respect to the parameters of the property-oriented decoder $p_{\theta'}$, which we are trying to learn.
However, we can overcome this challenge using the log-derivative trick~\cite{williams1992simple}:
 \begin{equation*}
% \nabla_{\theta'} \mathcal{C}(p_\theta|\Zcal) &= \nabla_{\theta'} \left\langle S_\theta(\Ecal, \Ycal, \Fcal|\Zcal) \right\rangle_{p_\theta}\\
\nabla_{\theta'} \EE_{\Ecal, \Ycal, \Fcal \sim p_{\theta'}( \cdot | \Zcal)} \left[ S_{\theta'}(\Ecal, \Ycal, \Fcal|\Zcal) \right]
 % = \nabla_{\theta'} \sum_{\Ecal, \Ycal, \Fcal} p_\theta(\Ecal, \Ycal, \Fcal|\Zcal)S_\theta(\Ecal, \Ycal, \Fcal|\Zcal)\\
% & = \sum_{\Ecal, \Ycal, \Fcal} \left(\nabla_{\theta'} p_{\theta'}(\Ecal, \Ycal, \Fcal|\Zcal)\right)S_{\theta'}(\Ecal, \Ycal, \Fcal|\Zcal) + \sum_{\Ecal, \Ycal, \Fcal} p_{\theta'}(\Ecal, \Ycal, \Fcal | \Zcal)\nabla_{\theta'} 
%\rho\log p_{\theta'}(\Ecal, \Ycal, \Fcal |\Zcal)\\
% &=\sum_{\Ecal, \Ycal, \Fcal} \left(p_{\theta'}(\Ecal, \Ycal, \Fcal | \Zcal) \nabla_{\theta'} \log p_{\theta'}(\Ecal, \Ycal, \Fcal |\Zcal)\right)S_{\theta'}(\Ecal, \Ycal, \Fcal|\Zcal) + \rho\nabla_{\theta'}\sum_{\Ecal, \Ycal, 
% \Fcal} p_{\theta'}(\Ecal, \Ycal, \Fcal|\Zcal)\\
% &=\sum_{\Ecal, \Ycal, \Fcal} \left(p_{\theta'}(\Ecal, \Ycal, \Fcal | \Zcal) \nabla_{\theta'} \log p_{\theta'}(\Ecal, \Ycal, \Fcal|\Zcal)\right)(S_{\theta'}(\Ecal, \Ycal, \Fcal |\Zcal) + \rho)\\
=    \EE_{\Ecal, \Ycal, \Fcal \sim p_{\theta'}( \cdot | \Zcal)} \left[ (S_{\theta'}(\Ecal, \Ycal, \Fcal | \Zcal) + \rho)\nabla_{\theta'} \log p_{\theta'}(\Ecal, \Ycal, \Fcal | \Zcal) \right].
\end{equation*}
The above expression readily yields the following unbiased finite sample Monte Carlo estimator:
\begin{align}
  \nabla_{\theta'} \EE_{\Ecal, \Ycal, \Fcal \sim p_{\theta}( \cdot | \Zcal)} \left[ S_{\theta'}(\Ecal, \Ycal, \Fcal|\Zcal) \right] \approx \frac{1}{M}\sum_{i=1} ^M (S_\theta(\Ecal_i, \Ycal_i, \Fcal_i | \Zcal) + \rho)\nabla_{\theta'} \log p_{\theta'}(\Ecal_i, \Ycal_i, \Fcal_i | \Zcal),
\end{align}
where $M$ is total number of sampled molecules generated from the property-oriented decoder $p_{\theta'}$. 
Algorithm~\ref{alg:prop} summarizes the overall procedure.
\begin{algorithm}[t!]
\caption{\textsc{PropertyOrientedDecoder}: it trains a parameterized property-oriented decoder.}
\label{alg:prop}
  \begin{algorithmic}[1]
    \STATE \textbf{Given: }The loss function $\ell(\cdot)$, parameter $\rho$, original decoder $p_{\theta}$, \# of iterations $M$, mini batch size $B$, and learning rate $\gamma$
        \STATE $\theta'_0 \leftarrow \theta$
        \FOR{$j=1,\ldots, M$}  
        \STATE $\Zcal_j \sim p_z( \cdot)$
        %\STATE $p_{\phi^j} \leftarrow $
        \STATE $\Dcal \leftarrow \textsc{Minibatch} ( p_{\theta'_j}( \cdot| \Zcal_j), B)$  
        \STATE $\nabla \leftarrow 0$
        \FOR{$(\Ecal_i, \Ycal_i, \Fcal_i) \in \Dcal$}        
                \STATE $S \leftarrow \ell(\Ecal_i, \Ycal_i, \Fcal_i) + \rho \log \left ( p_{\theta'_j}(\Ecal_i, \Ycal_i, \Fcal_i | \Zcal_j) / p_{\theta}(\Ecal_i, \Ycal_i, \Fcal_i | \Zcal_j ) \right)$
                \STATE $\nabla \leftarrow \nabla + \left( S + \rho \right) \nabla_{\theta'} \log p_{\theta'_j}(\Ecal_i, \Ycal_i, \Fcal_i | \Zcal_j )$
         \ENDFOR
         \STATE $\theta'_{j+1} \leftarrow \theta'_{j} + \gamma \, \frac{\nabla}{B}$
        \ENDFOR
        \STATE \textbf{Return} $\theta'_{M}$
  \end{algorithmic}
\end{algorithm}
% \section{Experiments on Synthetic Graphs}
% \label{sec:experiments-syn}
% \input{050expt-synthetic}

\section{Experiments}
\label{sec:experiments-real}
In this section, we first show that \ourmodel\ beats several state of the art machine learning models for molecule design~\cite{dai2018syntax,gomez2016automatic,kusner2017grammar,graphvae,jin2018junction,liu2018constrained} 
in terms of several relevant quality metrics, \ie, \emph{validity}, \emph{novelty} and \emph{uniqueness}. Then, we also 
show that the continuous latent representations of molecules that our model finds are smooth.
Finally, we demonstrate that the property-oriented decoder provided by Algorithm~\ref{alg:prop} is able to generate molecules 
that maximize certain desirable properties more effectively than several baselines based on Bayesian optimization and 
reinforcement learning.
Appendix~\ref{app:experiments-syn} contains additional experiments on synthetic data.

\subsection{Experimental setup}
We sample $\sim$$10{,}000$ drug-like commercially available molecules from the ZINC dataset~\cite{irwin2012zinc} with $\EE[n]=44$ atoms 
and $\sim$$10{,}000$ molecules from the QM9 dataset~\cite{ramakrishnan2014quantum,ruddigkeit2012enumeration} with $\EE[n]=21$ atoms.
For each molecule, we construct a molecular graph, where nodes are the atoms, the node features are the type
of the atoms \ie\ $\fb_u \in \{C, H,  N, O\}$, edges are the bonds between {two} atoms, and the weight associated with an edge is the type of bonds (single, double 
or triple)\footnote{\scriptsize We have not selected any molecule whose bond types are others than these three.}.
Then, for each dataset, we train our variational autoencoder for molecular graphs using batches comprised of molecules 
with the same number of nodes\footnote{\scriptsize We batch graphs with respect to the number of nodes for efficiency 
reasons since, every time that the number of nodes changes, we need to change the size of the computational graph in Tensorflow.}.
Finally, we sample $10^6$ molecular graphs from each of the (two) trained variational autoencoders using: (i) $\Gcal \sim p_{\theta}(\Gcal | \Zcal)$, 
where $\Zcal \sim p(\Zcal)$ and (ii) $\Zcal \sim p_{\theta}(\Zcal | G=G_T)$, where $G_T$ is a molecular graph from the corresponding (training) dataset. 
In the above procedure, we only use masking on the weight (\ie, type of bond) distributions both during training and sampling to ensure that the valence of the nodes 
at both ends are valid at all times, \ie, $x_m(u, v) = \II( m + n_k(u) \leq m_{max}(u) \wedge m + n_k(v) \leq m_{max}(v) )$, where $n_k(u)$ is the current valence of 
node $u$ and $m_{max}(u)$ is the maximum valence of node $u$, which depends on its type $\fb_u$.
Moreover, during sampling, if there is no valid weight value for a sampled edge, we reject it. 
To assess to which extent masking helps, we also train and sample from our model without masking. Here, we would like to highlight that, while using masking during test 
does not lead to a significant increase in the time it takes to generate a graph, using masking during training does lead to an increase of $5$\% in training time.

\subsection{Quality of the generated molecules}
We first make a quantitative analysis of our model by comparing the quality of the molecules generated by our trained models against the molecules generated by 
several state of the art competing methods and then provide a qualitative analysis by demonstrating that the latent space of the molecules inferred by our model is 
smooth.
For the quantitative  analysis, we use eight baselines for comparison:
 (i) GraphVAE~\cite{graphvae}, 
(ii) GrammarVAE~\cite{kusner2017grammar}, (iii) CVAE~\cite{gomez2016automatic}, (iv) SDVAE~\cite{dai2018syntax}, (v) JTVAE~\cite{jin2018junction}, (vi) CGVAE~\cite{liu2018constrained},
 %(vii) ORGAN ~\cite{organ}
 (vii) MOLGAN ~\cite{decao2018molgan} , (viii) ORGAN ~\cite{organ}, and (ix) GCPN ~\cite{gcpn}.
% , and (viii) GCPN ~\cite{gcpn}.
%
% The last two methods utilize chemical SMILES while GraphVAE uses molecular graphs. 
Among them, GraphVAE, JTVAE, CGVAE, MOLGAN and GCPN use molecular graphs and GrammarVAE, CVAE, SDVAE, JTVAE and ORGAN use SMILES strings, a 
domain specific textual representation of molecules. 
\begin{table*}[t]
    %     \hspace*{-0.5cm}
    \centering
    \scalebox{0.6}{
     \begin{tabular}{| l | l | l | l | l | l | l | l |l|}
         \hline
         \multicolumn{1}{|c|}{} &
         \multicolumn{8}{c|}{\textbf{Novelty}}
         \\
         \hline
         {\textbf{Dataset}}&
         {\textbf{\ourmodel}}&
         {\textbf{\ourmodel$^{*}$}}&
         {\textbf{GraphVAE}}&
         {\textbf{GrammarVAE}} &
         {\textbf{CVAE}} &
         {\textbf{SDVAE}} &
         {\textbf{JTVAE}} & \textbf{CGVAE} \\
         \hline
         \textbf{ZINC} &
         \specialcell{1.000} &
         \specialcell{1.000}& \specialcell[]{-} &\specialcell{ 1.000}& \specialcell{0.980} &\specialcell{1.000} & \specialcell[]{0.999} &1.000 \\
         \hline
         \textbf{QM9} &
         \specialcell{1.000} &
         \specialcell{1.000}&
         \specialcell{0.661}
         & \specialcell{1.000}& \specialcell{0.902}&\specialcell{-} & \specialcell[]{1.000}&0.943\\ 
         \hline
	 \hline
			\multicolumn{1}{|c|}{} & \multicolumn{8}{c|}{\textbf{Uniqueness}} \\
			\hline
			{\textbf{Dataset}}&
			{\textbf{\ourmodel}}&
			{\textbf{\ourmodel$^{*}$}}&
			{\textbf{GraphVAE}}&
			{\textbf{GrammarVAE}} &
			{\textbf{CVAE}} &
			{\textbf{SDVAE}}&
			{\textbf{JTVAE}}& \textbf{CGVAE}\\
			\hline
			\textbf{ZINC} &
			\specialcell{0.999 } & \specialcell[]{0.588}&
			\specialcell{-}&
			\specialcell{0.273} & \specialcell{0.021} &
			\specialcell{1.000 } &  \specialcell[]{0.991 }&0.998\\ 
			\hline
			\textbf{QM9} &
			\specialcell{0.998} &
			\specialcell{0.676}&
			\specialcell{0.305 } &\specialcell{0.197}&
			\specialcell{0.031}&
			\specialcell{-} & \specialcell[]{0.371}& 0.986  \\ 
			\hline
			
		\end{tabular}
	}
	\caption{Novelty and Uniqueness of the molecules generated using  \ourmodel\ and all baselines.
		The sign $^{*}$ in\-di\-cates no masking.
		For both the datasets, we report the number over $10^6$ valid sampled molecules.}
	\label{tab:unique}
\end{table*}
\begin{table*}[t]
    \centering
    \resizebox{\hsize}{!}{
          \begin{tabular}{| l | l | l | l | l | l | l | l |l|l|l|l|l|}
              \hline
              \multicolumn{2}{|c|}{} & \multicolumn{11}{c|}{\textbf{Validity}} \\
              \hline
              {\textbf{Dataset}}&
              \textbf{Sampling type}&
              {\textbf{\ourmodel}}&
              {\textbf{\ourmodel$^{*}$}}&
              {\textbf{GraphVAE}}&
              {\textbf{GrammarVAE}} &
              {\textbf{CVAE}} &
              {\textbf{SDVAE}}&
              {\textbf{JTVAE}}&{\textbf{CGVAE}}& \textbf{ORGAN} &  \textbf{MOLGAN}& \textbf{GCPN}\\
              \hline
              \textbf{ZINC} &
              \specialcell{\textbf{$\Zb\sim P(\Zb)$} \\ \textbf{$\Zb\sim P(\Zb |G_T)$}} &
              \specialcell{1.000 \\ 1.000} & \specialcell[]{0.590  \\ 0.580}&
              \specialcell{0.135\\-} & \specialcell{0.440\\0.381} & \specialcell{0.021\\0.175} &\specialcell{0.432 \\-} & \specialcell[]{1.000\\1.000}&\specialcell[]{1.000\\-} & \specialcell[]{0.240\\-} & \specialcell[]{1.000\\-}&\specialcell[]{1.000\\-}\\ % & 0.22 \\ %& 1.00 \\
              \hline
              \textbf{QM9} &
              \specialcell{\textbf{$\Zb\sim P(\Zb)$} \\ \textbf{$\Zb\sim P(\Zb | G_T)$}} &
              \specialcell{0.999\\ 0.999} &
              \specialcell{0.682 \\ 0.660}&
              \specialcell{0.458\\-}&
              \specialcell{0.200\\0.301}&
              \specialcell{0.031\\0.100} & \specialcell{-\\-}& \specialcell[]{0.997\\0.965}& \specialcell[]{1.000\\-}& \specialcell[]{-\\-} &\specialcell[]{-\\-} & \specialcell[]{-\\-}\\ % & - \\ % & 1.00 \\
              \hline
             
          \end{tabular}
    }
    \caption{Validity of the molecules generated using \ourmodel\ and all baselines. The sign $^{*}$ in\-di\-cates no masking. For both the datasets, we 
    report the numbers over $10^6$ sampled molecules.
    }
       \vspace*{-5mm}
    \label{tab:validity}
\end{table*}
Moreover, we use the following evaluation metrics for performance comparison:

\begin{itemize}[noitemsep,nolistsep,leftmargin=0.7cm]
\item[(i)] \emph{Novelty}: we use this metric to evaluate to which degree a method generates novel molecules, \ie, molecules which were not present in the (training) 
	dataset, \ie\ $\text{Novelty} = 1 -  {| \Ccal_s \cap \Dcal | }/{| \Ccal_s |}$,
	where $\Ccal_s$ is the set of generated molecules which are chemically valid, $\Dcal$ is the training dataset, and $\text{Novelty} \in [0, 1]$.
\item[(ii)] {\emph{Uniqueness}}: we use this metric to evaluate to what extent a method generates unique chemically valid molecules. We define, $\text{Uniqueness} =  {|\text{set}(\Ccal_s)|}/{n_s}$
	where $n_s$ is the number of generated molecules and $\text{Unique} \in  [0, 1]$.
\item[(iii)] \emph{Validity}: we use this metric to evaluate to which degree a method generates chemically valid molecules\footnote{\scriptsize We used the opensource cheminformatics suite RDkit (\url{http://www.rdkit.org}) 
to check the validity of a generated molecule.}. That is, $\text{Validity} =  {|{\Ccal_s}|}/{n_s}$ where $n_s$ is the number of generated molecules, $\Ccal_s$ is the set of generated molecules which are chemically valid, and 
note that $\text{Validity} \in  [0, 1]$.
\end{itemize}

Tables~\ref{tab:unique}--\ref{tab:validity} compare our trained models to the state of the art methods above in terms of novelty, uniqueness, and validity.
For GraphVAE and CGVAE we report the results reported in the paper and, for SDVAE, since there is no public domain implementation of these methods at the
time of writing, we have used the sampled molecules from the prior provided by the authors for the ZINC dataset. For CVAE, GrammarVAE, JTVAE, ORGAN, MOLGAN and GCPN, we run their public domain implementations in the same set of molecules that we used. For MOLGAN, ORGAN and GCPN, we only report the validity on the discovered molecules and
refrain comparing their performance in terms of novelty and uniqueness given that their focus is on generating molecules maximizing certain property value.

We find that, in terms of novelty, both our trained models and all competing methods except for the GraphVAE, which assumes a fixed number of nodes, are able to (almost) always 
generate novel molecules. 
However, we would also like to note that novelty is only defined over chemically valid molecules. Therefore, despite having (almost) perfect novelty scores, GraphVAE, 
GrammarVAE, CVAE and SDVAE generate significantly fewer novel molecules than our method.
{In terms of uniqueness, which is defined over the set of sampled molecules, we observe that all baseline methods, except CGVAE (for ZINC and QM9) and JTVAE (for ZINC), perform 
very poorly in both datasets in comparison with our method.}
In terms of validity, our trained models significantly outperform four competing methods--- GraphVAE, GrammarVAE, CVAE, SDVAE and ORGAN--- even without the use of masking,
and achieve comparable performance to JTVAE, CGVAE and GCPN.

We would like to highlight that, in contrast to our model, GrammarVAE, CVAE and SDVAE use SMILES, a domain specific string based representation, and 
thus they may be constrained by its limited expressiveness. Among them, GrammarVAE and SDVAE achieve better performance by using grammar to favor 
valid molecules.
GraphVAE generates molecular graphs, as our model, however, its performance is inferior to our method because it assumes a fixed number of nodes, it samples 
edges independently from a Bernoulli distribution, and is not permutation invariant.

\begin{figure}[!t] 
	\centering
		\scalebox{0.4}{\begin{tabular}{cccccc}
			\setlength{\fboxsep}{-0.3pt}
			\setlength{\fboxrule}{-0.2pt}
			\includegraphics[width=0.22\textwidth]{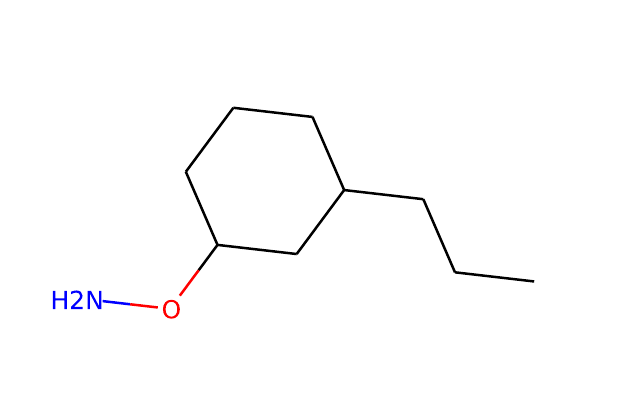}\hspace*{-0.8cm} & %\vspace{-1mm} &
			\includegraphics[width=0.22\textwidth]{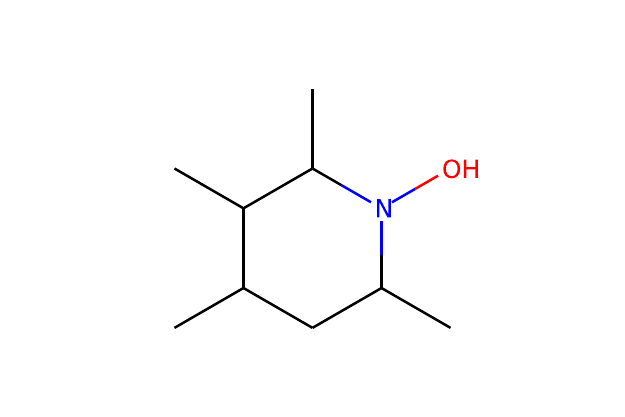}\hspace*{-0.8cm} & %\vspace{-1mm} &
			\includegraphics[width=0.22\textwidth]{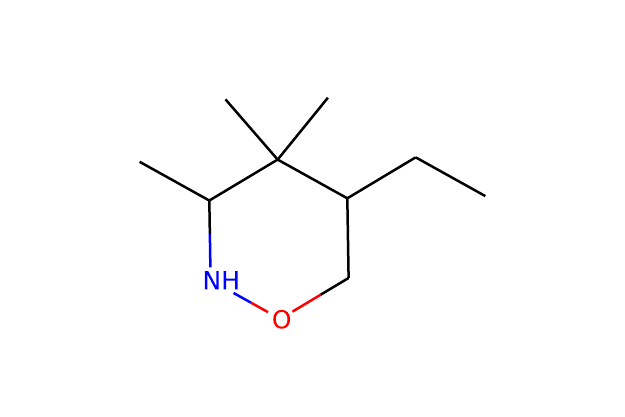}\hspace*{-0.8cm} &
			\includegraphics[width=0.22\textwidth]{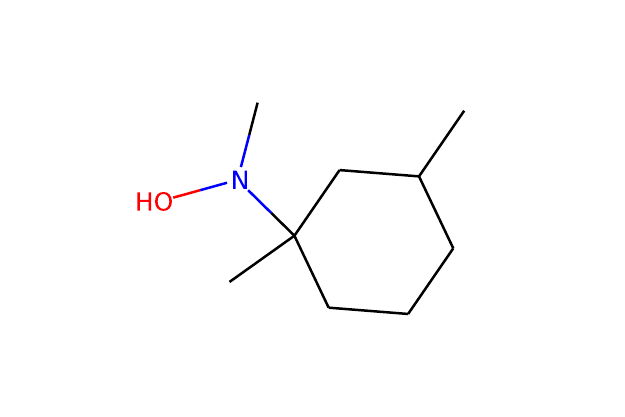}\hspace*{-0.8cm} &
			\includegraphics[width=0.22\textwidth]{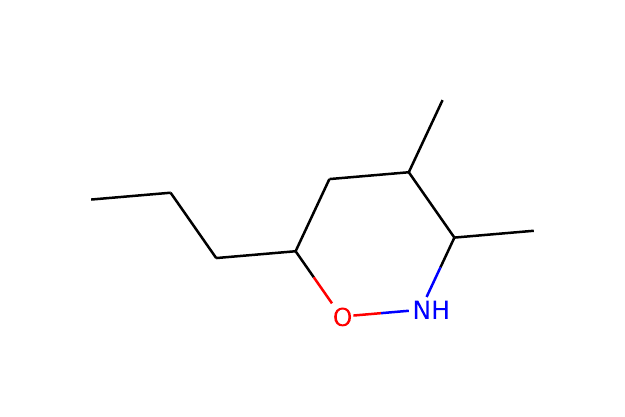}\hspace*{-0.3cm}
			\\
			\includegraphics[width=0.22\textwidth]{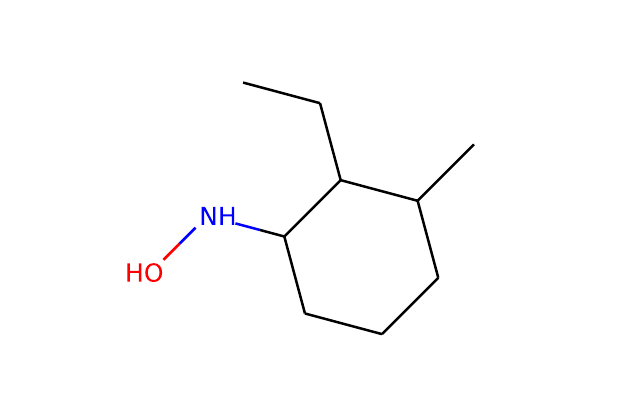}\hspace*{-0.8cm}&
			\includegraphics[width=0.22\textwidth]{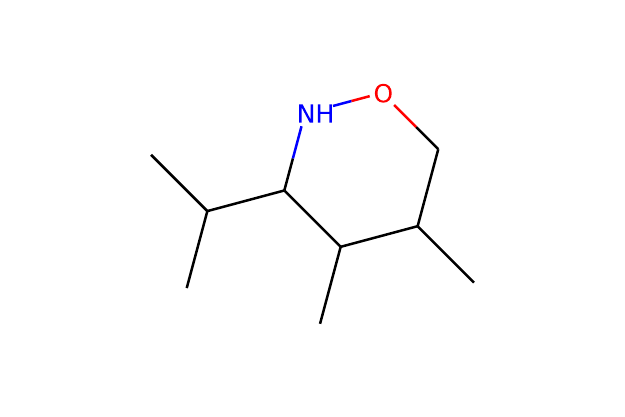}\hspace*{-0.8cm}&
			\includegraphics[width=0.22\textwidth]{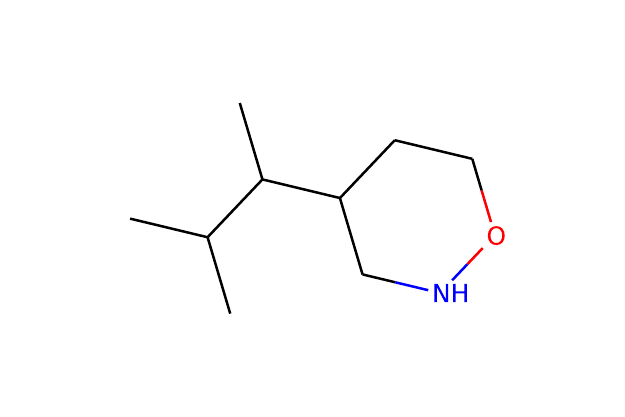}\hspace*{-0.8cm}&
			\includegraphics[width=0.22\textwidth]{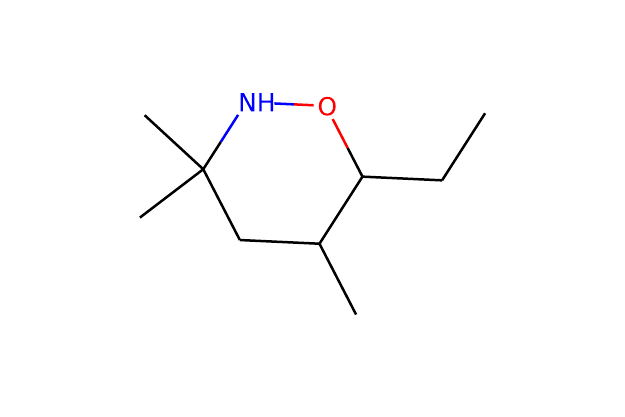}\hspace*{-0.8cm} &
			\includegraphics[width=0.22\textwidth]{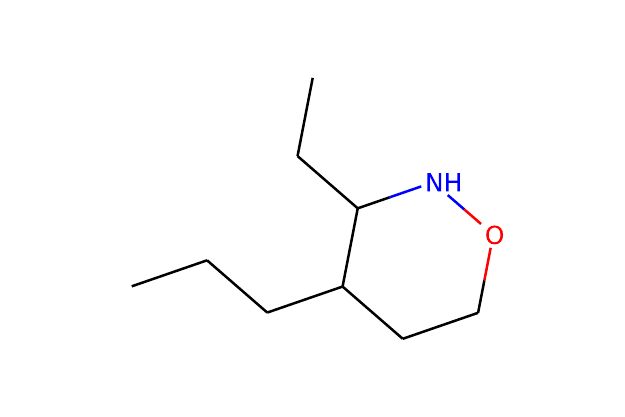}\hspace*{-0.8cm} 
			\\ 
			\includegraphics[width=0.22\textwidth]{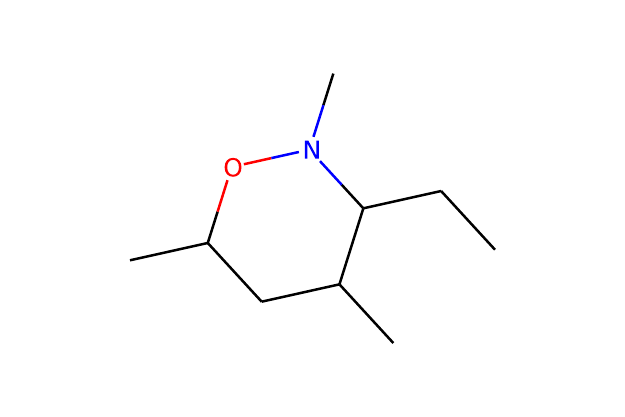}\hspace*{-0.8cm}&
			\includegraphics[width=0.22\textwidth]{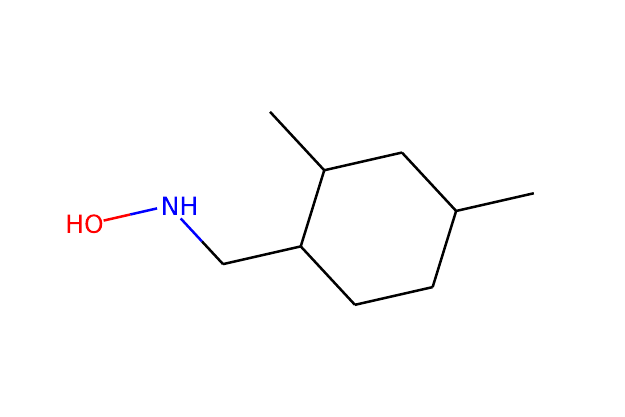}\hspace*{-0.8cm}& 
		        {\includegraphics[width=0.22\textwidth]{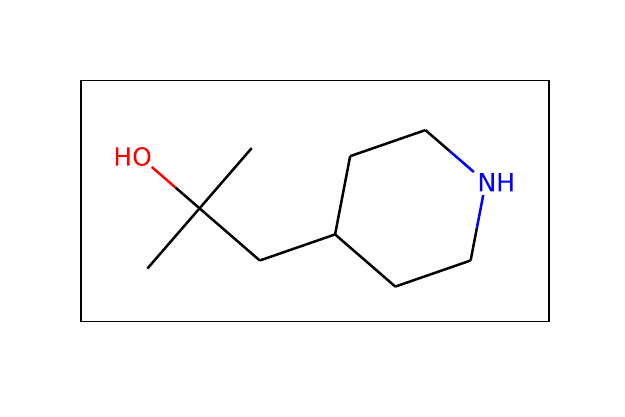} } \hspace*{-0.8cm}&	
			\includegraphics[width=0.22\textwidth]{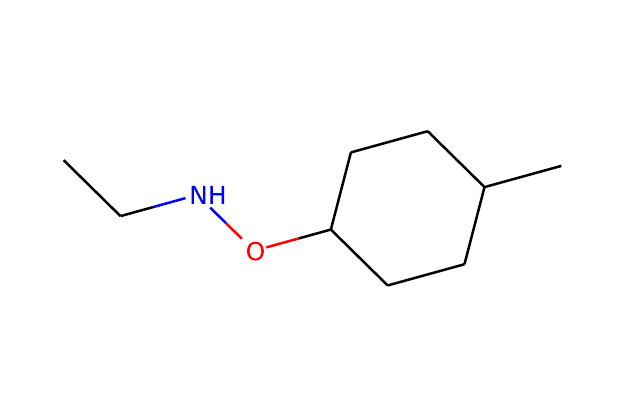}\hspace*{-0.8cm}&
			\includegraphics[width=0.22\textwidth]{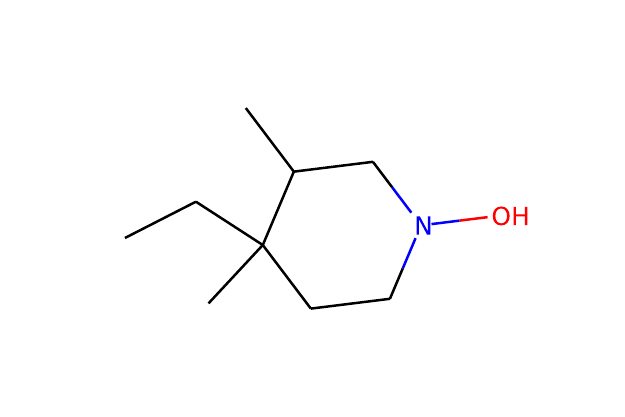}\hspace*{-0.8cm}
			\\
			\includegraphics[width=0.22\textwidth]{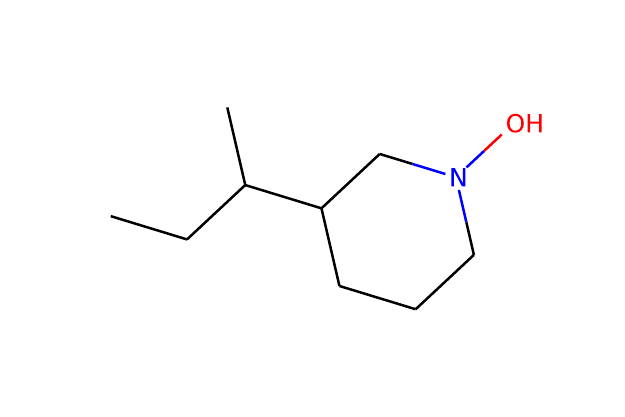}\hspace*{-0.8cm}&
			\includegraphics[width=0.22\textwidth]{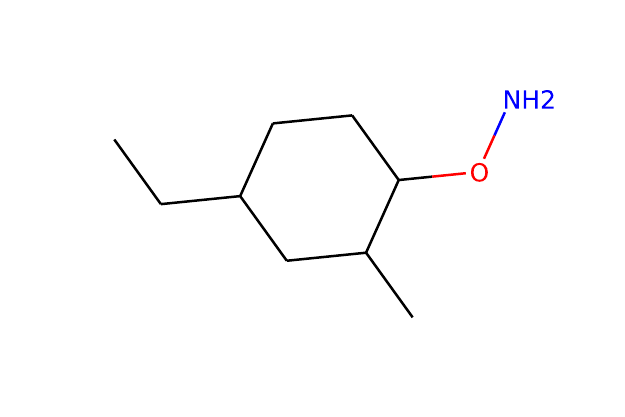}\hspace*{-0.8cm}&
			\includegraphics[width=0.22\textwidth]{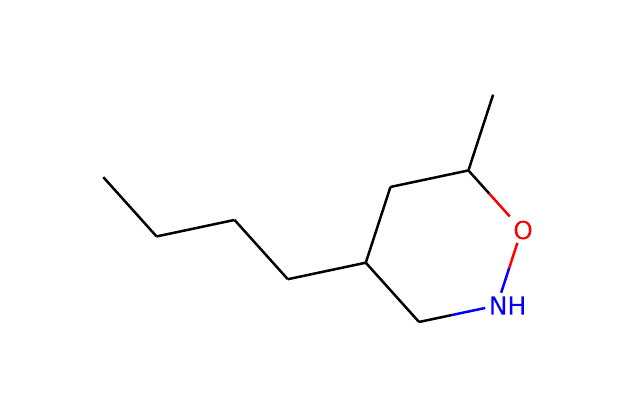}\hspace*{-0.8cm}&
			\includegraphics[width=0.22\textwidth]{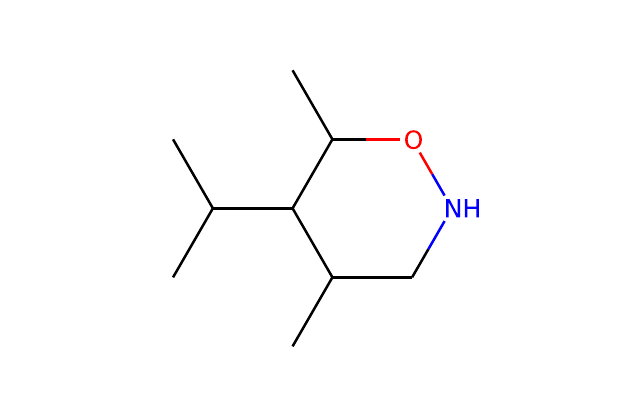}\hspace*{-0.8cm} &
			\includegraphics[width=0.22\textwidth]{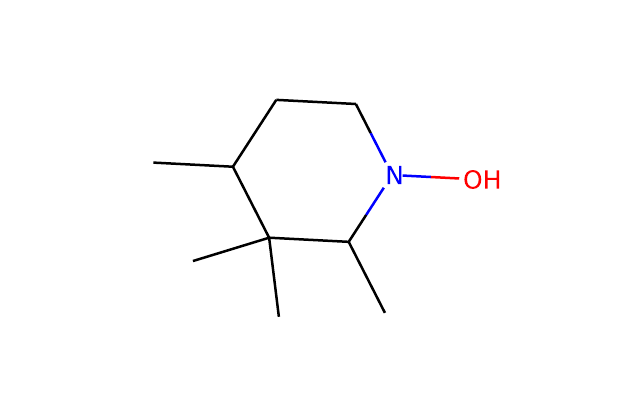}\hspace*{-0.8cm}
			\\
			\includegraphics[width=0.22\textwidth]{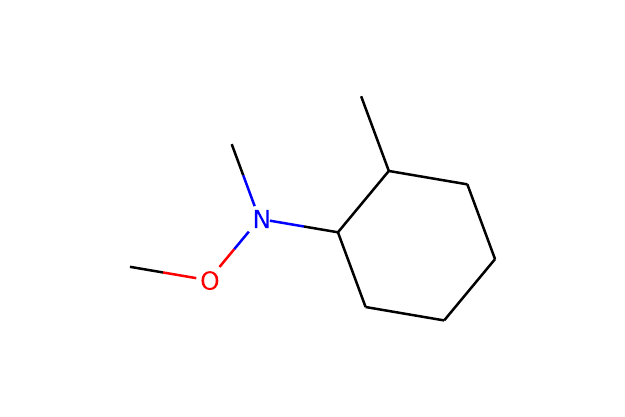}\hspace*{-0.8cm}&
			\includegraphics[width=0.22\textwidth]{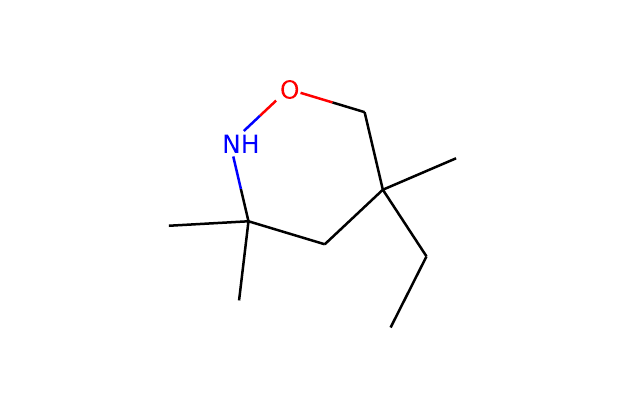}\hspace*{-0.8cm} &
			\includegraphics[width=0.22\textwidth]{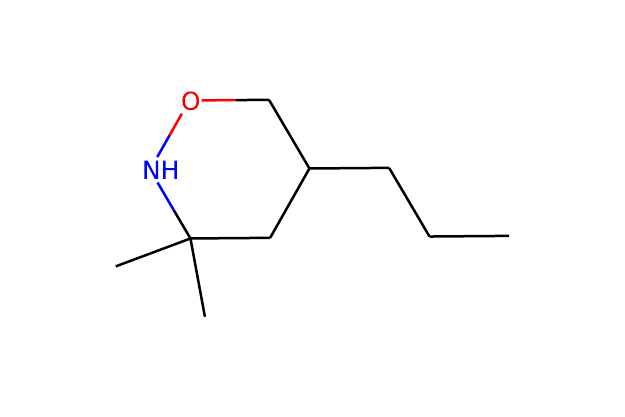}\hspace*{-0.8cm}&
			\includegraphics[width=0.22\textwidth]{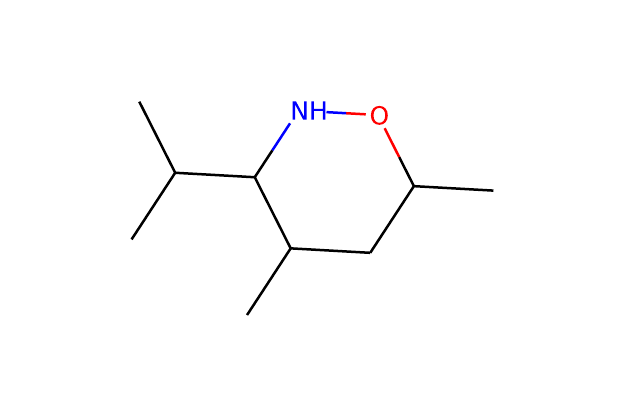}\hspace*{-0.8cm}&
			\includegraphics[width=0.22\textwidth]{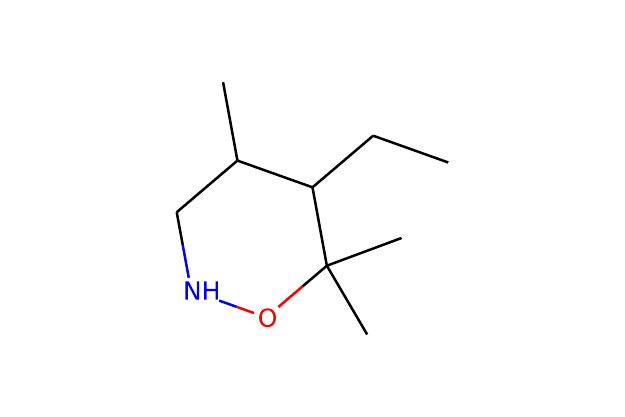}\hspace*{-0.8cm} 
			\\
	\end{tabular}}
%	}
		\caption{Molecules sampled using the probabilistic decoder $\Gcal_i\sim p_\theta (\Gcal|\Zcal)$, where $\Zcal$ is the (sampled) latent representation 
		of a given molecule $\Gcal$ (boxed) from the ZINC dataset.
%	VICEN: package changes conflicts with \boxed
%		of a given molecule $\mbox{\Gcal}$ from the ZINC dataset.
%
		The sampled molecules are topologically similar to each other as well the original. This provides qualitative evidence for the smooth latent space of 
		molecules provided by NeVAE.}
	\label{fig:posterior-zinc}
	\vspace*{-4mm}
\end{figure}

\begin{figure*}[!t]
	\centering
	\subfloat[ZINC dataset]{
	 {\scalebox{0.45}{\begin{tabular}{cccccc}	
	 			
	 		{\includegraphics[width=0.2\textwidth]{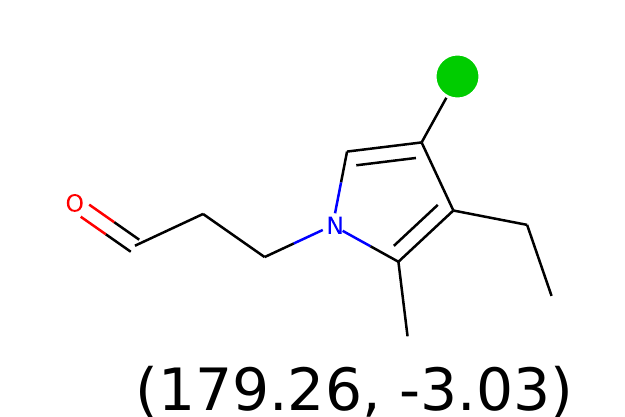}} &
	 		\includegraphics[width=0.2\textwidth]{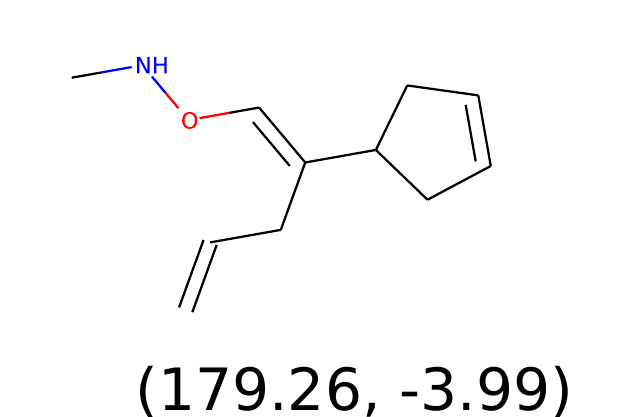}\hspace*{-0.3cm} &
	 		\includegraphics[width=0.2\textwidth]{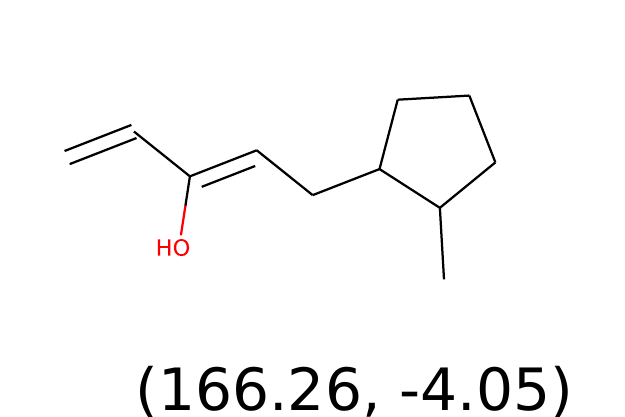}\hspace*{-0.3cm} &
	 		\includegraphics[width=0.2\textwidth]{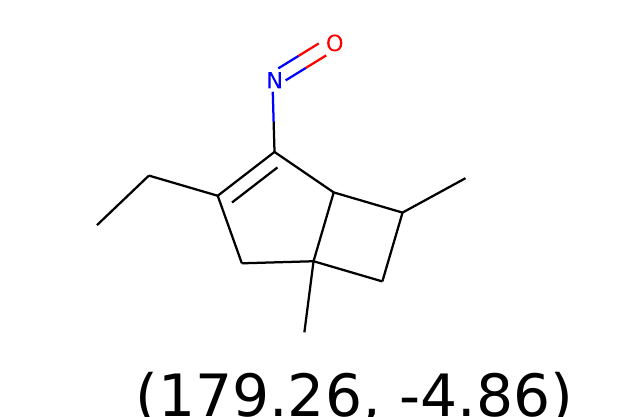}\hspace*{-0.3cm}&	
	 		\includegraphics[width=0.2\textwidth]{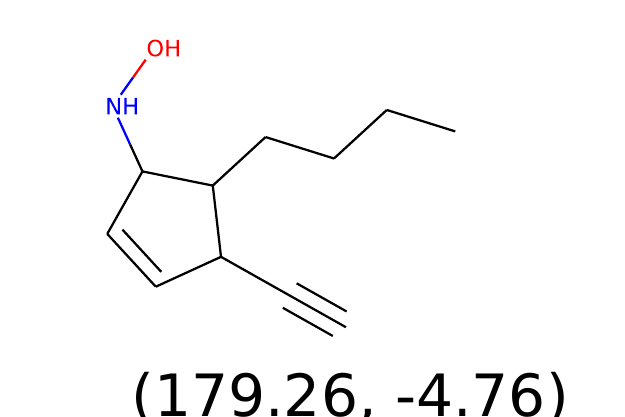}\hspace*{-0.3cm} 
			\\
		{\includegraphics[width=0.2\textwidth]{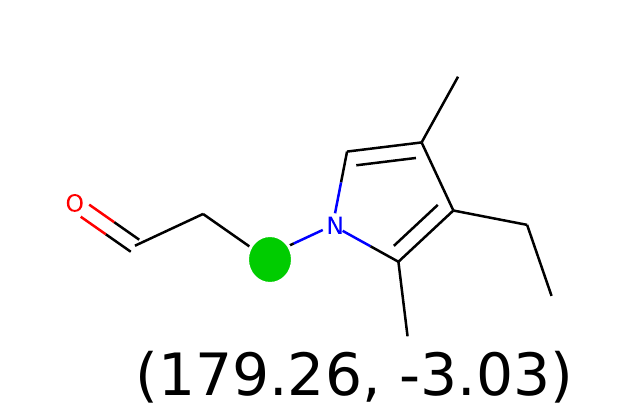}}\hspace*{-0.3cm}&
		\includegraphics[width=0.2\textwidth]{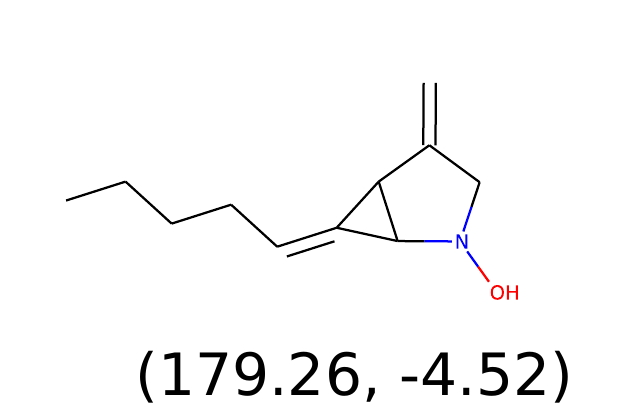}\hspace*{-0.3cm}&
		\includegraphics[width=0.2\textwidth]{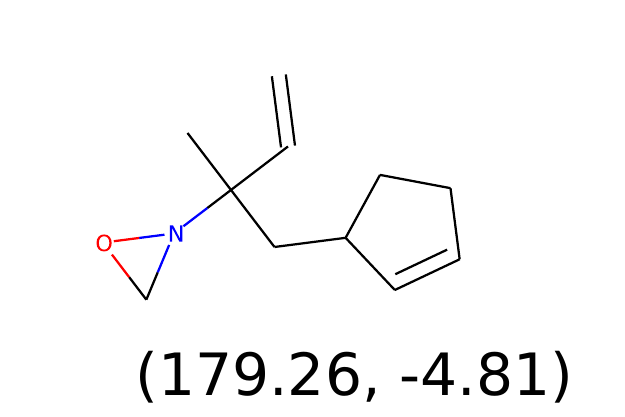}\hspace*{-0.3cm}&
		\includegraphics[width=0.2\textwidth]{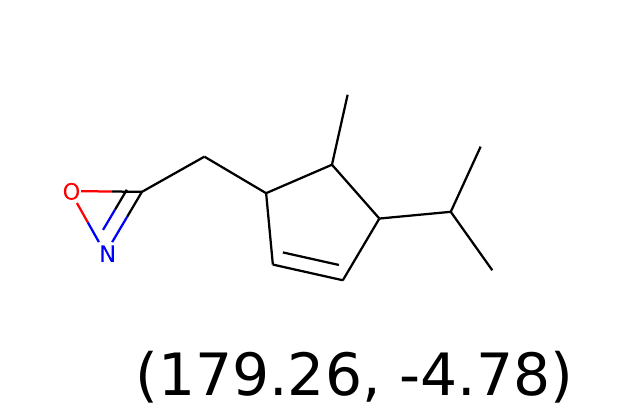}\hspace*{-0.3cm} &
		\includegraphics[width=0.2\textwidth]{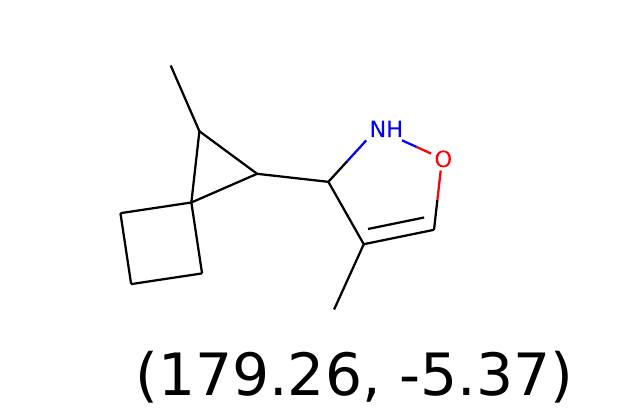}\hspace*{-0.3cm}
		\\
		{\includegraphics[width=0.2\textwidth]{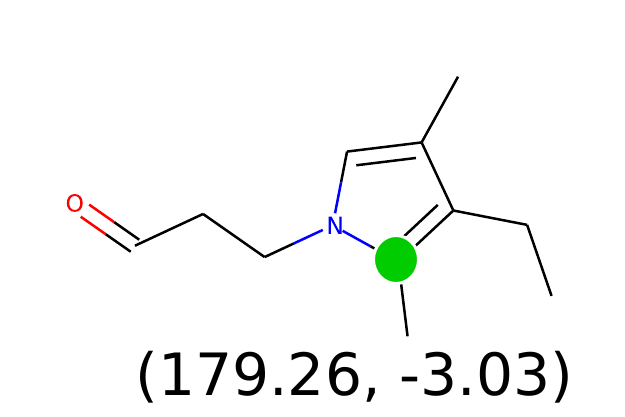}} \hspace*{-0.3cm}&
		\includegraphics[width=0.2\textwidth]{FIG/ZINCDOUBLE/approach_1_node_1_3828_new.pdf}\hspace*{-0.3cm}&
		\includegraphics[width=0.2\textwidth]{FIG/ZINCDOUBLE/approach_1_node_1_228_new.pdf}\hspace*{-0.3cm}&
			\includegraphics[width=0.2\textwidth]{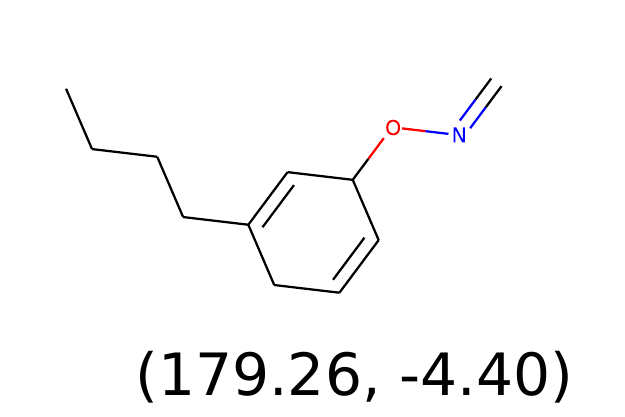}\hspace*{-0.3cm}&
		\includegraphics[width=0.2\textwidth]{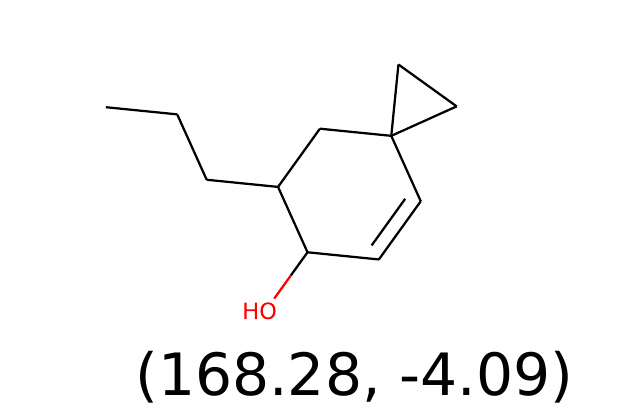}\hspace*{-0.3cm}
	\\
   {\includegraphics[width=0.2\textwidth]{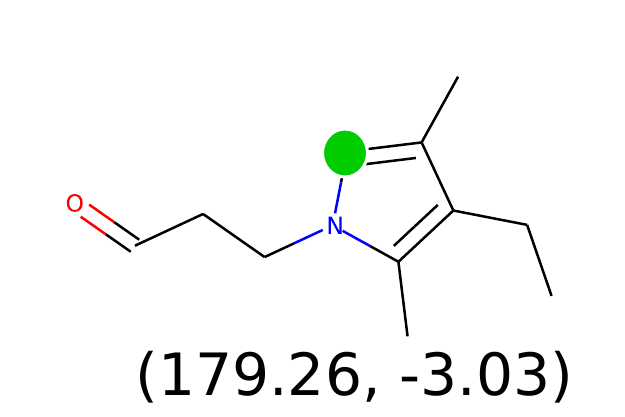}}\hspace*{-0.3cm} &
   \includegraphics[width=0.2\textwidth]{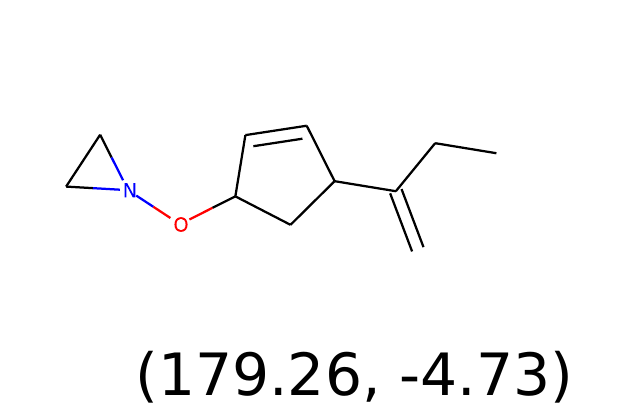}\hspace*{-0.3cm} &		
   \includegraphics[width=0.2\textwidth]{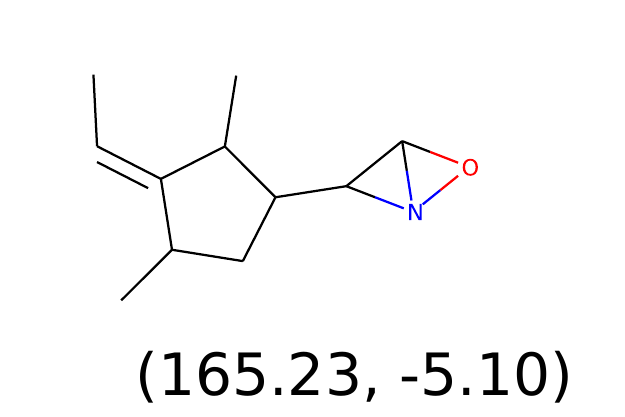}\hspace*{-0.3cm} & 
   \includegraphics[width=0.2\textwidth]{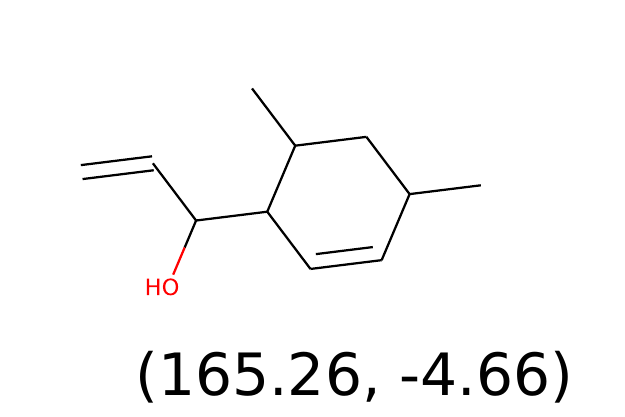}\hspace*{-0.3cm} &
   \includegraphics[width=0.2\textwidth]{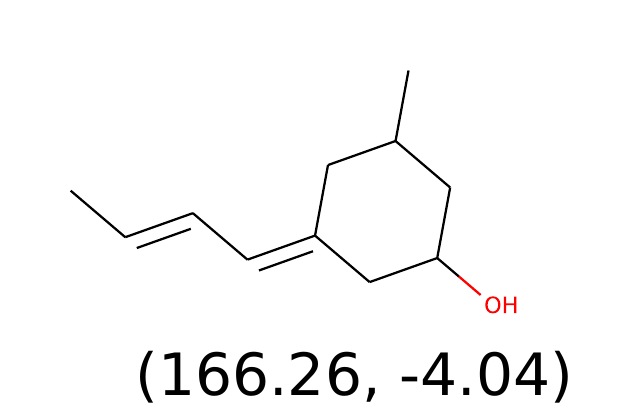}\hspace*{-0.3cm} 
   \\
{\includegraphics[width=0.2\textwidth]{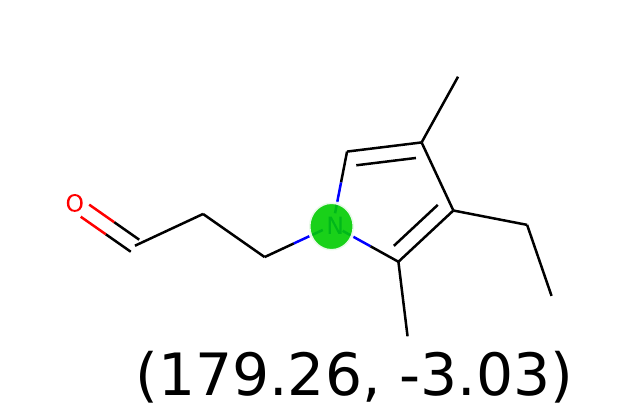}} \hspace*{-0.3cm}&
\includegraphics[width=0.2\textwidth]{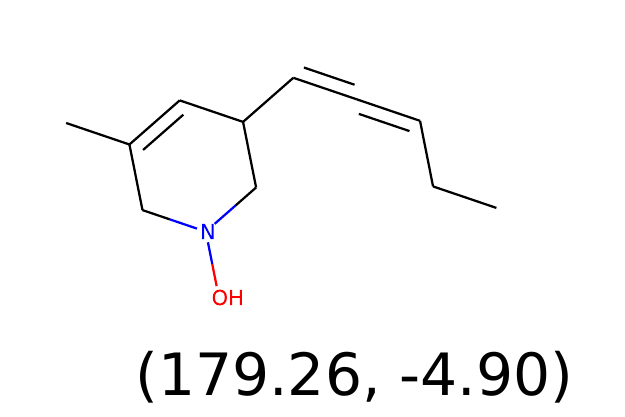}\hspace*{-0.3cm}&
\includegraphics[width=0.2\textwidth]{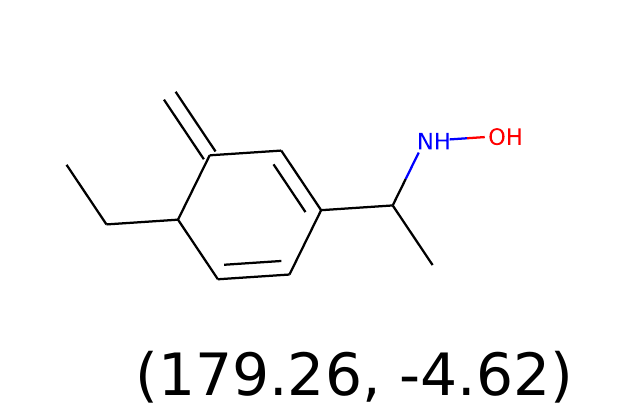}\hspace*{-0.3cm}&
\includegraphics[width=0.2\textwidth]{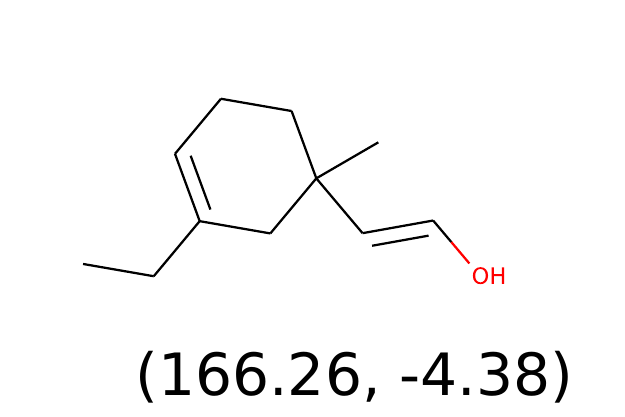}\hspace*{-0.3cm}&
\includegraphics[width=0.2\textwidth]{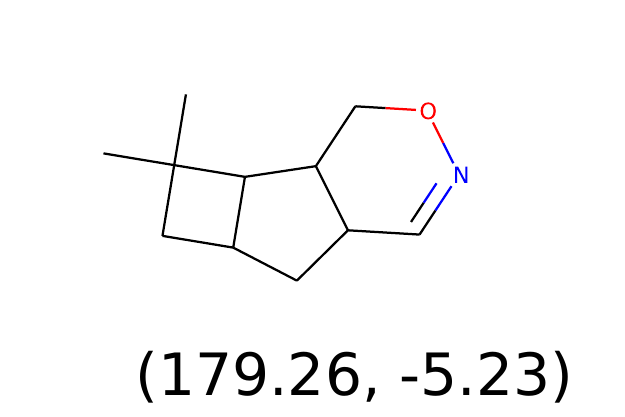}\hspace*{-0.3cm}\\
	\end{tabular}}}
	}
\hspace*{0.7cm}	\subfloat[QM9 Dataset]{\scalebox{0.45}{\hspace*{-1cm} \begin{tabular}{cccccc}
	      {\includegraphics[width=0.2\textwidth]{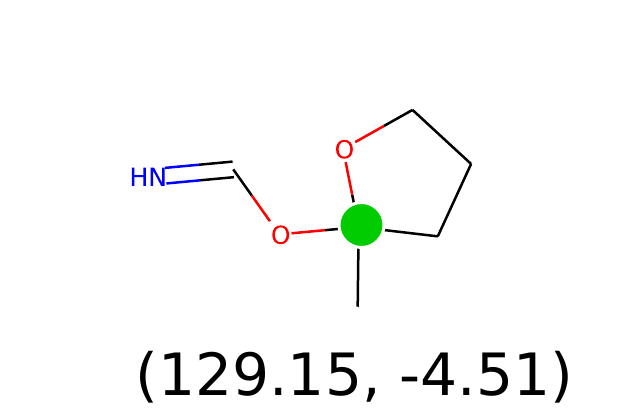}}\hspace*{-0.3cm}&
		\includegraphics[width=0.2\textwidth]{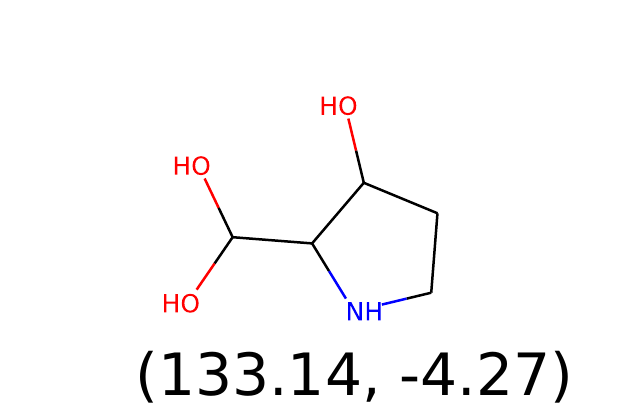}\hspace*{-0.3cm}&
	        \includegraphics[width=0.2\textwidth]{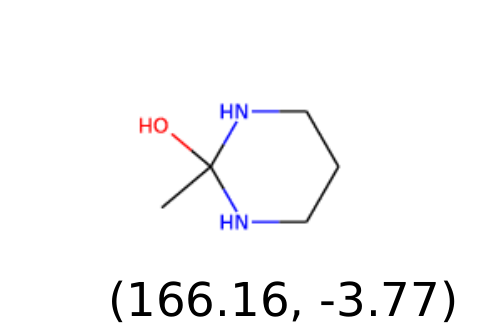}\hspace*{-0.3cm}&
		\includegraphics[width=0.2\textwidth]{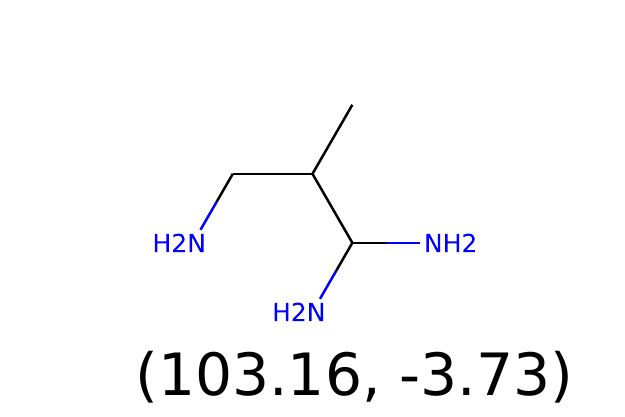}\hspace*{-0.3cm}&
		\includegraphics[width=0.2\textwidth]{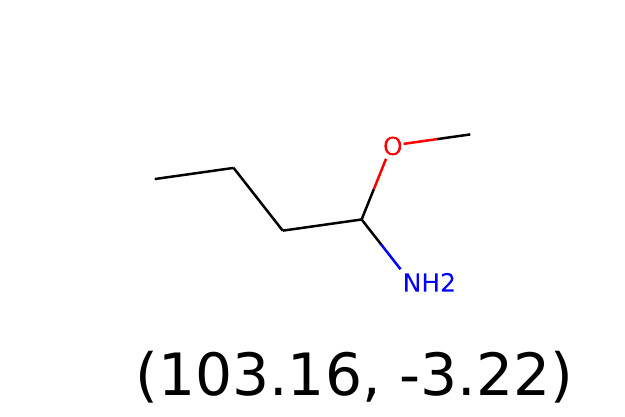}\hspace*{-0.3cm}
		\\
		{\includegraphics[width=0.2\textwidth]{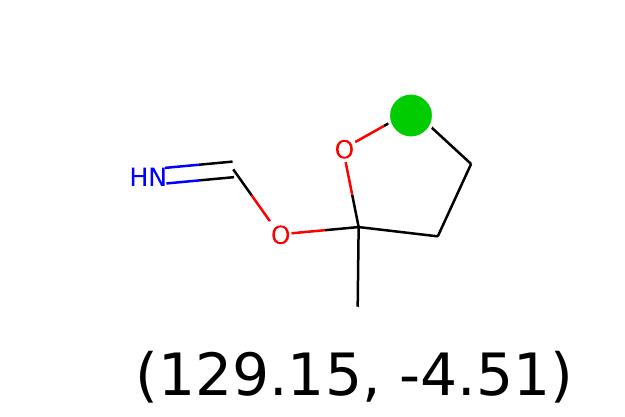}}\hspace*{-0.3cm} &
		\includegraphics[width=0.2\textwidth]{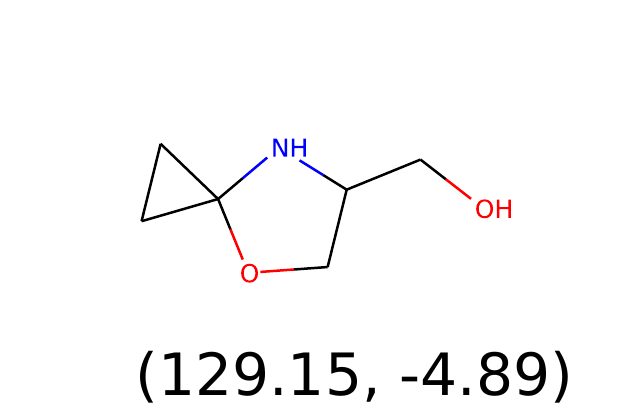}\hspace*{-0.3cm}&
		\includegraphics[width=0.2\textwidth]{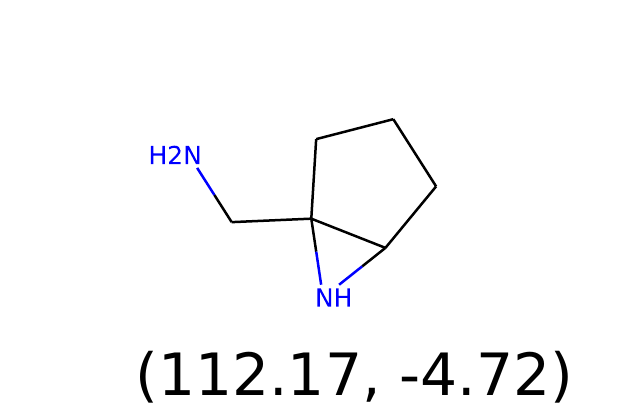}\hspace*{-0.3cm}&
		\includegraphics[width=0.2\textwidth]{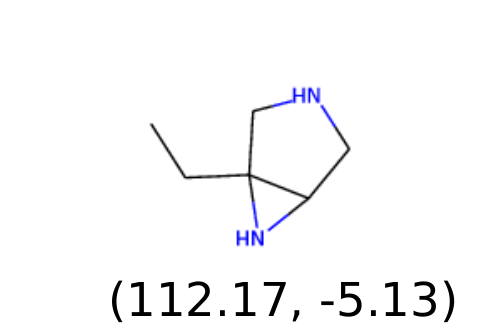}\hspace*{-0.3cm}&
		\includegraphics[width=0.2\textwidth]{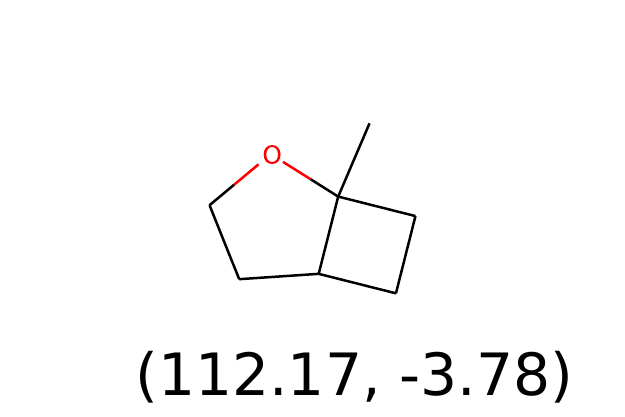}\hspace*{-0.3cm}
		\\
		{\includegraphics[width=0.2\textwidth]{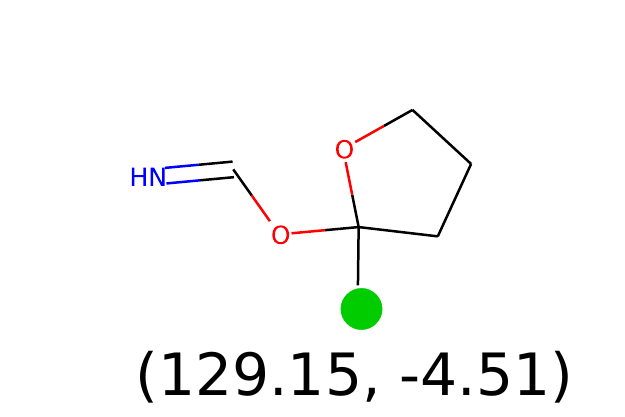}}\hspace*{-0.3cm}&
		\includegraphics[width=0.2\textwidth]{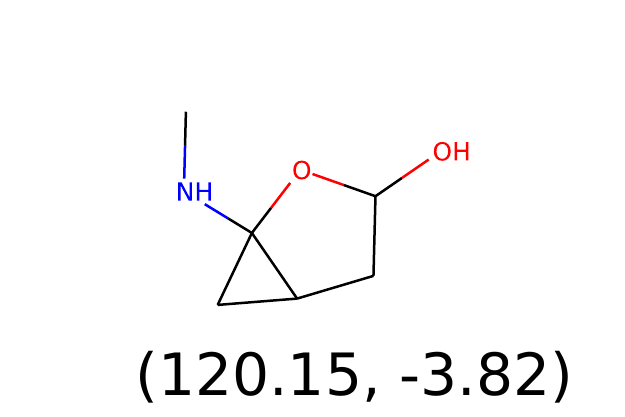}\hspace*{-0.3cm}&
                \includegraphics[width=0.2\textwidth]{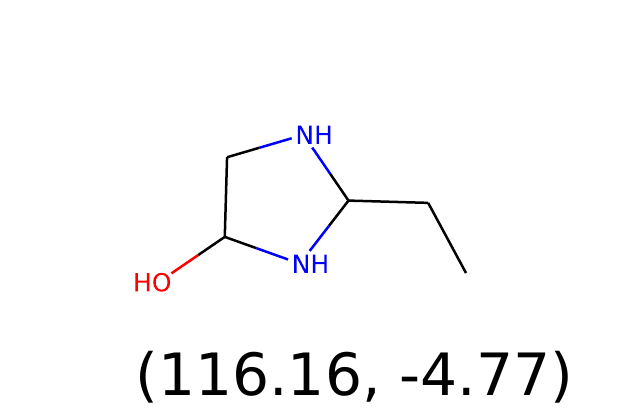}\hspace*{-0.3cm}&
                \includegraphics[width=0.2\textwidth]{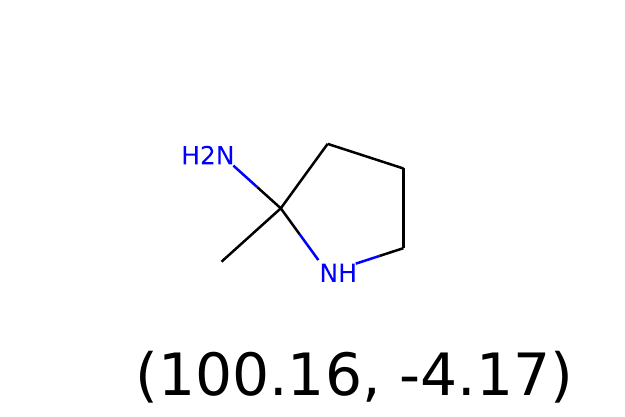}\hspace*{-0.3cm} &
                \includegraphics[width=0.2\textwidth]{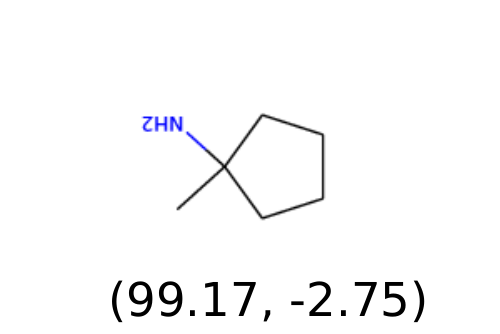}\hspace*{-0.3cm}
		\\ 
	         {\includegraphics[width=0.2\textwidth]{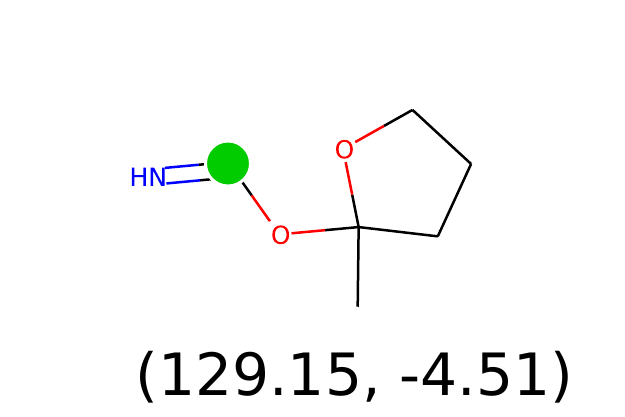}}\hspace*{-0.3cm}&
	         \includegraphics[width=0.2\textwidth]{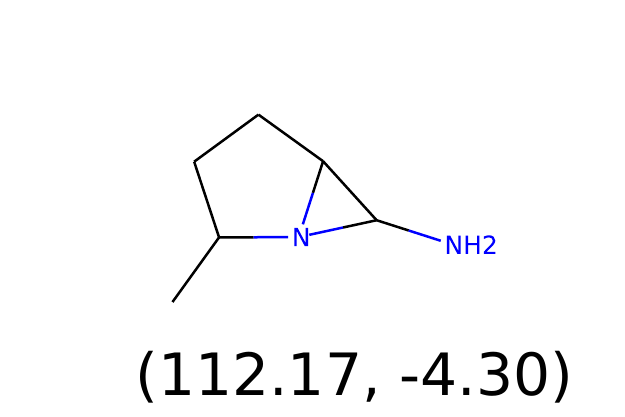}\hspace*{-0.3cm} &
                 \includegraphics[width=0.2\textwidth]{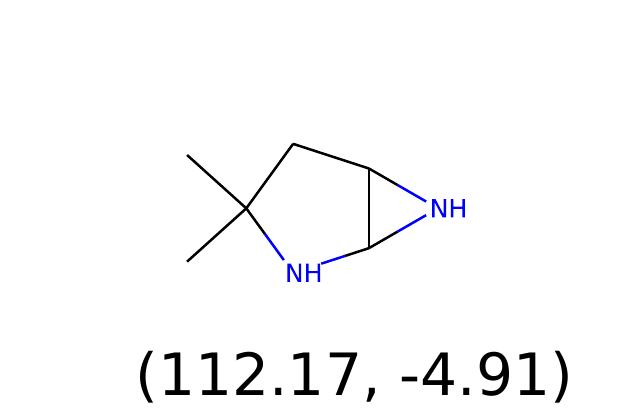}\hspace*{-0.3cm}&
		  \includegraphics[width=0.2\textwidth]{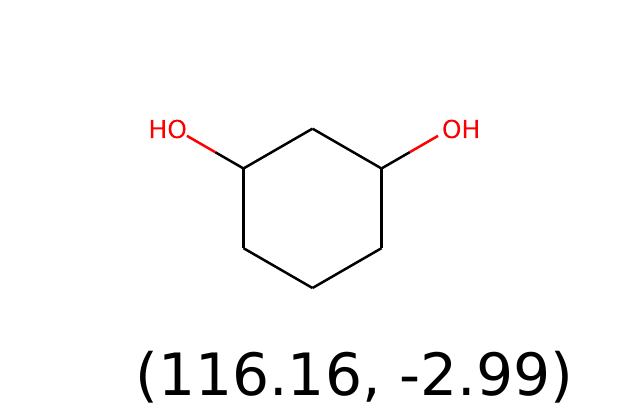}\hspace*{-0.3cm}&
		\includegraphics[width=0.2\textwidth]{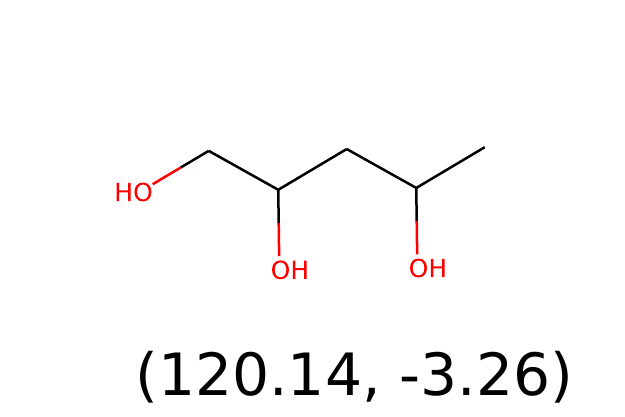}\hspace*{-0.3cm}
		\\
		{\includegraphics[width=0.2\textwidth]{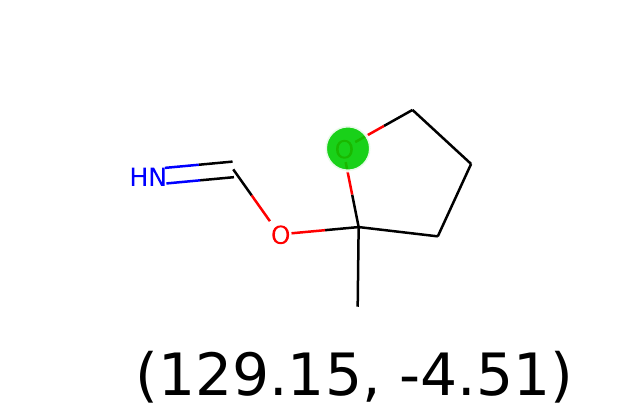}}\hspace*{-0.3cm}&
		\includegraphics[width=0.2\textwidth]{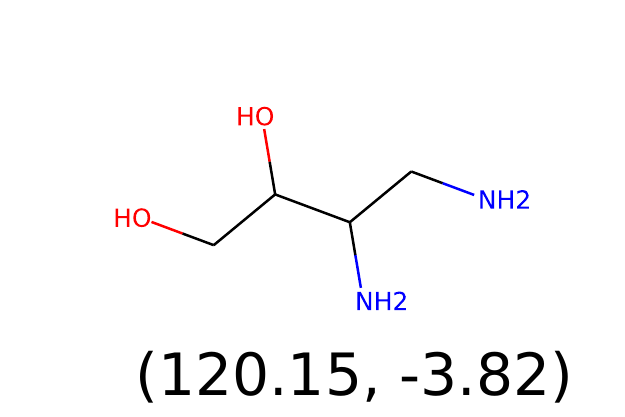}\hspace*{-0.3cm} &
		\includegraphics[width=0.2\textwidth]{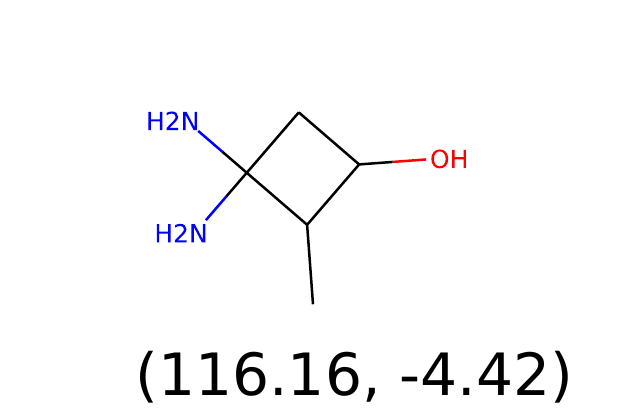}\hspace*{-0.3cm}&
		\includegraphics[width=0.2\textwidth]{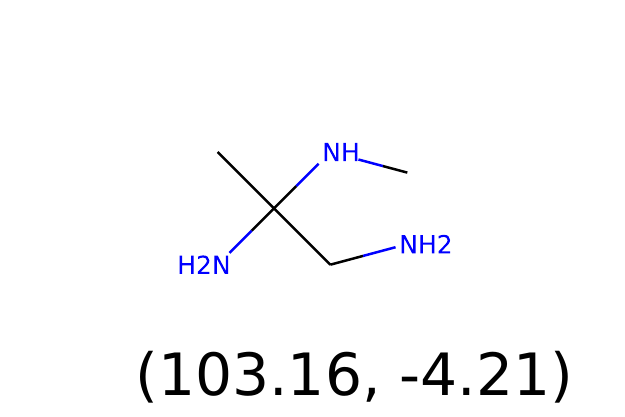}\hspace*{-0.3cm}&
		\includegraphics[width=0.2\textwidth]{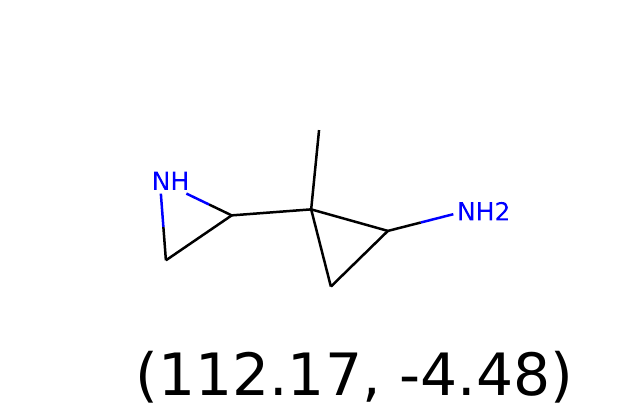}\hspace*{-0.3cm}
		\\
		\end{tabular}}}
	%}
	\vspace{-2mm}
	\caption{ 
	Molecules sampled using the probabilistic decoder $\Gcal \sim p_{\theta}(\Gcal | \Zcal)$, where $\Zcal = \{ \zb_i + a_i \zb_i | \, \zb_i \in \Zcal_0, a_i \geq 0\}$ and $a_i$ are given parameters.
	In each row, we start from the same molecule, set $a_i > 0$ for a single arbitrary node $i$ (denoted as \ngreen{$\bullet$}) and set $a_j = 0, j \neq i$ for the 
         remaining nodes. Under each molecule we report molecular weight and synthetic accessibility score.
	}
	\label{fig:interp-real-qm9}
\end{figure*} 

Next, we qualitatively demonstrate that the latent space of molecules inferred by our model is smooth. To that aim, given a molecule, along with its associated graph $\Gcal$, node 
features $\Fcal$ and edge weights $\Ycal$, we first sample its latent representation $\Zcal$ using our probabilistic encoder, \ie, $\Zcal \sim q_\phi (\Zcal|G,\Fcal,\Ycal)$. 
Then, given this latent representation, we generate various molecular graphs by sampling from our probabilistic decoder, \ie, $\Gcal_i\sim p_\theta (\Gcal|\Zcal)$. 
Figure~\ref{fig:posterior-zinc} summarizes the results for one molecule from  ZINC dataset, which shows that the sampled molecules are topologically similar to the given 
molecule. Finally, we also show that our encoder, once trained, creates a latent space representation of molecules with powerful semantics. In particular, 
since each node in a molecule has a latent representation, we can make fine-grained changes to the structure of a molecule by perturbing the latent
representation of single nodes.
To this aim, we proceed as follows.

First, we select one molecule with $n$ nodes from the ZINC dataset. Given its corresponding graph, node features and edge weights, $\Gcal$, $\Fcal$ and $\Ycal$, we sample its 
latent representation $\Zcal_0$. 
Then, we sample new molecular graphs $\Gcal$ from the pro\-ba\-bi\-lis\-tic decoder % latent values $\Zcal$, by smoothly perturbing the latent representation of one of the nodes. 
% in between these latent representations using a fined-grained node level linear interpolation
% Specifically we sample 
$\Gcal \sim p_{\theta}(\Gcal | \Zcal)$, where $\Zcal = \{ \zb_i + a_i \zb_i | \, \zb_i \in \Zcal_0, a_i \geq 0\}$ and $a_i$ are given parameters.
% and the node labels, 
% which define the matching between pairs of nodes in both graphs, are arbitrary.
%
Figure~\ref{fig:interp-real-qm9} provides several examples across both datasets, which show that the latent space representation is smooth and, as the distance from the initial molecule increases 
in the latent space, the resulting molecule differs more from the original.

\subsection{Property oriented molecule generation}
In this section, we first use our gradient-based algorithm (refer to Algorithm~\ref{alg:prop}) to design property-oriented decoders that maximize the following two 
properties:
\begin{itemize}[noitemsep,nolistsep,leftmargin=0.7cm]
\item[(i)] the octanol-water partition coefficient, penalized by synthetic accessibility (SA) score and number of long cycles (penalized logP, $y_1(m)$); 
and, 
\item[(ii)] the  quantitative estimation of drug-likeness (QED, $y_2(m)$).
\end{itemize}
Then, we use our gradient-based algorithm to design property-oriented decoders that, given a mo\-le\-cule of interest, are able to optimize the spatial configuration of 
its atoms for greater stability, \ie, lower potential energy. 

\begin{table}[!t]
	\centering
	\small
	\scalebox{0.8}{
		\begin{tabular}{l|l|l|l|l|l|l|l}
			\hline
			\multicolumn{2}{l|}{Method}                                                          & \multicolumn{3}{l|}{Penalized logP} & \multicolumn{3}{l}{QED} \\ \hline
			& &1st        & 2nd        & 3rd       & 1st    & 2nd    & 3rd    \\ \hline
			%&\begin{tabular}[c]{@{}l@{}}ZINC \\ (Training Data)\end{tabular} & 2.582&  &   &0.758  &  &        \\ \hline
			%\cline{2-8}
			\multirow{4}{3cm}{Bayesian optimisation} & \ourmodel\   & 2.826 & 2.447 & 2.299 & 0.732 &0.705  &0.704 \\ 
			%\hline
			\cline{2-8}
			&GrammarVAE & 2.521  & 2.330 & 1.893 &0.724 &0.712 & 0.698       \\
			% \hline
			\cline{2-8}
			&CVAE &  1.835 & 1.776  & 1.234 & 0.712 & 0.698 & 0.508 \\ 
			%\hline
			\cline{2-8}
			&JTVAE &  3.503& 3.224& 2.960 &  0.848& 0.831 &0.776 \\ \hline
			\multirow{4}{3cm}{Property oriented decoder} &MOLGAN& 0.259 & 0.233 & 0.231 & 0.398 &0.368 &0.344 \\ 
			%\hline
			\cline{2-8}
			%\multirow{3}{3cm}{Property oriented decoder} 
			&ORGAN& 3.148 & 2.334 & 2.145 &0.812 &0.807 &0.745 \\ 
			%\hline
			\cline{2-8}
			%\cline{2-8}
			%\multirow{3}{3cm}{Property oriented decoder} &ORGAN& 3.148 & 2.334 & 2.145 &0.812 &0.807 &0.745 \\ 
			%\hline
			%\cline{2-8}
			&GCPN  & 4.284 & 3.621& 2.902 & 0.913 & 0.885  & 0.775   \\ %\hline
			\cline{2-8}
			&\ourmodel\ (Algorithm~\ref{alg:prop}) & \textbf{6.82} & 6.65 & 6.55 & \textbf{0.920} & 0.916 & 0.912 \\
			\hline
		\end{tabular}
	}
	\caption{Penalized logP and QED scores for the best three molecules generated by our property oriented decoder, \ourmodel\ (Algorithm~\ref{alg:prop}),
	and all baselines. 
	For our property oriented decoder, we used $\rho=10^{-4}$ and $\rho=5\times 10^{-6}$ for penalized logP and QED scores, respectively.}
	\label{tab:propval}
\end{table}

\begin{figure}[!h]
	\centering
	%\vspace*{-1mm}
	
	\resizebox{\hsize}{!}{
		\begin{tabular}{ccc}
			
			%\hspace*{-2cm}
% 			$\rho = 1e^{-4}$ &
			\includegraphics[width=0.4\hsize, clip,height=35mm]{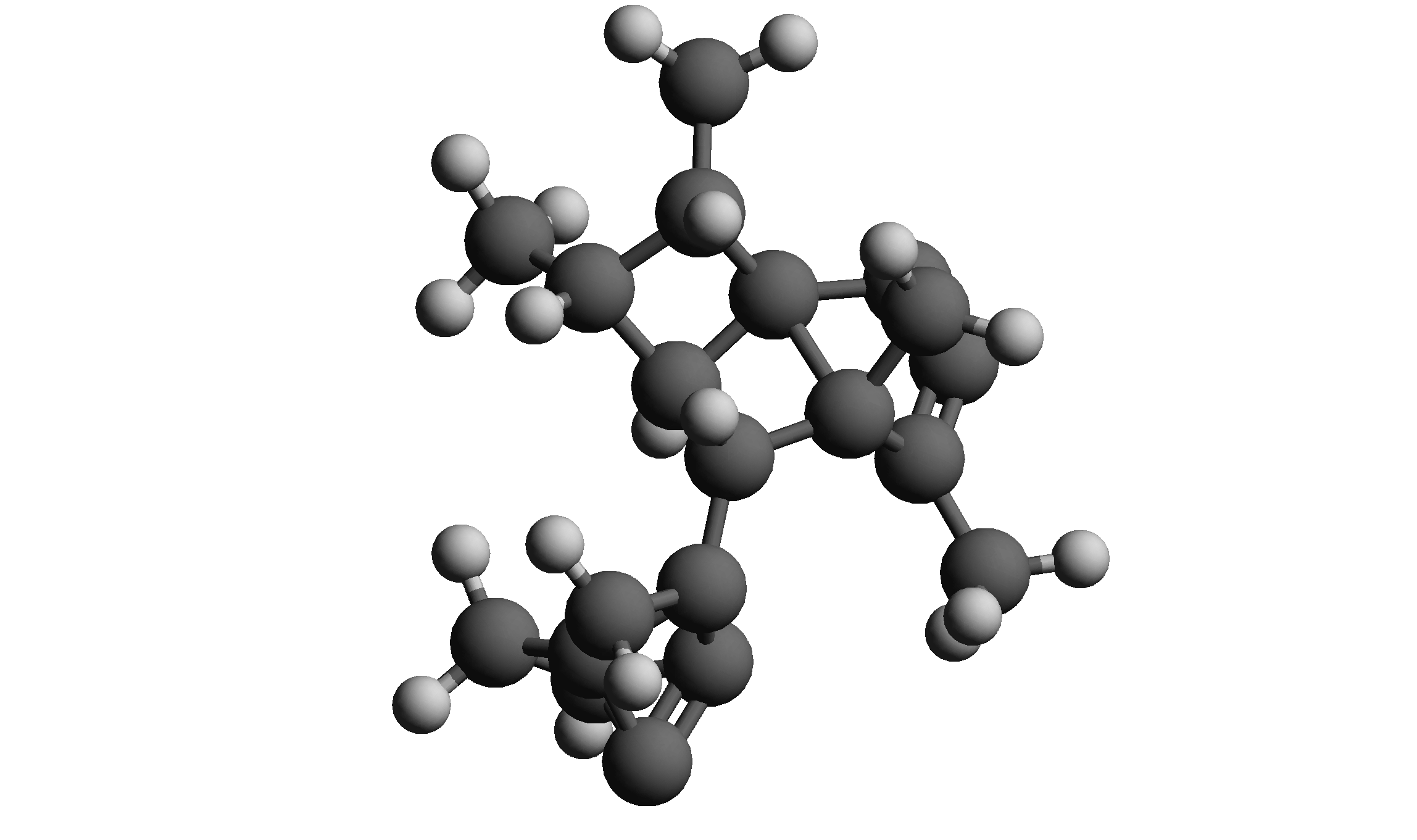} \hspace*{-1cm} &
			\includegraphics[width=0.4\hsize, clip,height=35mm]{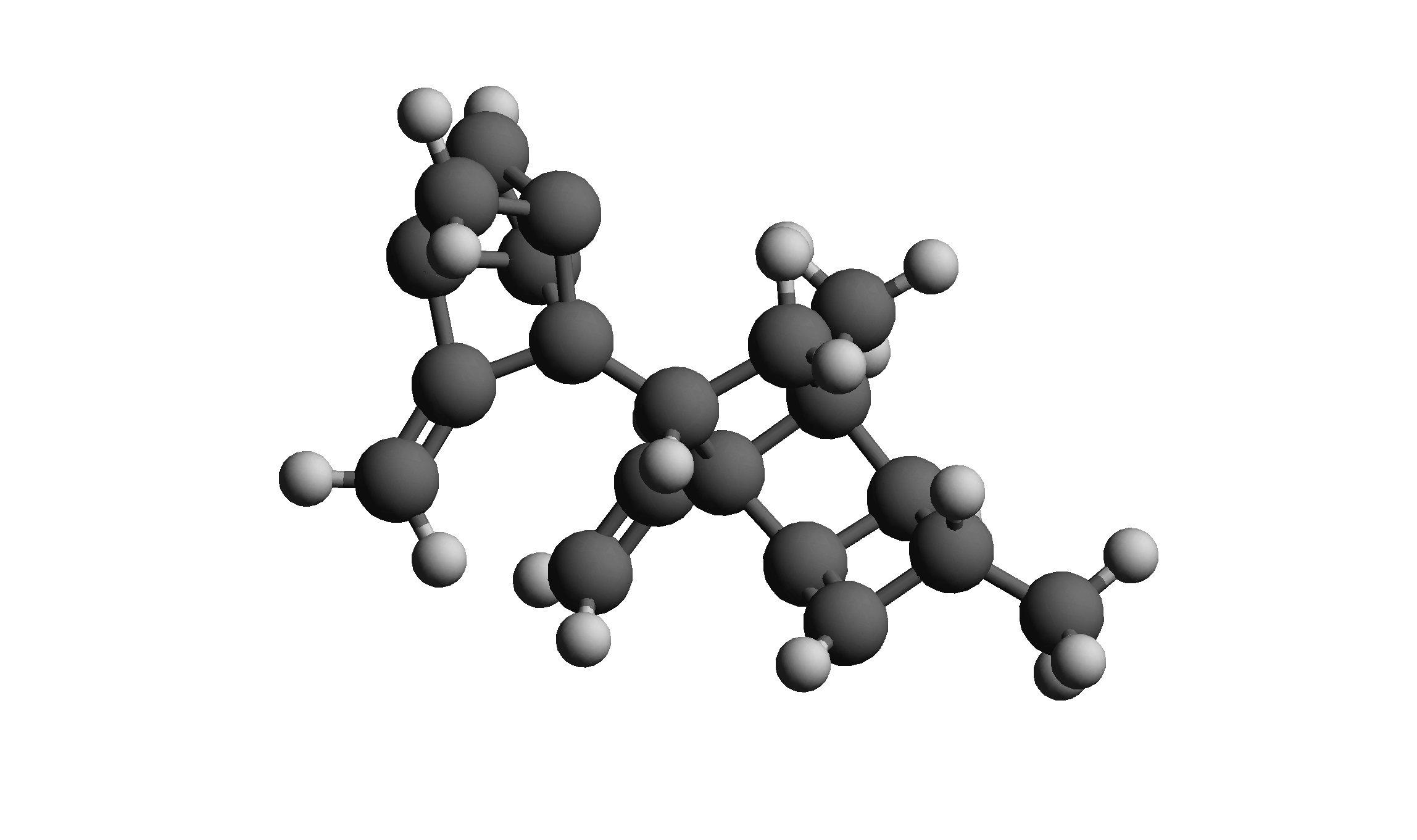} \hspace*{-1cm} &
			\includegraphics[width=0.4\hsize, clip,height=35mm]{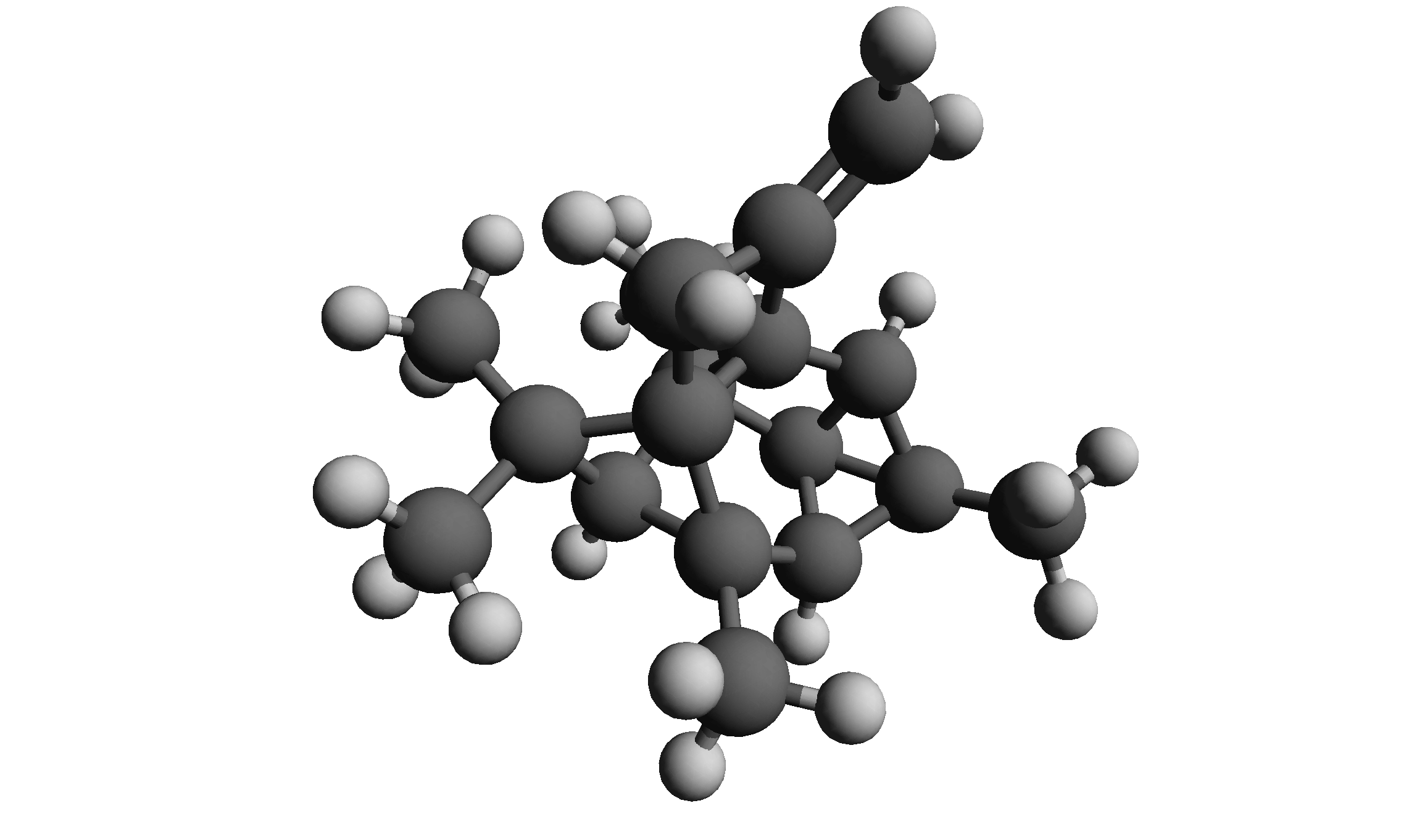} \hspace*{-1cm} 
			
			\vspace{-2mm} \\
			
			%\hspace*{-2cm}
			\includegraphics[width=0.2\textwidth]{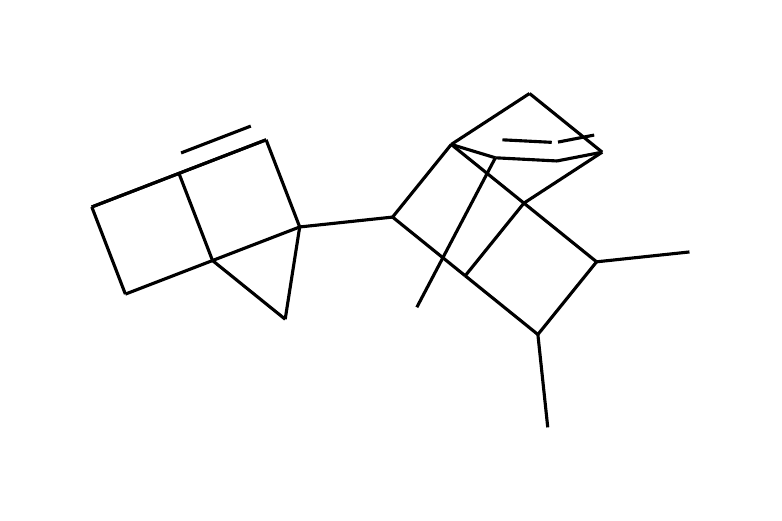}\hspace*{-1cm}&
			\includegraphics[width=0.2\hsize, clip]{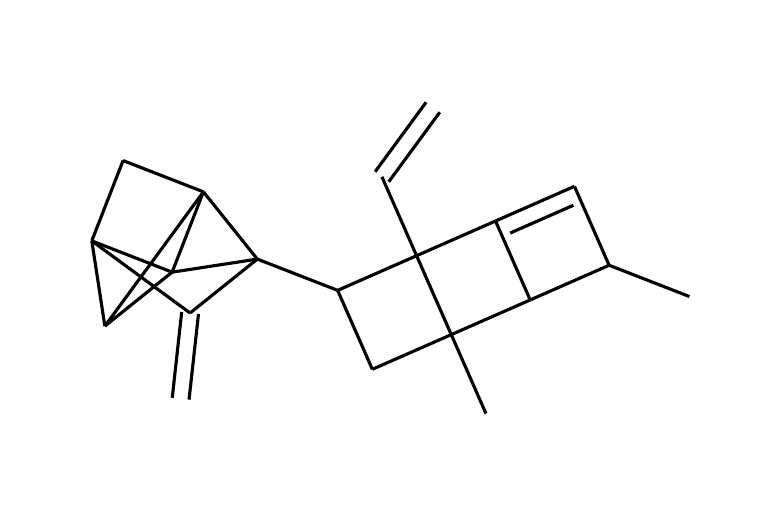} \hspace*{-1cm} &
			\includegraphics[width=0.2\textwidth]{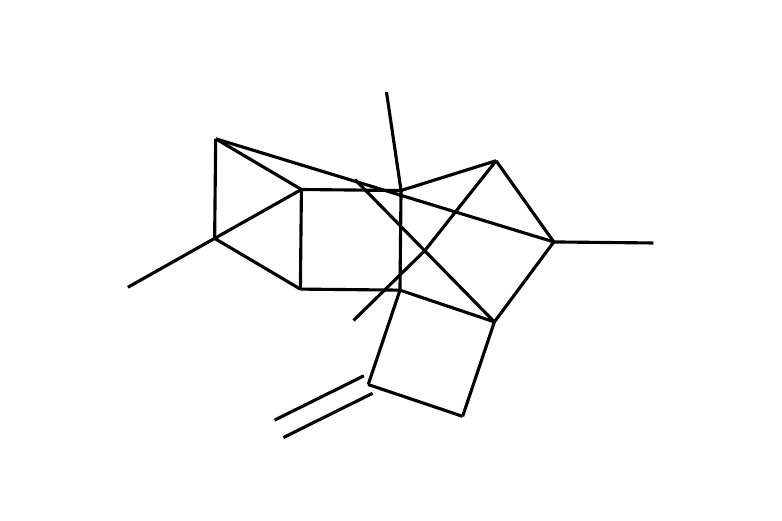}\hspace*{-1cm}
			\vspace{-2mm} \\
			
			%\hspace*{-2cm}
			\scriptsize $y_1(m) = 6.82$ (1st) &\scriptsize  $y_1(m) = 6.65$ (2nd)& \scriptsize  $y_1(m) = 6.55$ (3rd)
		 \\ \\ \\
			\includegraphics[width=0.4\hsize, clip,height=35mm]{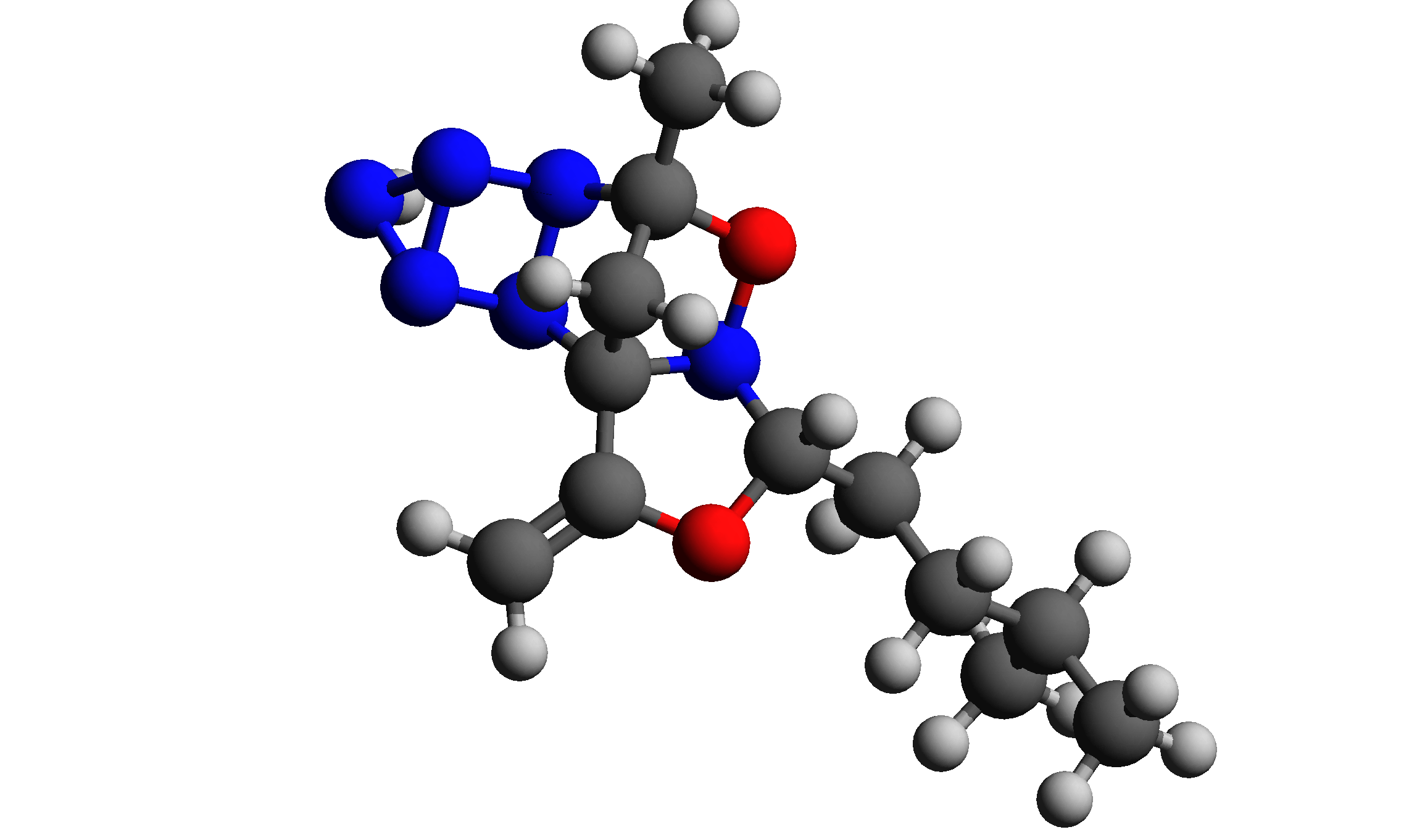} \hspace*{-1cm} &
			\includegraphics[width=0.4\hsize, clip,height=35mm]{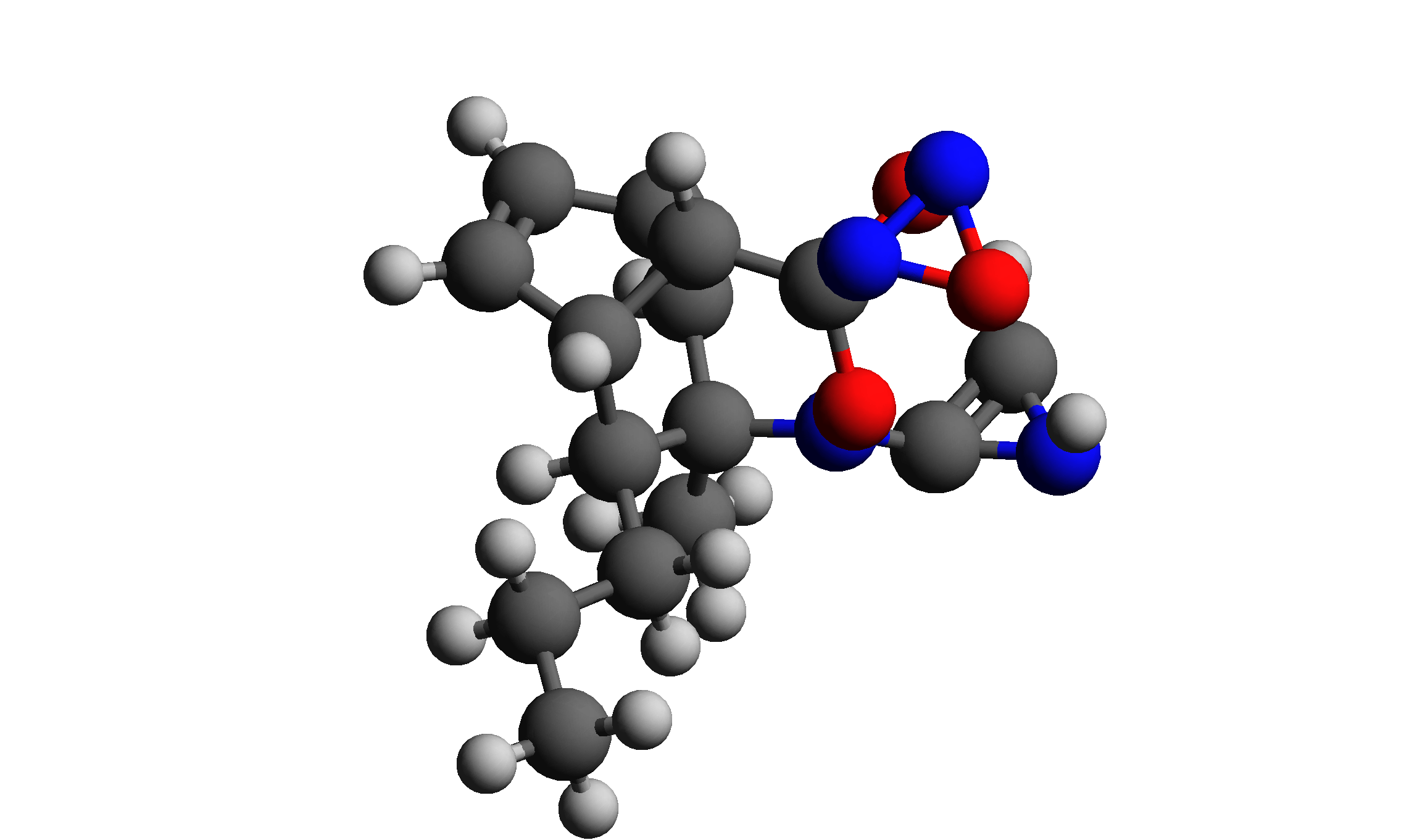} \hspace*{-1cm} &
			\includegraphics[width=0.4\hsize, clip,height=35mm]{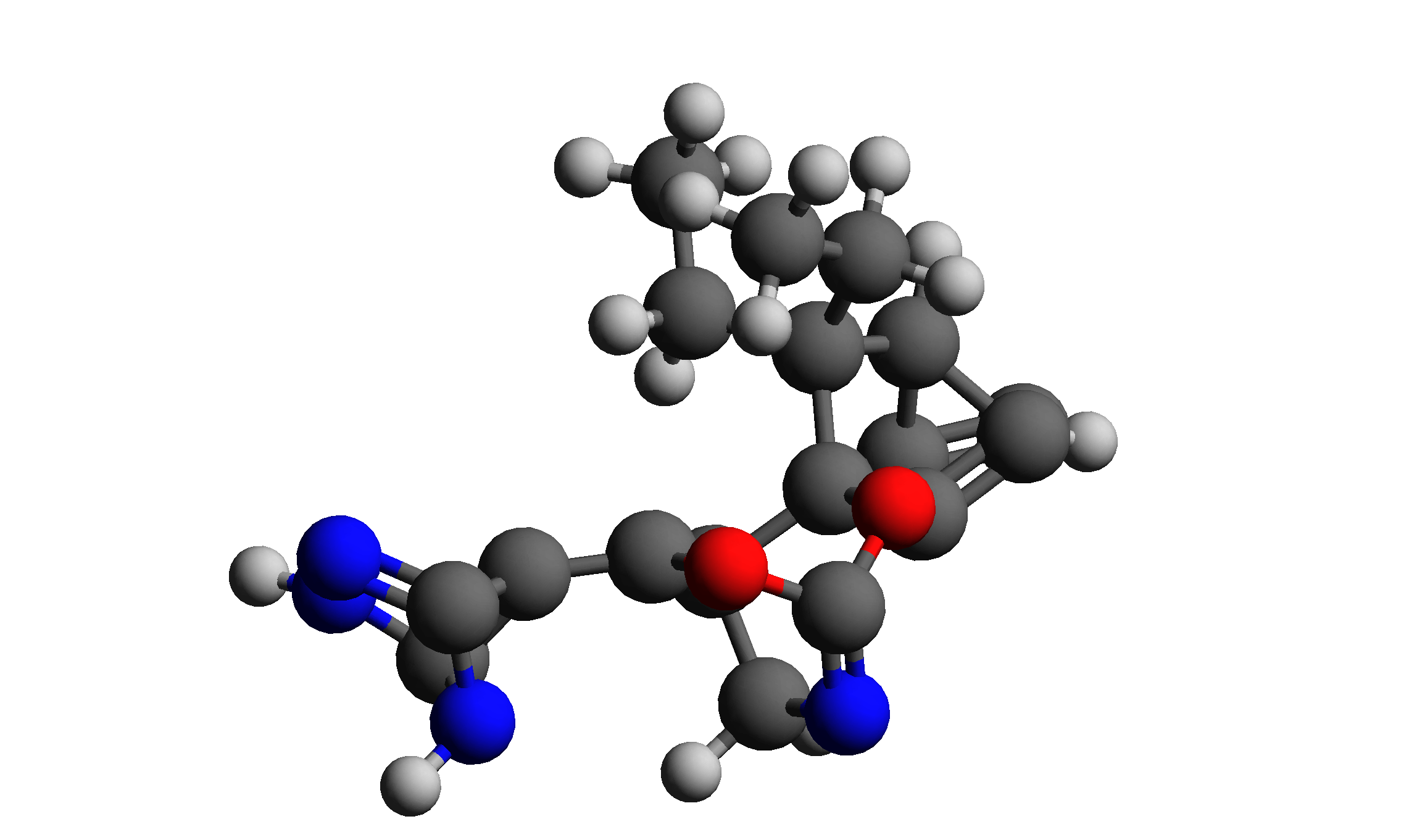} \hspace*{-1cm} 
			\vspace{-2mm} \\
			
			%\hspace*{-2cm}
			\includegraphics[width=0.2\textwidth]{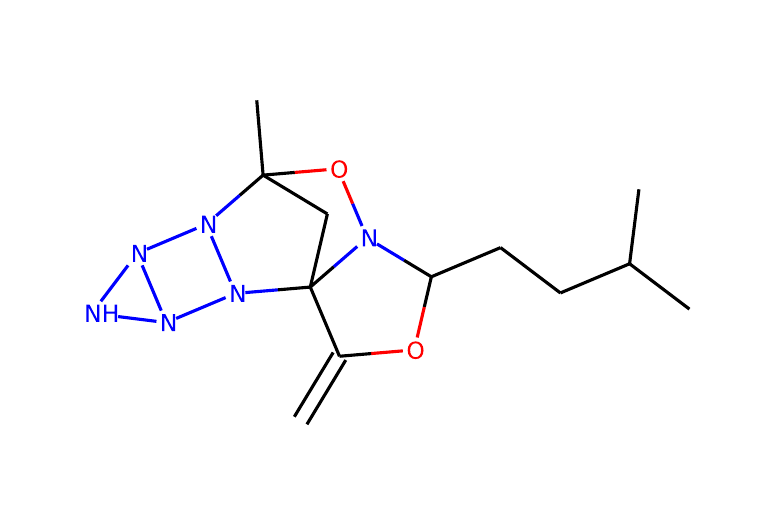}\hspace*{-1cm}&
			\includegraphics[width=0.2\hsize, clip]{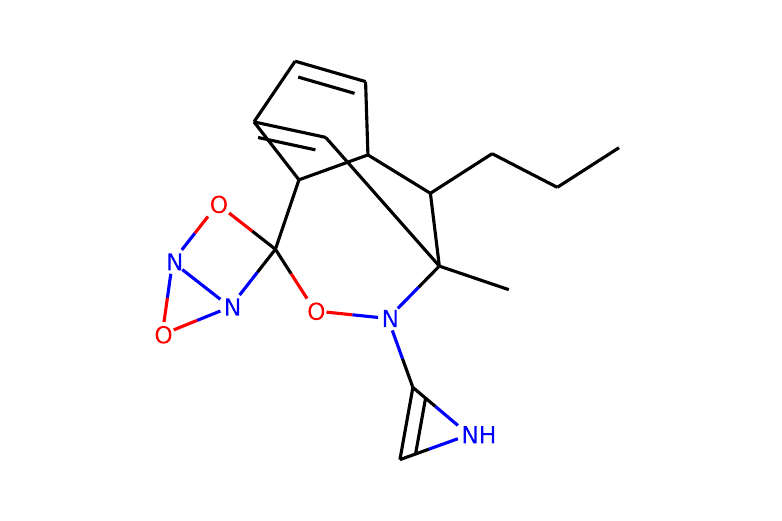} \hspace*{-1cm} &
			\includegraphics[width=0.2\textwidth]{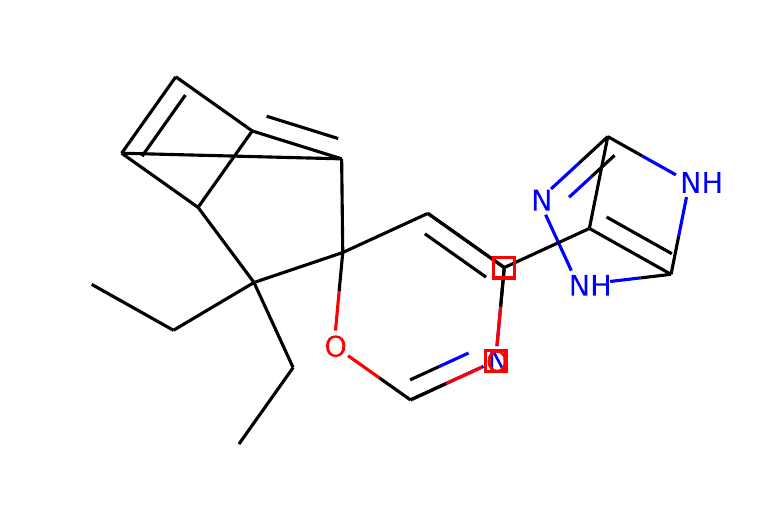}\hspace*{-1cm}
			\vspace{-2mm} \\
			
			%\hspace*{-2cm}
			\scriptsize $y_2(m) = 0.920$ (1st) &\scriptsize  $y_2(m) = 0.916$ (2nd)& \scriptsize  $y_2(m) = 0.912$ (3rd)
		\end{tabular}
	}
	\vspace*{-2mm}
	\caption{Visualization of the best three molecules generated by our property oriented decoder optimized for penalized logP (top row) and QED (bottom row) 
	scores.
	%
	% In each row block, the first row indicates the three dimensional molecular structure, whereas the second row shows its two dimensional graph structure.
	}
	\vspace*{-4mm}
	\label{fig:bestMollogP}
\end{figure}

\xhdr{Discovering molecules with high penalized logP and QED scores} 
To maximize the logP and QED, we consider the loss function $\ell(m)=-y_1(m)$ and $\ell(m)=-y_2(m)$, respectively, and compare our proposed property oriented decoder with three reinforcement learning 
based methods--- MOLGAN~\cite{decao2018molgan}, ORGAN~\cite{organ} and GCPN~\cite{gcpn}---and Bayesian optimization over the latent space of molecules under the encoders of several VAE based models---\ourmodel, GrammarVAE, 
CVAE and JTVAE. 
Appendix~\ref{app:bayes} contains additional details regarding our implementation of Bayesian optimization.

Table~\ref{tab:propval} shows the values of penalized logP and QED for the best three molecules generated by each method. 
The results show that our property oriented decoder, \ourmodel\, (Algorithm~\ref{alg:prop}), is able to identify molecules with property values $121$\% higher than those identified by the best performing competitor, \ie,
Bayesian optimization over the latent spaced of molecules generated by JTVAE.
Finally, note that, in contrast with all the competitors, our property oriented decoder is also able to provide a (plausible) spatial configuration for the atoms of each of the identified molecules, as shown in 
Figure~\ref{fig:bestMollogP}.

\xhdr{Discovering molecules with low potential energy values}
The stability of a molecule depends on its potential energy---a lower value of potential energy indicates higher stability.
In this section, our goal is to generate the most stable three dimensional structure for a given two dimensional molecular graph $\Gcal=(\Vcal, \Ecal)$ with atoms $\{\tb_u\}_{u\in\Vcal}$, initial spatial
coordinates $\xb'_{u}$ and bond-types $\Ycal$. % and initial spatial coordinates $\xb'_{u}$. 

To this aim, we only need to optimize the part of the decoder that generates the spatial coordinates rather than the entire decoder. Therefore, given a molecular graph, 
we solve the following optimization problem using the same gradient-descent algorithm described in Section~\ref{sec:prop}:
\begin{equation}
 \underset{p(\cdot | \Ecal, \Ycal, \Zcal)} {\text{minimize}} \quad   \EE_{\Zcal\sim q_{\phi}(.| \Vcal,\Ecal, \Ycal,\Fcal') } \left[ \EE_{ \{\xb_u\}_{ u\in\Vcal } \sim p(\cdot | \Ecal, \Ycal, \Zcal) } \left[ S( \{\xb_u\}_{ u\in\Vcal } |  \Ecal, \Ycal, \Zcal) \right]\right] , \label{eq:klx}
 \end{equation}
with
\begin{equation}
S( \{\xb_u\}_{ u\in\Vcal }| \Ecal, \Ycal, \Zcal) = \ell( \{\xb_u\}_{ u\in\Vcal }) + \rho \log \frac{p( \{\xb_u\}_{ u\in\Vcal }| \Ecal, \Ycal, \Zcal ) }{ p_{\theta} ( \{\xb_u\}_{ u\in\Vcal }| \Ecal, \Ycal, \Zcal)}, \label{eq:S-phix}
\end{equation}
where the loss $\ell(.)$ is the single point energy (SPE) in terms of Atomic Unit (a.u.), computed using the Gaussian toolbox by Frisch et al.~\cite{frisch2015gaussian}, and the outer expectation in Eq.~\ref{eq:klx} is computed over all possible latent vectors under the posterior distribution conditioned on the given molecular graph. 
% with features $\Fcal'=\{\tb_u,\xb'_u\}_{u\in\Vcal}$ and initial spatial coordinates $\xb' _{u}$.  

Figures~\ref{fig:SPE} and~\ref{fig:distr-spe} summarize the results for $100$ molecular graphs\footnote{\scriptsize The molecular graphs were not present in the 
training set used to train \ourmodel{}.}. 
The results show that, for a majority of the molecular graphs, we are able to find spatial configurations that increase their stability (\ie, decrease their potential 
energy). 

\begin{figure}[t]
	\centering
	\begin{tabular}{cc}
		\includegraphics[width=0.45\hsize, clip,height=35mm]{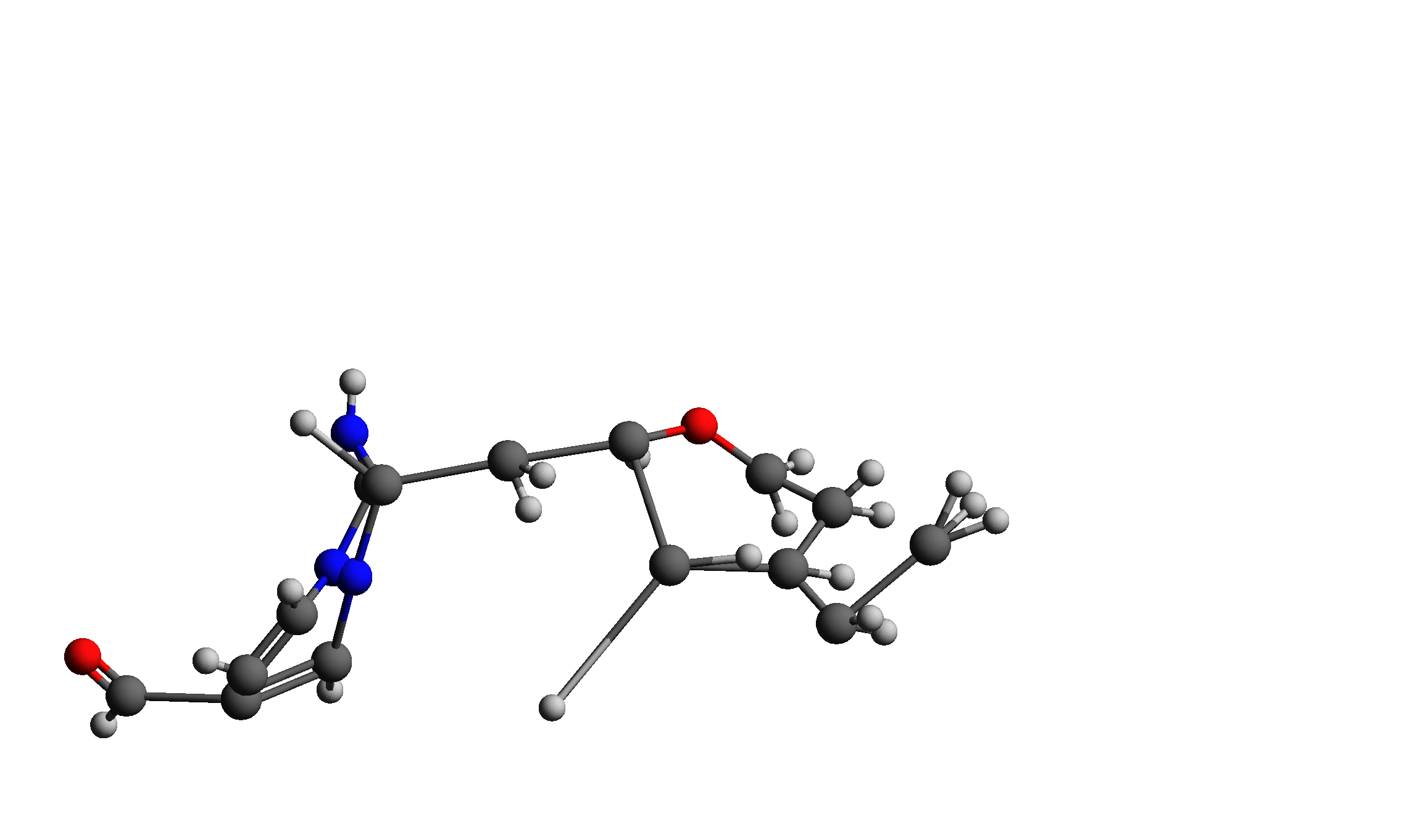} \hspace*{-1cm} &
		\includegraphics[width=0.45\hsize, clip, height=35mm]{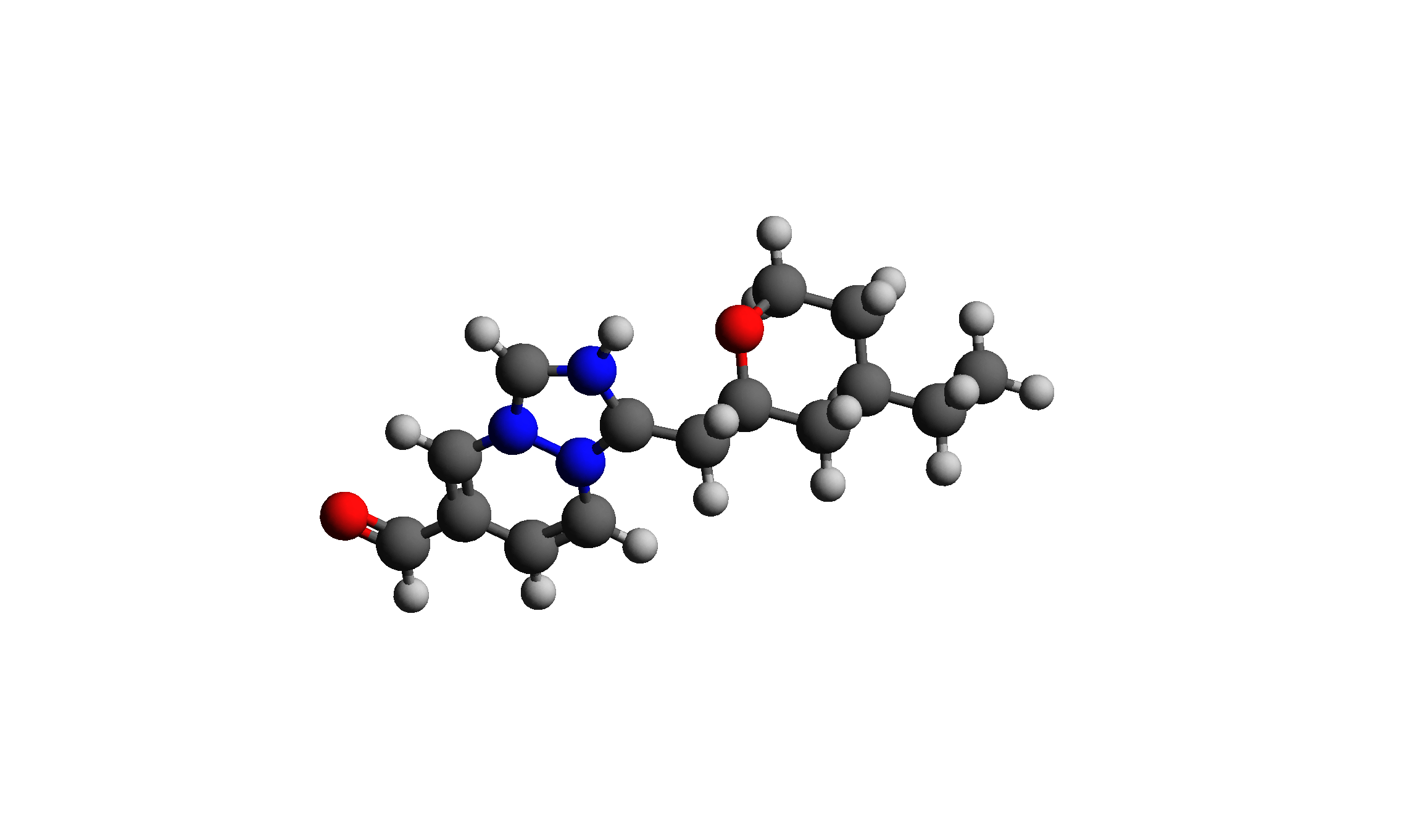} \hspace*{-1cm} \\ \vspace*{-0.4cm}
		 \scriptsize   Initial configuration,  $SPE = -895.7493$ a.u.  
		 & \scriptsize   Final configuration,  $SPE = -898.9146$ a.u.  \\ \\		 
		 	\includegraphics[width=0.45\hsize, clip]{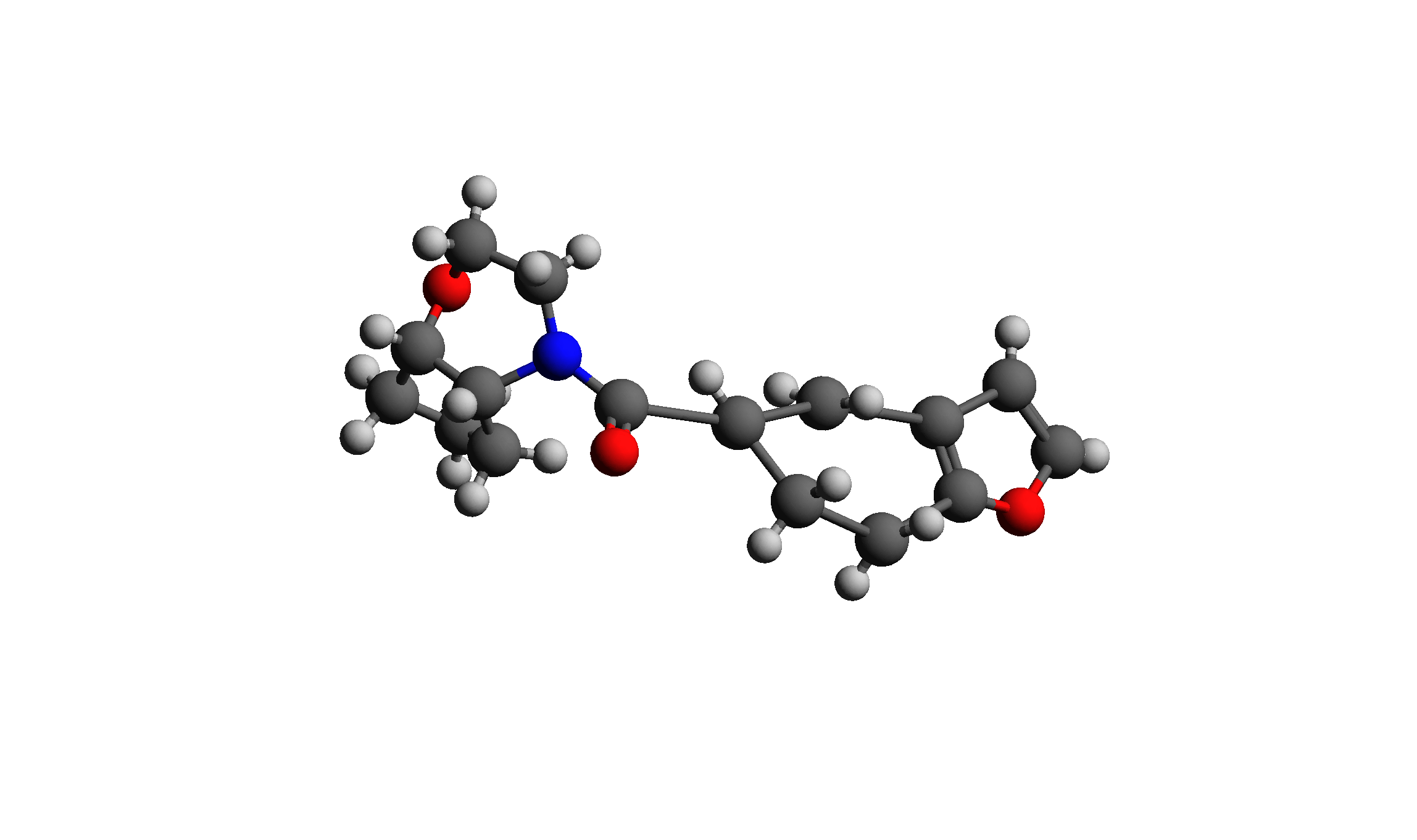} \hspace*{-1cm} &
		 	\includegraphics[width=0.40\hsize, clip]{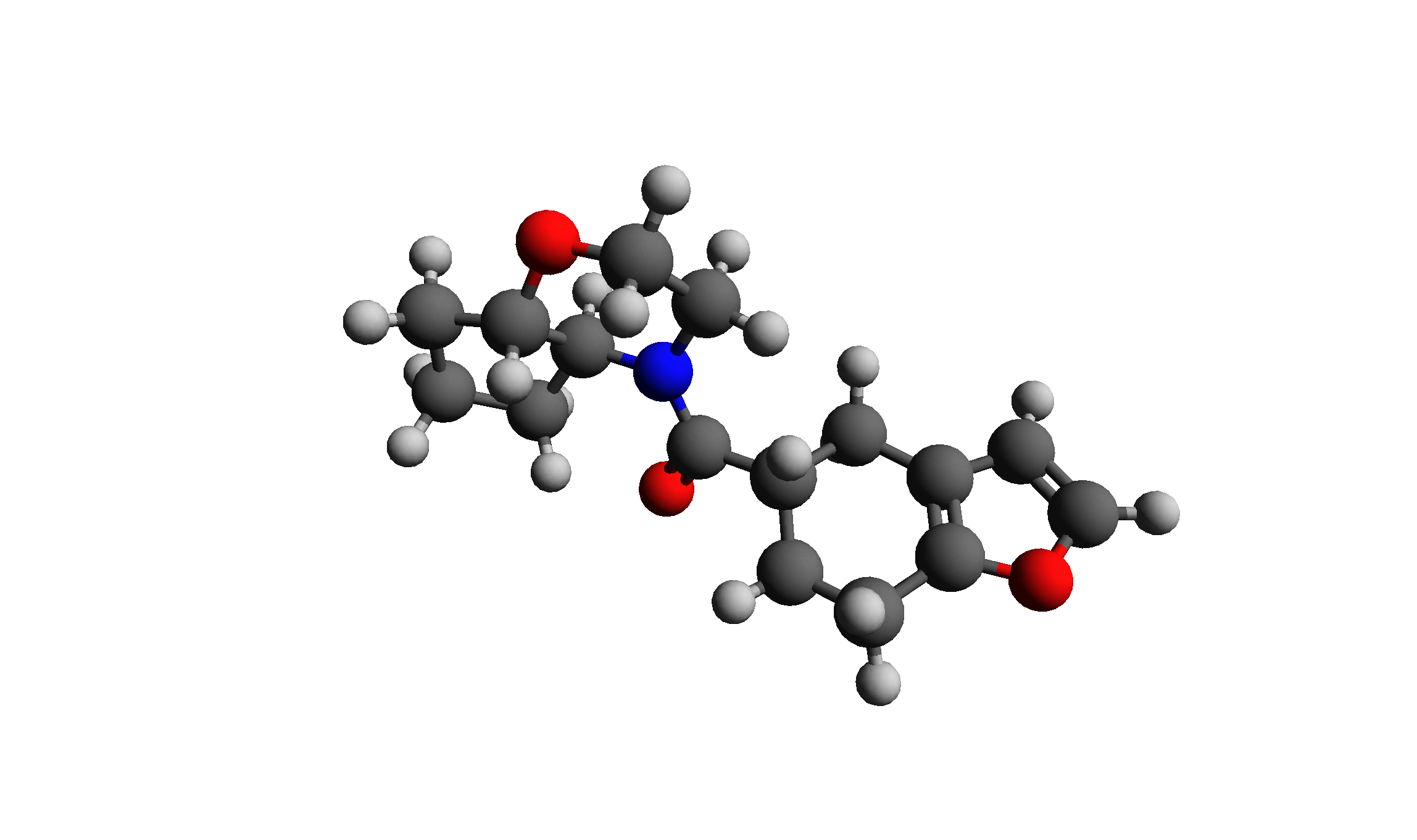} \hspace*{-1cm} \\ \vspace*{-0.4cm}
		 	\scriptsize Initial configuration, $SPE = -901.4993$ a.u. 
		 	& \scriptsize Final configuration, $SPE = -902.8904$ a.u.  \\ \\
	\end{tabular}
		\caption{Single point energy (SPE) for the initial spatial configuration (left column) and optimized spatial configuration (right column) for 
		two randomly sampled molecular graphs present in the ZINC dataset, where each row corresponds to each of the molecules.}
		\label{fig:SPE}
\end{figure}

\begin{figure}[t]
	\centering
	\includegraphics[width=0.4\textwidth]{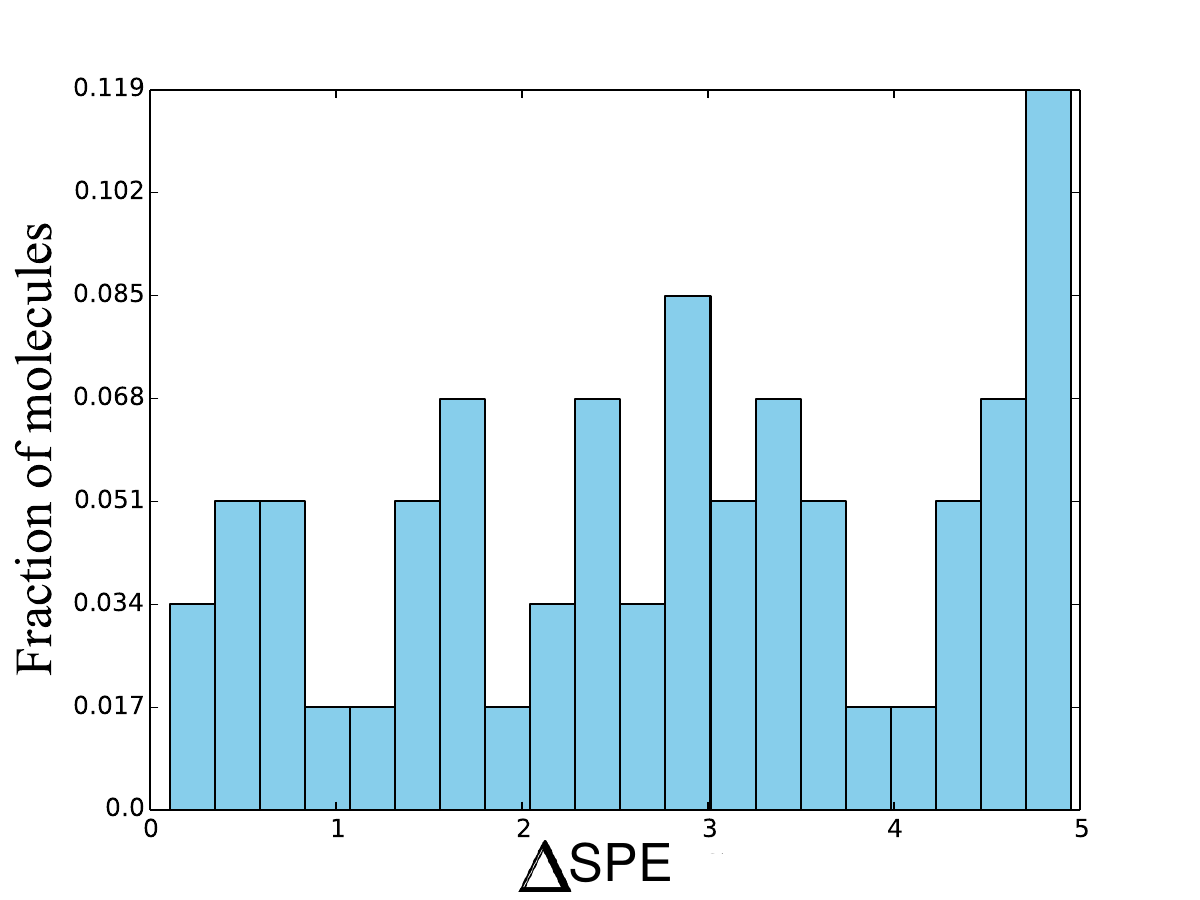} 
	\caption{Difference between single point energies between the initial and the optimized spatial configurations ($\Delta$SPE) for $100$ randomly 
	sampled molecular graphs present in the ZING dataset. The final optimized spatial configuration has always lower SPE.}% It shows that, XX}
	\label{fig:distr-spe}
\end{figure}
\iffalse
\begin{figure}[t]
	\centering
	%\vspace*{-1mm}
	%\resizebox{\hsize}{
	%\resizebox{\hsize}{!}{
	\begin{tabular}{ccccc}
		%\hspace*{-2cm}
		\includegraphics[width=0.22\textwidth]{FIG/RL/SPE/40_best_2_new-min.pdf} 
		&\includegraphics[width=0.22\textwidth]{FIG/RL/SPE/40_best_5_modified-min.pdf}
%		&\includegraphics[width=0.9\textwidth]{FIG/RL/SPE/40_best_3_modified.pdf}
	    &\includegraphics[width=0.22\textwidth]{FIG/RL/SPE/40_test_modified-min.pdf}
		&\includegraphics[width=0.22\textwidth]{FIG/RL/SPE/40_original_modified-min.pdf}\\
		 \scriptsize $SPE = -770.7718\text{ }a.u$  
		& \scriptsize $SPE = -770.7931\text{ }a.u$  
%		&  $SPE = -770.8463 a.u$
		& \scriptsize $SPE = -770.8506\text{ }a.u$ 
		& \scriptsize  $SPE = -770.8710\text{ }a.u$ 
	\end{tabular}
	%}
	\vspace*{-2mm}
	\caption{Transition of single point energy (SPE) during  stable structure discovery for a two dimensional molecular graph present in ZINC dataset. }
	\vspace*{-4mm}
	\label{fig:SPE}
\end{figure}
\fi

%\section{Experiments of KL control}
%\input
%\input{051expt-synthetic-RL}

% \section{Related Work}
% \label{sec:related}
% \input{060related}
% % \vspace*{-0.4cm}
\section{Conclusion}
% \vspace*{-0.3cm}
\label{sec:conclusions}
In this work, we have introduced a variational autoencoder for molecular graphs, that is permutation invariant of the nodes labels of the 
graphs they are trained with, and allow for graphs with different number of nodes and edges as well as three dimensional spatial coordinates for atoms. 
Moreover, the decoder is able to guarantee a set of local structural and functional properties in the generated graphs through masking. 
Then, we have developed a gradient based algorithm to optimize the decoder of our model so that it learns to generate molecules that
maximize the value of certain property of interest.
Finally, we have shown that our variational autoencoder is able to discover plausible, diverse and novel molecules more effectively than 
several state of the art methods and, for several properties of interest, our optimized decoder is able to identify molecules with property
values $121$\% higher than those identified by several state of the art methods.

Our work also opens many interesting venues for future work. 
For example, in the design of our variational autoencoder, we have assumed graphs to be static, however, it would be interesting 
to augment our design to dynamic graphs by, \eg, incorporating a recurrent neural network or long short-term memory (LSTM) units.
Moreover, we have focused on molecular graphs, however, we believe our methodology could be adapted to other real-world graphs.
Finally, there are other problems related to molecular design, such as retro synthesis~\cite{segler2018planning}, where machine learning 
may advance the state of the art.

\setlength{\abovedisplayskip}{4pt}
\setlength{\abovedisplayshortskip}{1pt}
\setlength{\belowdisplayskip}{4pt}
\setlength{\belowdisplayshortskip}{1pt}
\setlength{\jot}{3pt}

% \clearpage
% \newpage
\bibliographystyle{unsrt}
{
\small
\bibliography{vae}
}

\newpage
\appendix

\label{sec:appen}
\section{Implementation Details} \label{app:implementation}

\xhdr{Architecture details}
Table~\ref{tab:implementation} provides additional details on the architecture of our variational autoencoder for graphs, where it is important to notice that the parameters 
to be learned do not depend on the size of the graphs (\ie, the number of nodes and edges). Note that, $\rb$ and $\gb$ are linear forms and the aggregator function $\Lambdab$ 
is a sum, which is a symmetric function, for simplicity\footnote{\scriptsize We did experiment with other symmetric aggregator functions such as pooling, as in the inductive graph 
representation learning~\cite{hamilton2017inductive,lei2017}, and did not notice significant gains in practice.}.
\begin{table}[!h]
	\centering
	\small
	\scalebox{0.8}{
		\begin{tabular}{|l |l |l |l |l |l|}
			\hline
			\textbf{Layer} &  \textbf{Architecture} & \textbf{Inputs} &  \textbf{\specialcell{Type of \\non-linearity}} & \textbf{\specialcell{Parameters \\to be learned}} 
			& \textbf{Output}\\
			\hline
			Input & \specialcell{Feedforward \\ ($K$ layers)} & $\Ecal, \Fcal, \Ycal$ & \specialcell{$\rb(\cdot)$: Linear \\ $\gb(\cdot)$: Linear \\ $\Lambdab(\cdot)$: Sum} & $W^{\Tcal}_1, W^{\Xcal}_1,...,W^{\Tcal}_K, W^{\Xcal}_K$ & $\cb_1, \ldots, \cb_n$ \\
			\hline
			Encoder & \specialcell{Feedforward \\ (Two layers)} & $\cb_1, \ldots, \cb_n$ & \specialcell{Softplus \\ Softplus \\ Softplus}& \specialcell{$W_{h}, b_{h}$ \\ $W_{\mu}, b_{\mu}$, \\$W_{\sigma}, b_{\sigma}$}& \specialcell{$\mub_1, \ldots, \mub_n$ \\ $\sigmab_1, \ldots, \sigmab_n$} \\
			\hline
			Decoder & \specialcell{Feedforward \\ (One layer)} & $\Zcal$ & Softplus & $W, b$ & $\Ecal, \Ycal, \fb$ \\
			\hline
		\end{tabular}}
			\caption{Details on the architecture of \ourmodel.}
	\label{tab:implementation}
	\vspace{-3mm}
\end{table}

\xhdr{Hyperparameter tuning}
At the very outset, to train \ourmodel, we implemented stochastic gradient descent (SGD) using the Adam optimizer.
Therein, we had to specify four hyperparameters: (i) $D$  -- the dimension of $z_u$, (ii) $K$  -- the maximum number of hops used in encoder to aggregate information, (iii) $L$ -- the number of negative samples, 
(iv) $l_r$-- the learning rate. Note that, all the parameters $\Wb_{\bullet}$'s and $\bb_{\bullet}$'s in the input, hidden and output layers depend on $D$ and $K$.
We selected these hyperparameters using cross validation. More specifically, we varied $l_r$ in a logarithmic scale, \ie, $\{0.0005, 0.005, 0.05, 0.5\}$, and the rest of the hyperparameters in an arithmetic scale, 
and chose the hyperparameters maximizing the value of the objective function in the validation set. For synthetic (real) data, the resulting hyperparameter values were $D=7 (5)$, $K=3 (5)$, $L=10 (10)$ and 
$l_r = 0.005 (0.005)$. 
% Note that, a small learning rate facilitates convergence of the objective while using Adam optimizer.
%
% Furthermore, we used Softplus as the nonlinear activation functions (Table~\ref{tab:implementation}). 
% We have three hyper-parameters like the dimension of $dim(\zb)$, the random walk length $K$ and the negative sampling size $L$. We have first selected a small set of samples from real data and run our model with various combinations of $dim(\zb)$, $k$ and $L$. We found the best performance (minimum loss) for corresponding values of 7, 5 and 10 respectively. So, we selected these parameter values to train the entire set of samples.
%
To run the baseline algorithms, we followed the instructions in the corres\-pon\-ding repository (or paper). 
\begin{algorithm}[h]                    % enter the algorithm environment
	\small
	\caption{Training with Minibatches}%($G, k, dim(z), totalEpoch$)}          % give the 
	\label{alg:training}                           % and a label for \ref{} commands later in the document
	\begin{algorithmic}[1]                    % enter the algorithmic environment
		\STATE \textbf{Input: } Training graphs $\{\Gcal_i(\Vcal_i,\Ecal_i)_{i\in [N]}\}$, hyperparameters $\psi=\{D,K,L,l_r\}$, $\Theta$
		\STATE \textbf{Output: }Inferred parameters $\hat{\Theta}$.
		\vspace{1mm}
% 		\STATE \texttt{/*Create batches*/}
		\STATE $\hat{\Theta}\leftarrow\text{Initialize} (\Theta)$ 
		\STATE $\Bcal \leftarrow \text{CreateBatches}(\{\Gcal_i(\Vcal_i,\Ecal_i)_{i\in [N]}\})$
%		\FOR{$i\in [N]$}
%		  \STATE $\Bcal_{|\Vcal_i|}\leftarrow \Bcal_{|\Vcal_i|} \cup\{\Gcal_i\}$
%		\ENDFOR
% 		\STATE Generate a list $L$ with unique node numbers 
% 		\STATE $epoch$ $\leftarrow$ 0
% 		\STATE $\theta$ $\leftarrow  \emptyset$ 
% 		\WHILE {$epoch < totalEpoch$}
		%\FOR{$ \in L$}
		\FOR{$\Bcal_k \in \Bcal$}
		\STATE NeVAE$_{\hat{\Theta}}$ $\leftarrow \text{BuildComputationalGraph}(\text{Nodes}(\Bcal_k), \hat{\Theta})$
% 		\IF{$isEmpty$($\theta$)}		 
% 		\STATE $\theta \leftarrow$ $Initialize(W_1,...,W_k, W_h, b_h, W_\mu, b_\mu, W_\sigma,b_\sigma, W, b)$
% 		\ELSE
% 		\STATE $\hat{\Theta}_{k} \leftarrow$ $\text{Restore}(\Theta)$		
		%\FOR
		\STATE $\hat{\Theta} \leftarrow \text{Train}(\text{\ourmodel}_{{\hat{\Theta}}}, \Bcal_k)$
%		\STATE $\hat{\Theta}\leftarrow\hat{\Theta}_k$
% 		\STATE $\text{Store}(\Theta)$
		\ENDFOR
		%\ENDFOR
% 		\ENDWHILE
		
	\end{algorithmic}
	\label{alg:opDynamicsSim} 
\end{algorithm}
%
% \clearpage
% \newpage

\xhdr{Training with minibatch}
We implemented stochastic gradient descent (SGD) using minibatches, where each batch contained graphs with the same number of nodes. 
More specifically, we first group the trai\-ning graphs $\Gcal_i$'s into batches $\Bcal = \{\Bcal_k\}$ such that $|\Vcal_i|=|\Vcal_j|$ for all $\Gcal_i, \Gcal_j \in \Bcal_k$. Then, at each 
iteration, we select a batch at random, build a computation graph for the number of nodes corresponding to the batch using the parameters estimated in the previous iteration, 
and update the parameters using the computation graph and the batch of graphs. 
%
% For the next batch $\Bcal_{k+1}$, it is again restored and used as the initial value, to obtain $\hat{\Theta}_{k+1}$. 
% Hence, the network building overhead time is reduced from per sample to per batch.
Such a procedure helps to reduce the overhead time for building the computational graph, from per sample to per batch. 
%
% As Tensorflow has certain limitations in terms of handling variable length inputs, we process the data and batch them based on the number of nodes. 
%
% Then for each batch, we build the neural network and train the network for all graphs with same number of nodes. 
% For the next batch, we restore the learned parameters and build the new network. 
%
This batching and training process is summarizedd in Algorithm~\ref{alg:training}, where ``CreateBatches(...)" group the training graphs into batches, ``BuildComputationalGraph(...)'' builds the 
computation graph ``NeVAE" using the parameters from the previous iteration and a given number of nodes, ``Nodes(...)" returns the number of nodes of the graphs in a batch, and ``Train(...)'' 
updates the parameters given the computation graph and the parameters from the previous iteration.

\xhdr{Hardware and software specifications}
We carried out all our experiments for \ourmodel\ using Tensorflow 1.4.1, on a 64 bit Debian distribution with 16 core Intel Xenon CPU (E5-2667 v4 @3.20 GHz) and 512GB RAM.

% \clearpage\newpage

% 
% \clearpage
% \newpage
\section{Additional details on Bayesian optimization}\label{app:bayes}
%
%Here, we leverage our model to discover novel molecules with desirable properties. 
%
To implement Bayesian optimization (BO) for property-oriented molecule generation, we proceed similarly as in previous work~\cite{gomez2016automatic,kusner2017grammar,jin2018junction}.
More specifically, we first sample $3{,}000$ molecules from our ZINC dataset, which we split into training (90\%) and test (10\%) sets. 
Then, for our model and each competing model with public domain implementations, we train a sparse Gaussian process (SGP)~\cite{snelson2006sparse} with the latent 
representations and $y(m)$ values of $100$ inducing points sampled from the training set.
The SGPs allow us to make predictions for the property values of new molecules in the latent spaces. 
Then, we run $5$ iterations of batch Bayesian optimization (BO) using the expected improvement (EI) heuristic~\cite{jones1998efficient}, with $50$ (new) latent vectors (molecules)
per iteration.

In this section, we complement the performance comparison between \ourmodel{}, GrammarVAE, CVAE, and JTVAE from Table~\ref{tab:propval} using two additional 
quality measures:
\begin{itemize}[noitemsep,nolistsep,leftmargin=0.9cm]
\item[(a)] the predictive performance of the trained SGPs in terms of log-likelihood (LL) and root mean square \-error (RMSE) on the test set; and,
\item[(b)] the average value $\EE\left[y(m)\right]$, fraction of valid molecules and fraction of \emph{good} molecules, \ie, $y(m) > 0$, among the molecules found using EI.
\end{itemize}
\indent Table~\ref{tab:score1}, Figure~\ref{fig:BO_score1} and Figure~\ref{fig:bestMol} summarize the results. In terms of log-likelihood and RMSE, the SGP trained using the latent 
representations provided by 
our model outperforms all baselines.
In terms of the property values $\EE\left[y_1(m)\right]$ of the discovered molecules and fraction of valid and good molecules, BO under \ourmodel\ also outperforms all baselines; however, in terms of $\EE\left[y_2(m)\right]$ this is the second best after JTVAE.
Here, we would like to highlight that, while BO under JTVAE is able to find a few molecules with larger property value than BO under \ourmodel, it is unable to discover a sizeable set
of unique molecules with high property values.
\begin{table}[h]
	\vspace{-1mm}
% 	\smaller
	\centering
	\scalebox{0.7}{
		\begin{tabular}{l|l|l|l|l||l |l | l | l}
			\hline
			{Objective} &  \multicolumn{4}{c||}{Penalized logP} & \multicolumn{4}{c}{QED} \\
			\hline
			
			&{\ourmodel}&  {GrammarVAE} &  {CVAE}&JTVAE  &{\ourmodel}&  {GrammarVAE} &  {CVAE}&JTVAE\\
			
			\hline\hline
			{LL} & \textbf{-1.45} & -1.75 & -2.29& \textit{-1.54}& \textbf{-1.27} &-1.75 & -1.77 & \textit{-1.48} \\ % &-1.697 \\
			% {LL} & {-1.631} & -1.658 & -1.980 & -2.153 \\
			\hline
			{RMSE} & \textbf{1.23} & 1.38 &1.80 & \textit{1.25}&  \textbf{0.85}& 1.37 &1.39 & \textit{1.07} \\ % & 1.366\\
			\hline\hline
			{Fraction of \emph{valid} molecules} &\textbf{1.00} & 0.77 & 0.53 &\textbf{1.00}& \textbf{1.00} &  0.42 &0.33 &\textbf{1.00}\\
			\hline
			%{Avg. scores} & \textit{0.80} & -4.30 & -14.40 & \textbf{2.10}\\
			%\hline 
			%{Fraction of \emph{good} molecules}&\textit{0.66}& 0.27 & 0.23 & \textbf{1.00}\\
			%\hline 
			{Fraction of \emph{unique} molecules} &\textbf{0.58} & 0.29 & 0.41 &{0.32} &\textbf{0.61}& 0.22 &0.20 & 0.42 \\
			\hline
% 			{Best scores}& 2.89& 2.27 &0.76 & \textbf{4.56} \\
% 			\hline
		\end{tabular}
	}
	\vspace*{-0.1cm}
	%\caption{Results finding best molecule. Column 4 indicates fraction of sampled molecules with score greater than 0 }
	%\end{tabular}}
	\caption{Property prediction performance (LL and RMSE) using Sparse Gaussian processes (SGPs) and
		property maximization using Bayesian Optimization (BO).} 
	\label{tab:score1}
\vspace*{-0.3cm}
\end{table}

\begin{figure}[h]
\vspace{-4mm}
	\centering
	\subfloat[Uniqueness ($y_1(m)$)]{\includegraphics[width=0.20\textwidth]{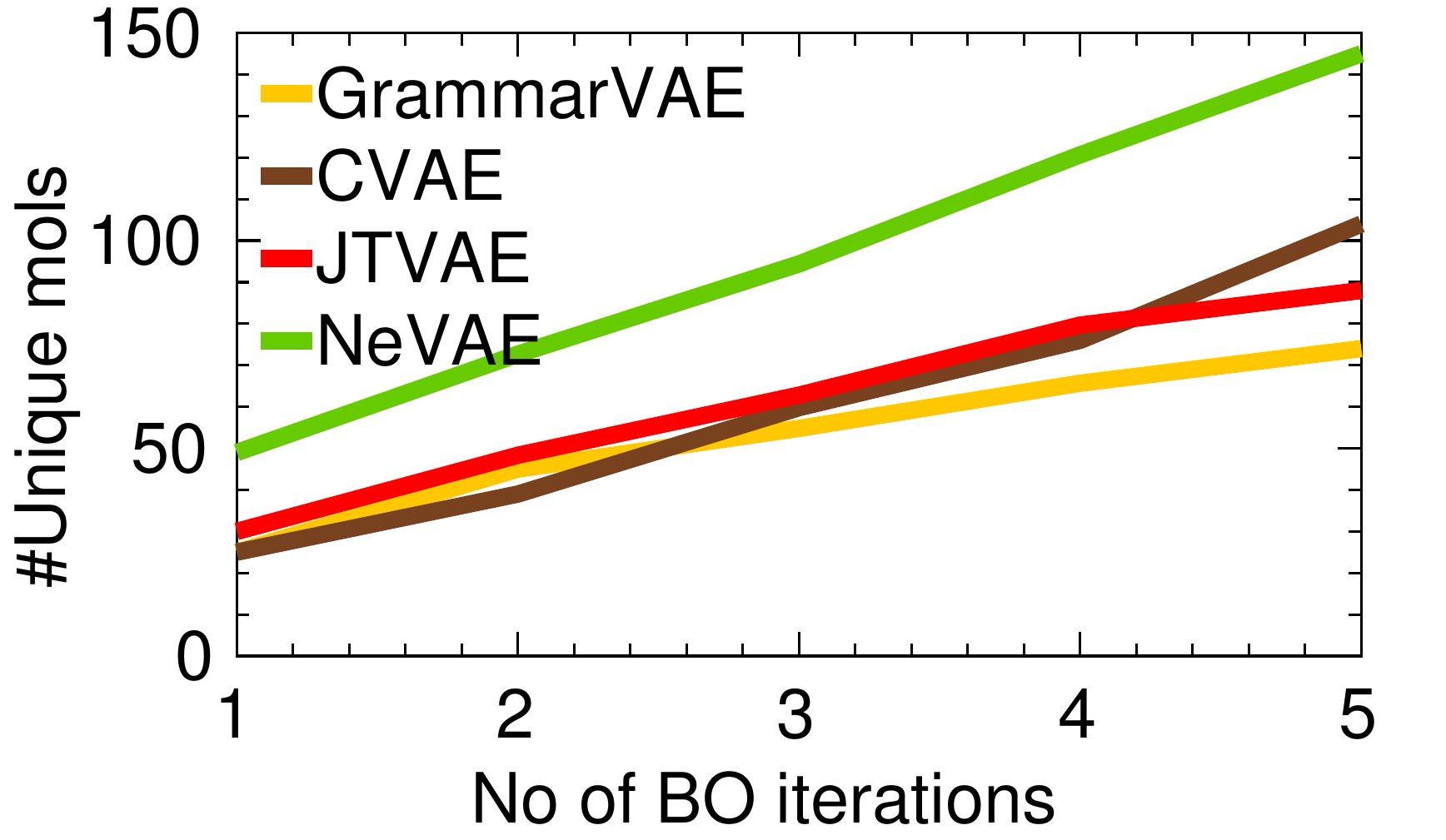}}
	\hspace*{0.5 cm}
	\subfloat[Uniqueness ($y_2(m)$)]{\includegraphics[width=0.20\textwidth]{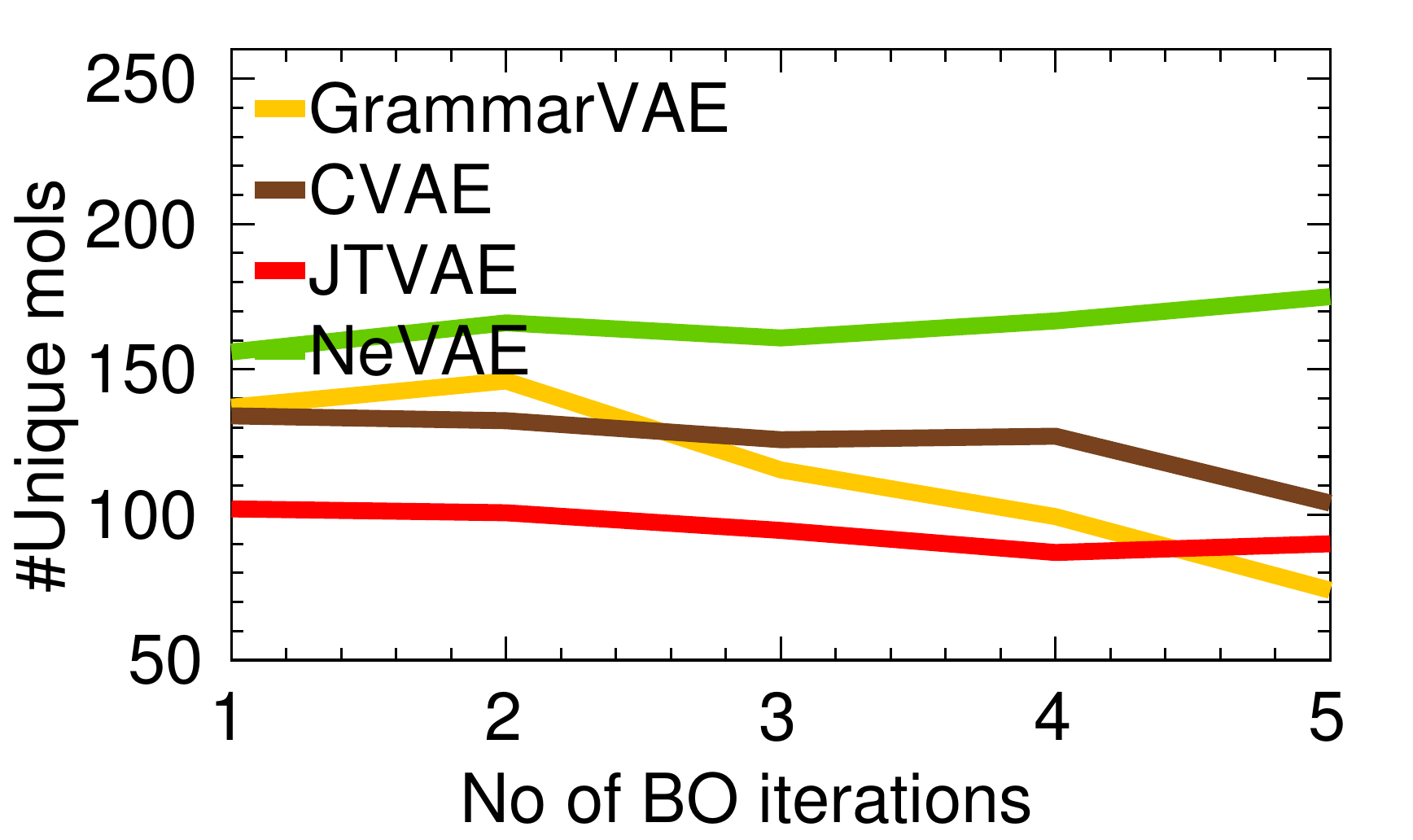}}
	\hspace*{0.5 cm}
% % 	
	\subfloat[Score ($y_1(m)$)]{\includegraphics[width=0.20\textwidth]{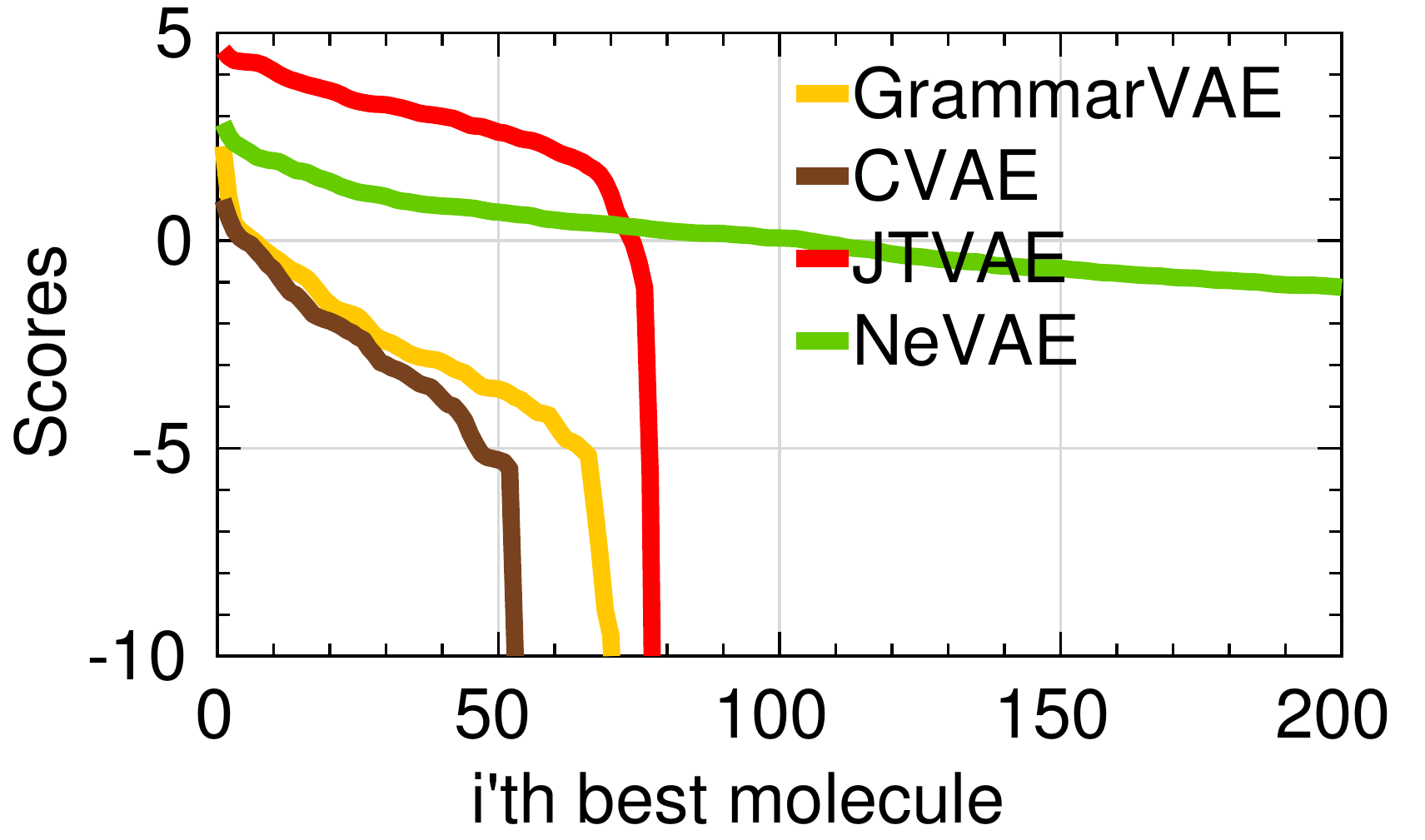}} 
	\hspace*{0.5 cm}
	\subfloat[Score ($y_2(m)$)]{\includegraphics[width=0.20\textwidth]{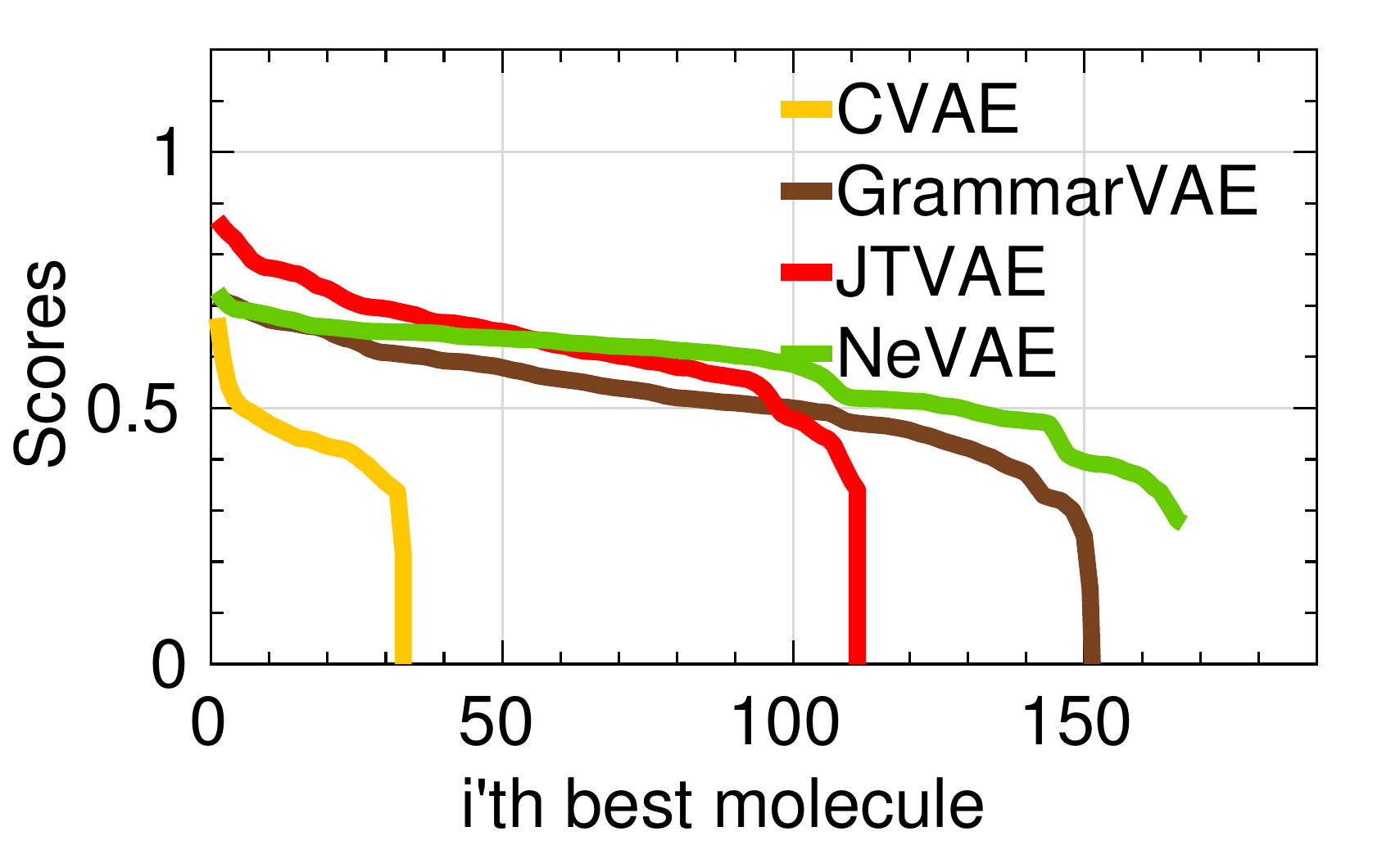}}\vspace*{-0.2cm}
% 			\subfloat[Score]{\includegraphics[width=0.20\textwidth]{FIG/BO/rule_based}}
	\caption{ Property maximization using Bayesian optimization.
% 		Each plot shows the values of $y_1(m)$ in decreasing order for unique molecules.
		%
		Panel (a) and (b) show the variation of Uniqueness with the no. of BO iterations for $y_1(m)$ and $y_2(m)$ respectively. 
		Panel (c) and (d) show the values of $y_1(m)$ and $y_2(m)$ sorted in the decreasing order. 	
	}
 	\vspace*{-0.2cm}
	\label{fig:BO_score1}
\end{figure}

\begin{figure}[h]
	\centering
	\vspace*{-4mm}
	\begin{tabular}{ccc}
		\includegraphics[width=0.12\textwidth]{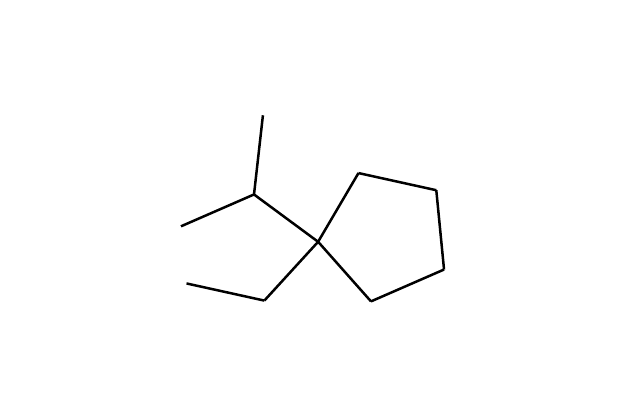} \vspace{-1mm} &
		\includegraphics[width=0.12\textwidth]{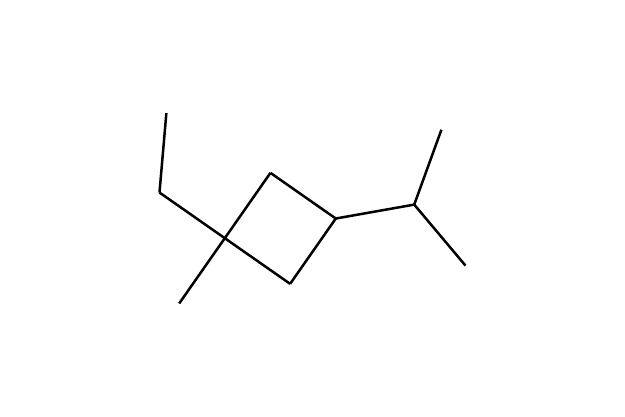} \vspace{-1mm} &
		\includegraphics[width=0.12\textwidth]{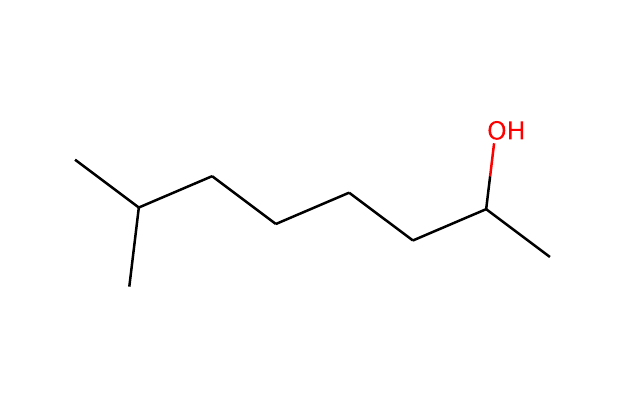} \vspace{-1mm} \\%  \vspace{-1mm}
		\scriptsize $y(m)=2.826$ (1st) &\scriptsize  $y(m)= 2.477$ (2nd) &\scriptsize  $y(m)= 2.299$ (3rd)
	\end{tabular}
	\vspace*{-2mm}
	\caption{Best molecules found by Bayesian Optimization (BO) using our model.}
       \vspace*{-4mm}
	\label{fig:bestMol}
\end{figure}

\section{Additional Experiments on Synthetic Graphs} \label{app:experiments-syn}
In this section, we first demonstrate that our original model \ourmodel\ is able to generate graphs with a predefined local topological property, \ie, graphs without
triangles. Then, we show that our model is able to learn smooth latent representations of a popular type of random graphs, Kronecker graphs~\cite{leskovec2010kronecker}.
Then, we  present additional quantitative results on the ability of our model to learn and mimic the generative processes 
that determine the absence or presence of nodes and edges in 
of Kronecker graphs and Barab{\'a}si-Albert graphs~\cite{barabasi1999emergence}, a scalability analysis and finally illustrate the effect of node label permutations on the decoder parameter estimation.
Finally, we show that the optimal property-oriented decoder designed using variational inference is able to generate synthetic graphs with certain structural properties.
% % 

\subsection{Experimental setup} 
We first generate two sets of synthetic networks, each containing $100$ graphs, with up to $n = 1000$ number of nodes. The first set contains triangle free graphs 
and the second set contains a 50\%-50\% mixture of Kronecker graphs with initiator matrices: 
$\Thetab_1= [0.9,0.6;0.3,0.2]$, and $\Thetab_2 = [0.6,0.6;0.6,0.6]$. 
For each dataset, we train our variational autoencoder for graphs by maximizing the corresponding evidence lower bound (ELBO).
Then, we use the trained models to generate three sets of $1000$ graphs by sampling from the decoders, \ie, $\Gcal \sim p_{\theta}(\Gcal | \Zcal)$, 
where $\Zcal \sim p(\Zcal)$. 

\subsection{Quality of the generated graphs}

% \xhdr{Graphs with a predefined local topological property} 
%
We first evaluate the ability of our model to generate triangle free graphs by measuring the validity of the generated graphs, \ie, $\text{Validity} :=$ $ {|\{\Gcal_i\in\GG \,|\, \Gcal_i \text{ has 
no triangles}\}|}/{|\GG|}$, where $\GG$ is the set of $1000$ graphs generated using the trained model. We experiment both with and without masking during training and during test time. 
% \niloy{What about if we lose the metrics a bit - with at most $k$ triangles}
% \niloy{If we use masking during training does the inference time increase?}
% \manuel{Add relative increase in % in running time due to masking when ready, bidisha}
% Table~\ref{tab:masking} summarizes the results, 
%
We observe that, if we train and test our model with masking, it achieves a validity of $100$\%, \ie, it always generates triangle free graphs. 
If we only use masking during training, our model is able to achieve a validity of $68$\%, and, if we do not use masking at all, our models achieves 
a validity of $57$\%.
Moreover, while using masking during test does not lead to significant increase in the time it takes to generate a graph, using masking during training
does lead to an increase of $18$\% in training time.
%
% \niloy{what not used}
% \manuel{rewritten}
%
Figure~\ref{fig:tree} shows several example of triangle free graphs generated by our model.
  \begin{figure}[!h]
	\centering
	 {\includegraphics[width=0.25\textwidth]{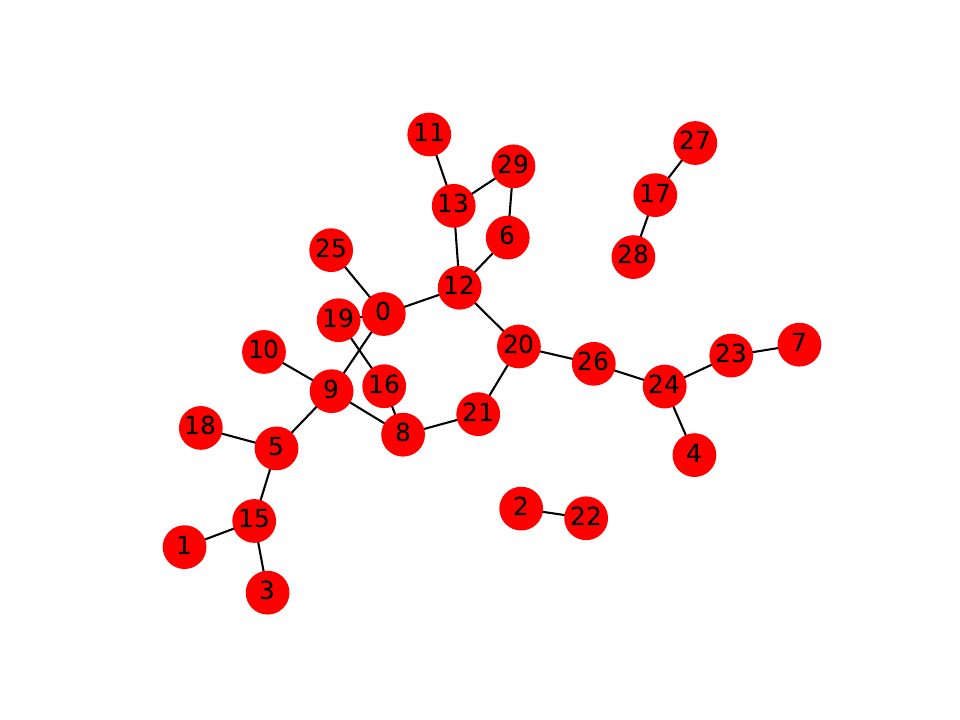}} 
	 {\includegraphics[width=0.25\textwidth]{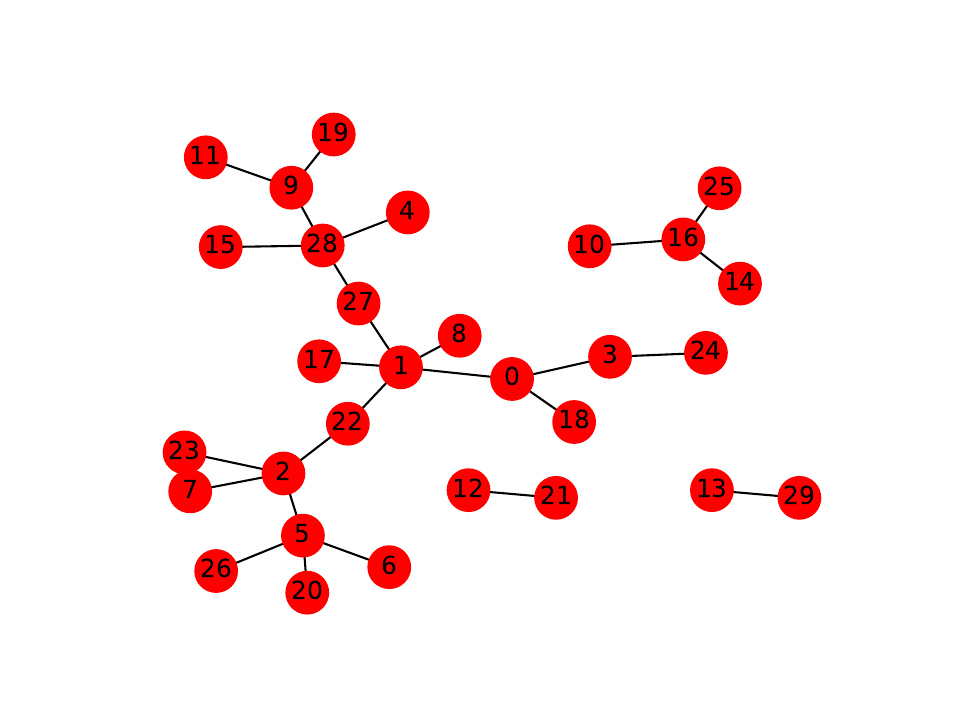}}
	{\includegraphics[width=0.25\textwidth]{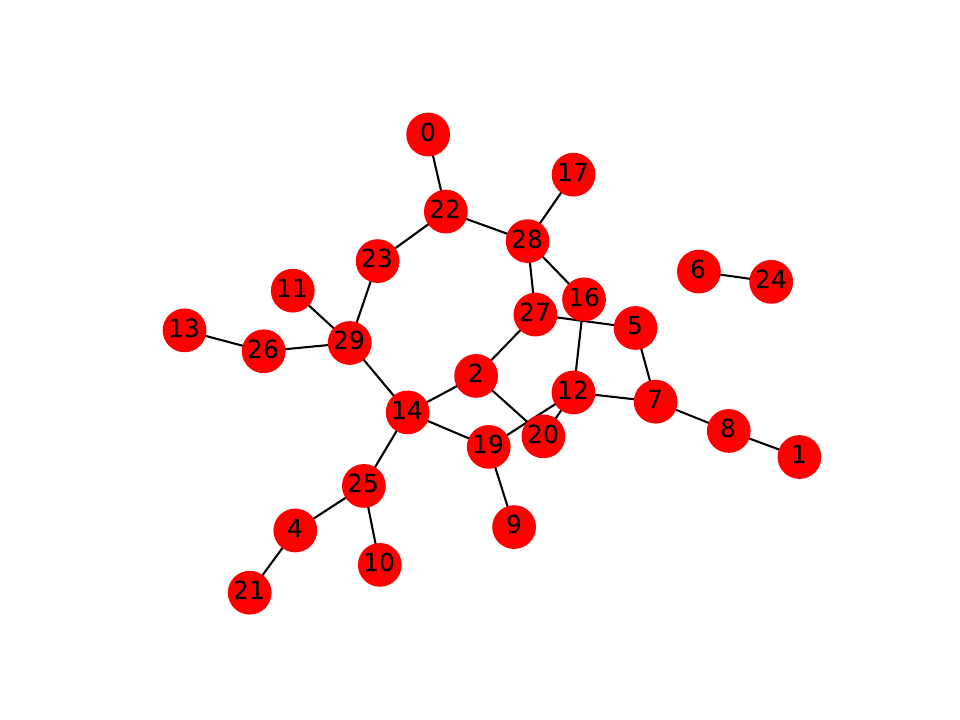}} 
	\caption{Graphs sampled using our variational autoencoder trained with a set of triangle free graphs. By using masking, our variational autoencoder is able
	to always generate triangle free graphs.}
	\label{fig:tree}
\end{figure} 

% \xhdr{Generalization ability}
%
Next, we evaluate the ability of our model to learn smooth latent representations of Kronecker graphs as follows.
% smoothly interpolate between 
% Kronecker graphs, as if they were generated by true Kronecker random graph models with different parameters. 
%
% More specifically, we proceed as follows.
%
% app:experiments-syn
%
First, we select two graphs ($\Gcal_0$ and $\Gcal_1$) from the training set, one generated using an initiator matrix $\Thetab_0 = [0.9,0.6;0.5,0.1]$ and the other 
using $\Thetab_1 = [0.6,0.6;0.6,0.6]$.
Then, we sample the latent representations $\Zcal_0$ and $\Zcal_1$ for $\Gcal_0$ and $\Gcal_1$, respectively, and sample new graphs from latent values $\Zcal$ in 
between these latent representations (using a linear interpolation), \ie, $\Gcal \sim p_{\theta}(\Gcal | \Zcal)$, where $\Zcal = a \Zcal_0 + (1-a) \Zcal_1$ and $a \in [0, 1]$, and the node labels, which define the matching between pairs of nodes in both graphs, are arbitrary.
Figure~\ref{fig:KronInter} provides an example, which shows that, remarkably, as $\Zcal$ moves towards $\Zcal_0$ ($\Zcal_1$), the sampled graph becomes similar to that of 
$\Gcal_0$ ($\Gcal_1$) and the inferred initiator matrices along the way smoothly interpolate between both initiator matrix. Here, we infer the initiator matrices of the graphs
generated by our trained decoder using the method by Leskovec et al.~\cite{leskovec2010kronecker}.
%
%\niloy{Table ~\ref{tab:results_synthetic} shows that the generated graph follow the general principles of BA or Kronecker
% and has nothing to do with interpolation - that needs to be disambiguated}
% I have clarified further.
Table~\ref{tab:results_synthetic} provides a quantitative evaluation of the quality of the generated graphs, \ie, it shows that the graphs our model generates are \emph{indistinguishable} from 
true Kronecker graphs.

\begin{figure*}[!t]
    \centering
\vspace{3mm} \hspace*{-1cm}\scalebox{0.9}{\begin{tabular}{cccc}
\includegraphics[width=0.23\textwidth]{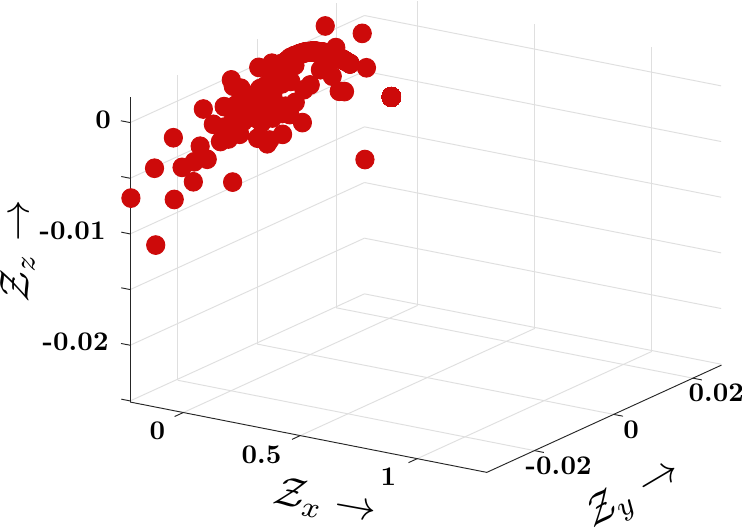} &
\hspace*{0.4cm}\includegraphics[width=0.23\textwidth]{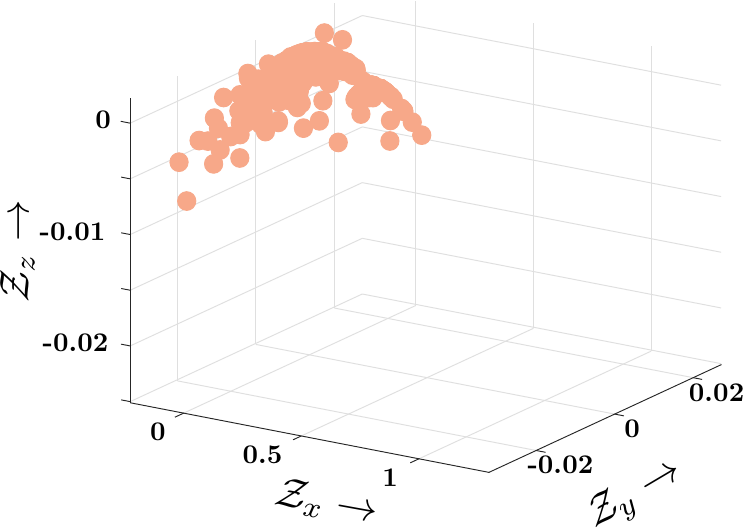} &
\hspace*{0.4cm}\includegraphics[width=0.23\textwidth]{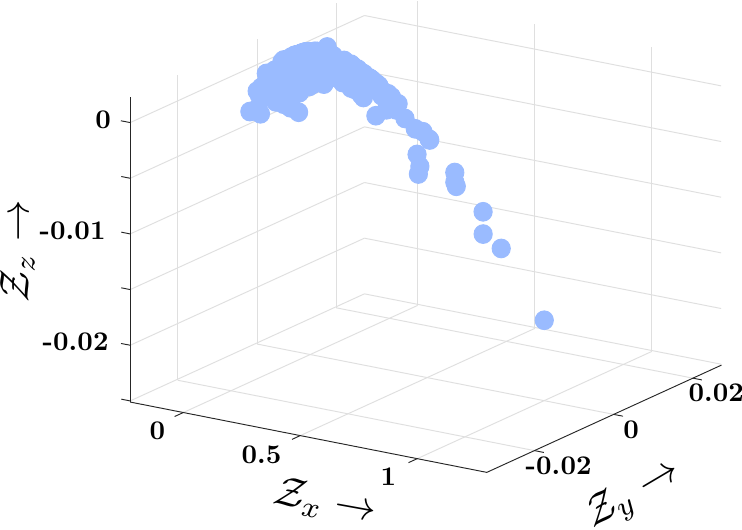} &
\hspace*{0.4cm}\includegraphics[width=0.23\textwidth]{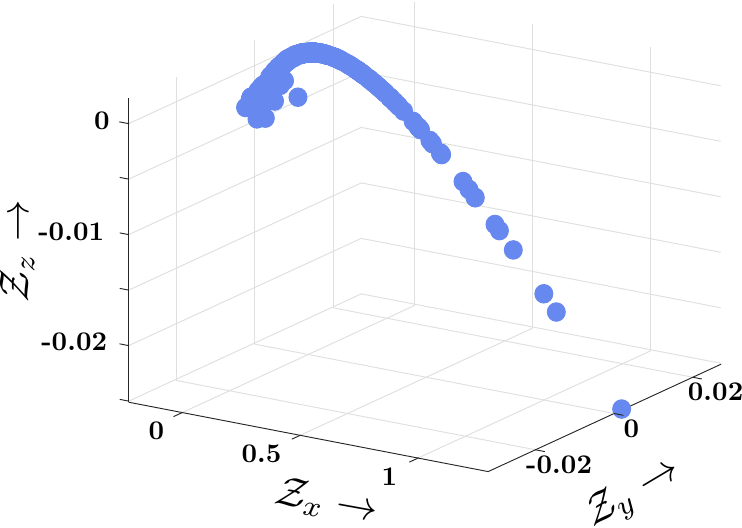} \\
\scriptsize $\bm{\Theta}=[0.9 \ 0.6;0.5\ 0.1]$ &
\hspace*{0.4cm}\scriptsize $\bm{\Theta}=[0.7 \ 0.6 ;0.5 \ 0.1]$ &
% \small $\bm{\Theta}=[0.4 \ 0.7;0.7 \ 0.4]$ &
\hspace*{0.4cm}\scriptsize $\bm{\Theta}=[0.4 \ 0.6;0.6 \ 0.4]$ &
\hspace*{0.4cm}\scriptsize $\bm{\Theta}=[0.6\ 0.6;0.6\ 0.6]$ \\
\includegraphics[width=0.11\textwidth]{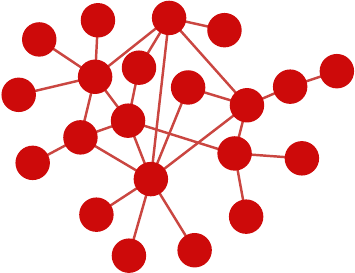} &
\hspace*{0.4cm}\includegraphics[width=0.11\textwidth]{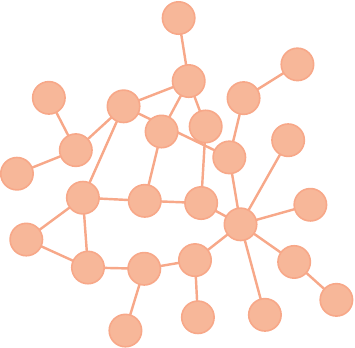} &
 \hspace*{0.4cm}\includegraphics[width=0.12\textwidth]{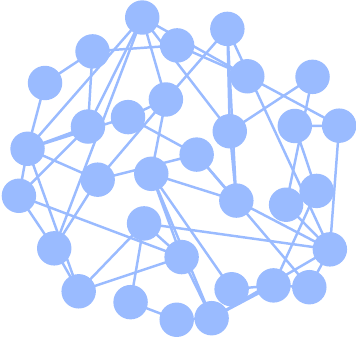} &
\hspace*{0.4cm}\includegraphics[width=0.11\textwidth]{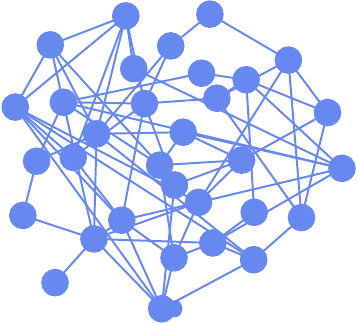}
\end{tabular}}
    \caption{Graph generation by sampling graphs from our probabilistic decoder whose latent representation lies in between the latent representation of two Kronecker 
    graphs, using a linear interpolation. 
    Each column corresponds to a graph, the top row shows the latent representation of all nodes for the graphs in the bottom row, and the middle row shows
    the (inferred) initiator matrix for the Kronecker graph model.h
    %app:experiments-syn
%    The leftmost and rightmost graphs are from the training set and, as the latent representation $\Zcal$ moves from the latent representation of the leftmost graph 
%    to that of the rightmost graph, the sampled graph becomes similar to the rightmost graph and the inferred initiator matrices along the way smoothly interpolate 
%    between both initiator matrix.}
    }
    \label{fig:KronInter}
\end{figure*}

% \subsection{Quality of the generated graphs}
%
Finally, we create a set of $100$ graphs with up to $n = 1000$ number of nodes sampled from the \ba\ graph model with generation parameter $m=1$.
For both \ba\ and Kronecker graphs, we evaluate the quality of the generated graphs using two quantitative evaluation metrics:
%
%
% \begin{table}[t]
% \center
%  \begin{tabular}{|l|c|c|c|}
%  \hline
%    & \multicolumn{2}{c}{\textbf{Generation}}& \\
%  	\hline\parbox[t]{2mm}{\multirow{3}{*}{\rotatebox[origin=t]{90}{\textbf{Train}}}}
% 	
% \multirow{1}{*}{}&   & $\Mcal=\emptyset$  &  $\Mcal\neq \emptyset$ \\
%   	\cline{2-4} 
%  	\cline{2-4}
%   	& $\Mcal=\emptyset$ & 0.57&--\\ 
%  	  \cline{2-4}
%           & $\Mcal\neq\emptyset$ & 0.68 &1.00 \\
%  	  %
%  	 \hline
% %
%  \end{tabular}
%   \caption{Validity achieved by our variational autoencoder trained using triangle free graphs with ($\Mcal\neq\emptyset$) and without ($\Mcal=\emptyset$) masking 
%   during training and during test time. Here, $\text{dim}(z_{i})=7$ and $K=3$.}
% \label{tab:masking}
% \end{table}
%   
%   \begin{figure}[t]
% 	\centeringapp:experiments-syn
% 	 {\includegraphics[width=0.25\textwidth]{sample1.pdf}}\hspace*{0.1 cm}
% 	 {\includegraphics[width=0.25\textwidth]{sample2.pdf}}
% 	{\includegraphics[width=0.25\textwidth]{sample3.pdf}}\hspace*{0.1 cm}
% 	\caption{Graphs sampled using our variational autoencoder trained with a set of triangle free graphs. By using masking, our variational autoencoder is able
% 	to always generate triangle free graphs.}
% 	\label{fig:tree}
% \end{figure} 
\begin{itemize}[leftmargin=0.9cm]
\item[(i)] Rank correlation:
we use this metric to test to which extent the models we trained using \ba\ and Kronecker graphs do generate \emph{plausible} \ba\ and Kronecker 
graphs, respectively.
Intuitively, if the trained models generate plausible graphs, we expect that a graph $\Gcal$ with a very high value of likelihood under the 
true model, $p(\Gcal|\Pcal)$, should also have a high value of likelihood, $\EE_{\Zcal \sim p(\Zcal)}\log p_{\theta}(\Gcal |\Zcal)$, and ELBO 
under our trained model. 
For a set of graphs, we verify this expectation by computing the rank correlation between lists of graphs as follows.
First, for each set of generated graphs $\GG$, we order them in decreasing order of $p(\Gcal|\Pcal)$ and keep the top $10\%$ in a
ranked list\footnote{\scriptsize We discard the remaining graphs since their likelihood is very similar.}, which we denote as $\Tcal_{p}$.
Then, we take the graphs in $\Tcal_{p}$ and create two ranked lists, one in decreasing order of $\EE_{\Zcal \sim p(\Zcal)}\log p_{\theta}(\Gcal|\Zcal)$, which we denote as $\Tcal_{p_{\theta}}$, and another one in decreasing order of ELBO, which we denote as $\Tcal_{\text{ELBO}}$.
Finally, we compute two Spearman'{}s rank correlation coefficients between these lists:
 \begin{equation*}
\rho_{p_{\theta}} := \frac{\text{Cov}(\Tcal_{p},\Tcal_{p_{\theta}})}{\sigma_{\Tcal_{p}}\sigma_{\Tcal_{p_{\theta}}}} \quad \quad
\rho_{\text{ELBO}} := \frac{\text{Cov}(\Tcal_{p},\Tcal_{p_{\text{ELBO}}})}{\sigma_{\Tcal_{p}}\sigma_{\Tcal_{\text{ELBO}}}}
\end{equation*}
where $\rho_{p_{\theta}}, \rho_{\text{ELBO}} \in [-1, 1]$. 

\item[(ii)] Precision:
we use this metric, which we compute as follows, as an alternative to the rank correlation above for \ba\ and Kronecker 
graphs.
For each set of generated graphs $\GG$, we also order them in decreasing order of $p(\Gcal|\Pcal)$ and 
create an ordered list $\Tcal_{p}$, and select $\Tcal_{p}^{\uparrow}$ 
as the top $10$\% and $\Tcal_{p}^{\downarrow}$ as the bottom $10$\% of $\Tcal_p$.
Then, we re-rank this list in decreasing order of $\EE_{\Zcal \sim p(\Zcal)}\log p_{\theta}(\Gcal|\Zcal)$ and ELBO 
to create two new ordered lists, $\Tcal_{p_{\theta}}$ and $\Tcal_{\text{ELBO}}$.
Here, if the trained models generate plausible graphs, we expect that each of the top and bottom halves of $\Tcal_{p_{\theta}}$ 
and $\Tcal_{\text{ELBO}}$ should have a high overlap with $\Tcal_{p}^{\uparrow}$ and $\Tcal_{p}^{\downarrow}$, respectively.
Then, we define top and bottom precision as:
\begin{equation*}
 \gamma^{\uparrow}=\frac{|\Tcal_x^{\uparrow} \cap \Tcal_{p}^\uparrow|}{|\Tcal_{p}^\uparrow|} \quad \quad
 \gamma^{\downarrow}=\frac{|\Tcal_x^{\downarrow} \cap \Tcal_{p} ^\downarrow|}{|\Tcal_{p}^\downarrow|}
\end{equation*}
where $\gamma^{\uparrow}, \gamma^{\downarrow} \in [0, 1]$ and $\Tcal_x^{\uparrow}$ ($\Tcal_x^{\downarrow}$) is the top (bottom) half of 
either $x = \Tcal_{p_{\theta}}$ or $x = \Tcal_{\text{ELBO}}$. 
\end{itemize}

Table~\ref{tab:results_synthetic} summarizes the results, which show that our model is able to learn the generative
process of \ba\ more accurately than Kronecker graphs. This may be due to the higher complexity of the generative process Kronecker graph use. That being
said, it is remarkable that our model is able to achieve correlation and precision values over $0.4$ in both cases.
\begin{table}[!h]
\center
 \scalebox{0.8}{\begin{tabular}{|l||c|c|c|c|c|c|c|}
\hline
 & $\rho_{p_{\theta}}$   & $\rho_{\text{ELBO}}$ & $\gamma^{\uparrow}_{p_{\theta}}$  & $\gamma^{\downarrow}_{p_{\theta}}$ & $\gamma^{\uparrow}_{\text{ELBO}}$  & $\gamma^{\downarrow}_{\text{ELBO}}$  \\ \hline
   \ba	   & $0.69$    &  $0.72$ &      $0.98$ &      $0.98$ & $1.00$ & $1.00$  \\ \hline
  Kronecker & $0.50$ &  $0.21$    & $0.47$  &   $0.47$ & $0.70$ & $0.70$ \\ \hline
 \end{tabular}} 
 \caption{Rank correlation ($\rho$), top precision ($\gamma^{\uparrow}$) and bo\-ttom precision ($\gamma^{\downarrow}$) achieved by our variational autoencoder trained 
with either \ba\ or Kronecker graphs. In both cases, $\text{dim}(z_{i})=7$ and $K=3$. Here, the higher the value of rank correlation and (top and bottom) precision, 
the more accurately the trained models mimic the generative processes for \ba\ and Kronecker graphs.}
\label{tab:results_synthetic}
\end{table}

\subsection{Effect of permuting node labels on decoder parameter estimation }

In this section, we evaluate the permutation invariant property of our decoder over two networks---a \ba\ graph with $1000$ nodes and a Kronecker graph with $1024$ nodes and 
an initiator matrix $\Theta=[0.6,  0.6; 0.6, 0.6]$. For each of these graphs $\Gcal$, we first generate $K$ isomorphic networks $\{\Gcal_{\pi}\}$ with different node labels and then train our decoder (with encoder)
for three different source node ($s$) sampling protocols $\zeta$: (i) degree- distribution based sampling \ie\ $\zeta(s)=d_s/\sum_{i\in\Vcal} d_i$; (ii) maximum degree based sampling \ie\ $\zeta(s)=\Ucal\{s|d_s=\max_{i\in\Vcal} {d_i}\}$;
and uniform distribution \ie\ $\zeta(s)=\Ucal\{s\in\Vcal\}$.
Then, we investigate the variation of $\EE(||\theta_{\Gcal_{\pi}-\Gcal_{\pi'}}||)$, the mean difference between the estimated decoder parameters over the pairs of $\{\Gcal_{\pi}\}$,
with the number of training iterations.

{Figure~\ref{fig:weightchange} summarizes the results which show that degree based methods perform best in case of \ba\ graph and uniform distribution performs best in Kronecker graph. This is because, the degree distribution of \ba\ 
graph is skewed and as a result, a very few source nodes are sampled again and again, thereby giving similar parameter values. On the other hand,
for  homogeneous Kronecker graph, the degree distribution is more or less uniform. Consequently, the degree based methods perform worse in that case.}
\begin{figure}[t]
	\centering
	\hspace*{-0.5cm}
	\subfloat[\ba]{\includegraphics[width=0.3\textwidth]{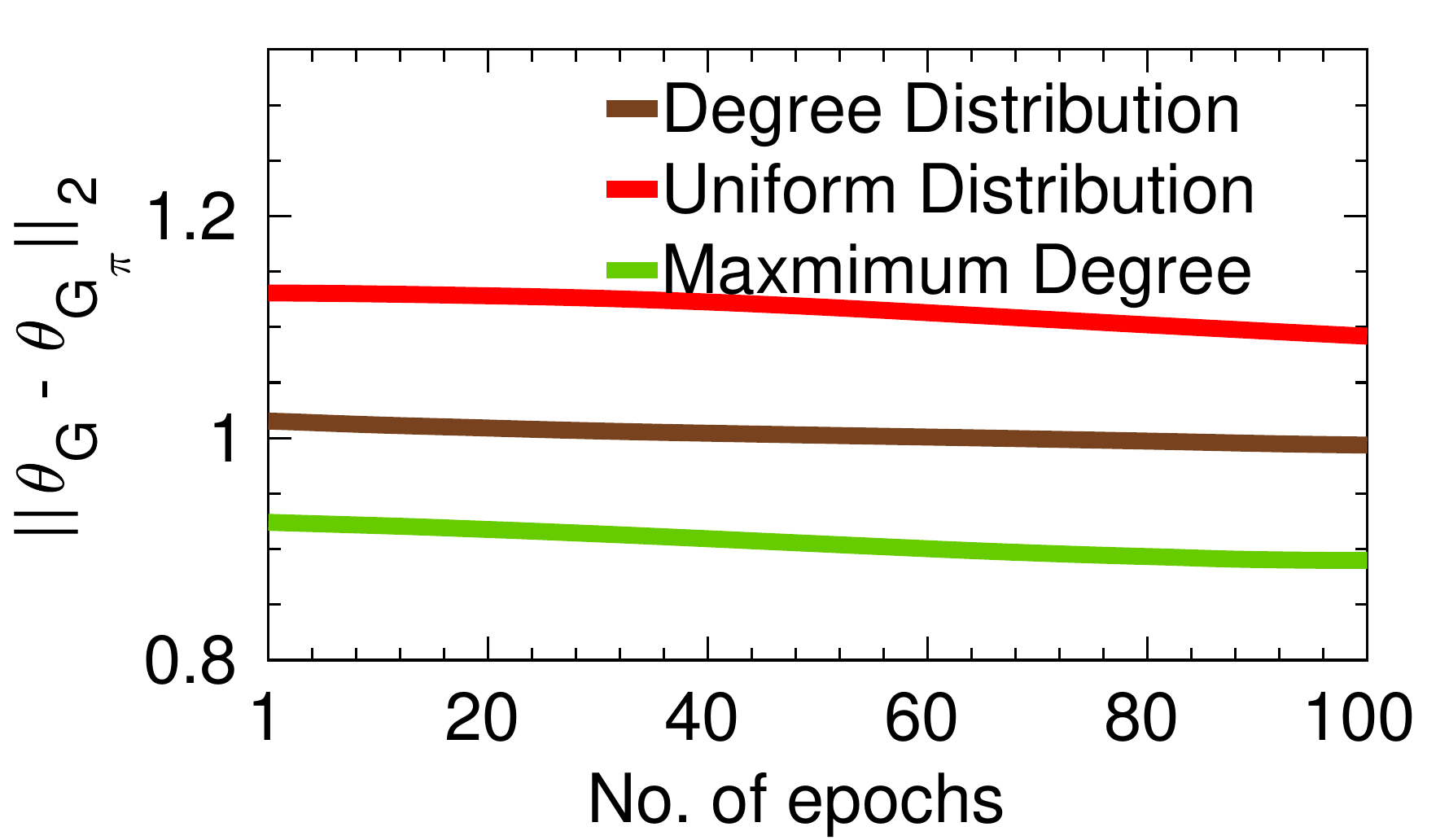}}\hspace*{ 0.9cm}
% 	\subfloat[Legend]{\includegraphics[width=0.12\textwidth]{FIG/weight_diff/Legend.pdf}}\hspace*{-0.5cm}
	\subfloat[Kronecker]{\includegraphics[width=0.3\textwidth]{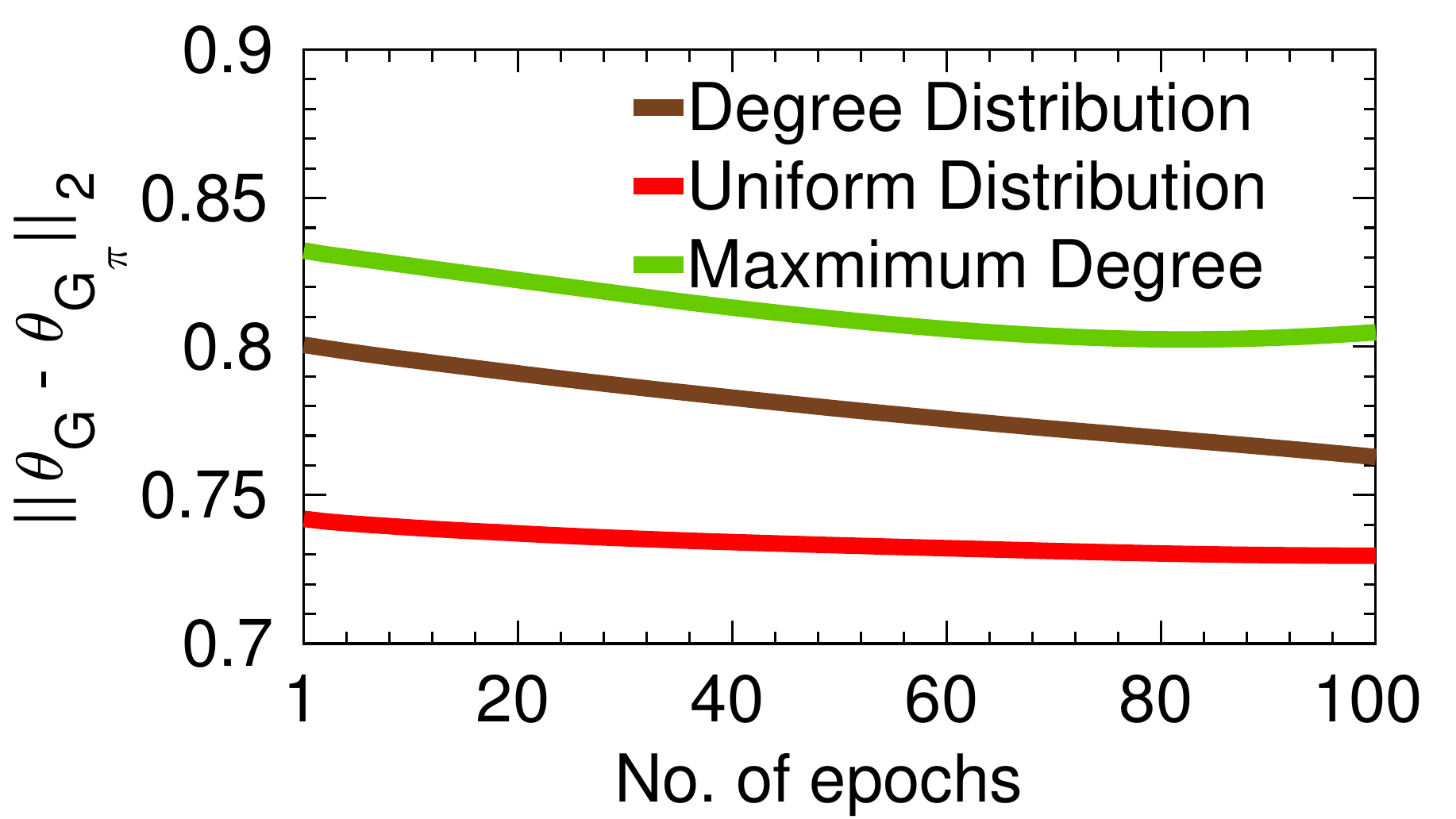}}		
	\caption{Effect of node label permutations on the decoder parameter estimation, for different source node distribution $\zeta$.
		Each plot shows the variation of mean difference in learned decoder weights $\EE\big(||\theta_{\Gcal_1}-\theta_{\Gcal'}||\big)$
		with the number of training iterations, where  $\Gcal$ and $\Gcal'$ are the representation of same graphs different node labels.
		Panel (a) shows this variation for \ba\ graphs with $X$ nodes. Panel (b) shows this variation for a Kronecker graph
		with $Y$ nodes and initiator matrix $\Theta=[0.6\ 0.6;0.6\ 0.6]$. 
%  
% 	
% 	Panel (c) shows the difference in learned decoder weights 
% 		for two set of Kronecker graphs $G$, and the permutation graphs $G_\pi$,  with ${\Theta}=[0.6\ 0.6;0.6\ 0.6]$ for different types of edge order sampling. 
% 		Panel (b) shows the different type of sampling. For  staring node sampling of BFS, we uses 
% 			%
% 			DD = Degree distribution, MD$=$Maximum Degree, UD = Uniform distribution.
% 			%
% 			From each of these distribution we either sample Single starting node or Multiple nodes.
	}
	\label{fig:weightchange}
\end{figure}

\subsection{Effect of $K$ (search-depth in encoder) on model performance} 
In this section, we investigate the behavior of our model with respect to the search depths $K$ used in the decoder.
Figure~\ref{fig:hyper} summarizes the results, which show that, for \ba\ graphs, our model performs consistently well for low 
values of $K$, however, for Kronecker graphs, the performance is better for high values of $K$. 
A plausible explanation for this is that \ba\ networks are generated sequentially using only local topological features (only 
node-degrees), whereas the generation process of Kronecker graphs incorporates global topological features.
\begin{figure}[t]
	\centering
	\subfloat[\ba ]{\includegraphics[width=0.3\textwidth]{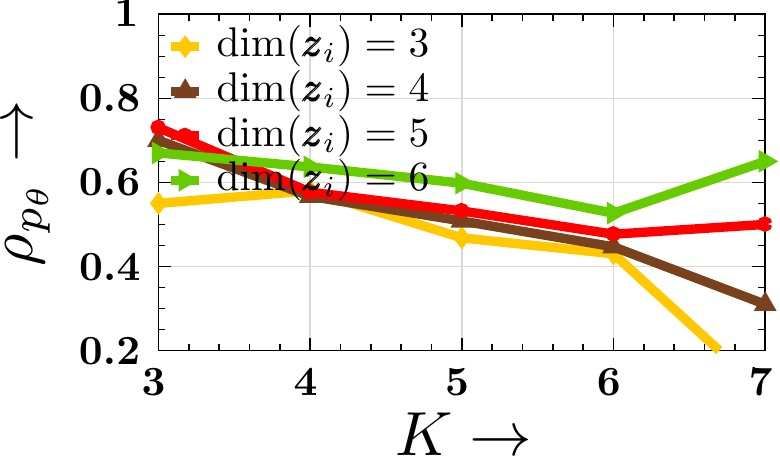}}\hspace*{0.5 cm}
	\subfloat[Kronecker]{\includegraphics[width=0.3\textwidth]{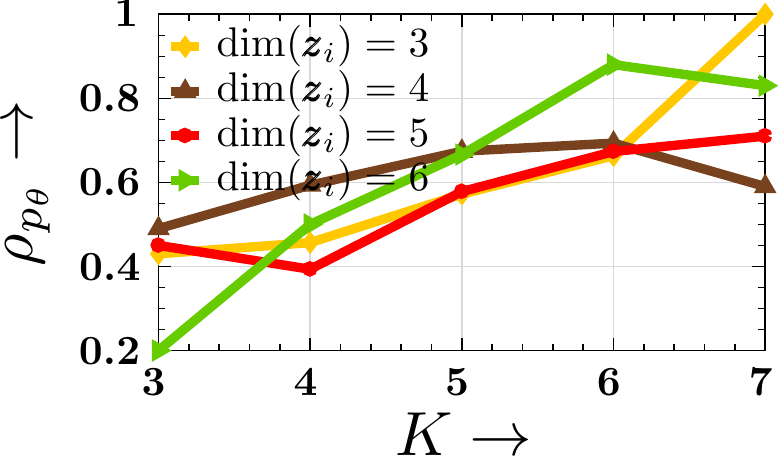}}
	\caption{Rank correlation ($\rho_{p_{\theta}}$) with respect to the search depths $K$ used in the decoder
	for \ba\ graphs, small values of $K$ achieve better performance, whereas for Kronecker graphs, a larger $K$ provides better performance. 
	}
	\label{fig:hyper}
\end{figure}
\begin{figure}[!t]
	\centering
	\subfloat[Training time vs. \# of nodes]{\includegraphics[width=0.26\textwidth]{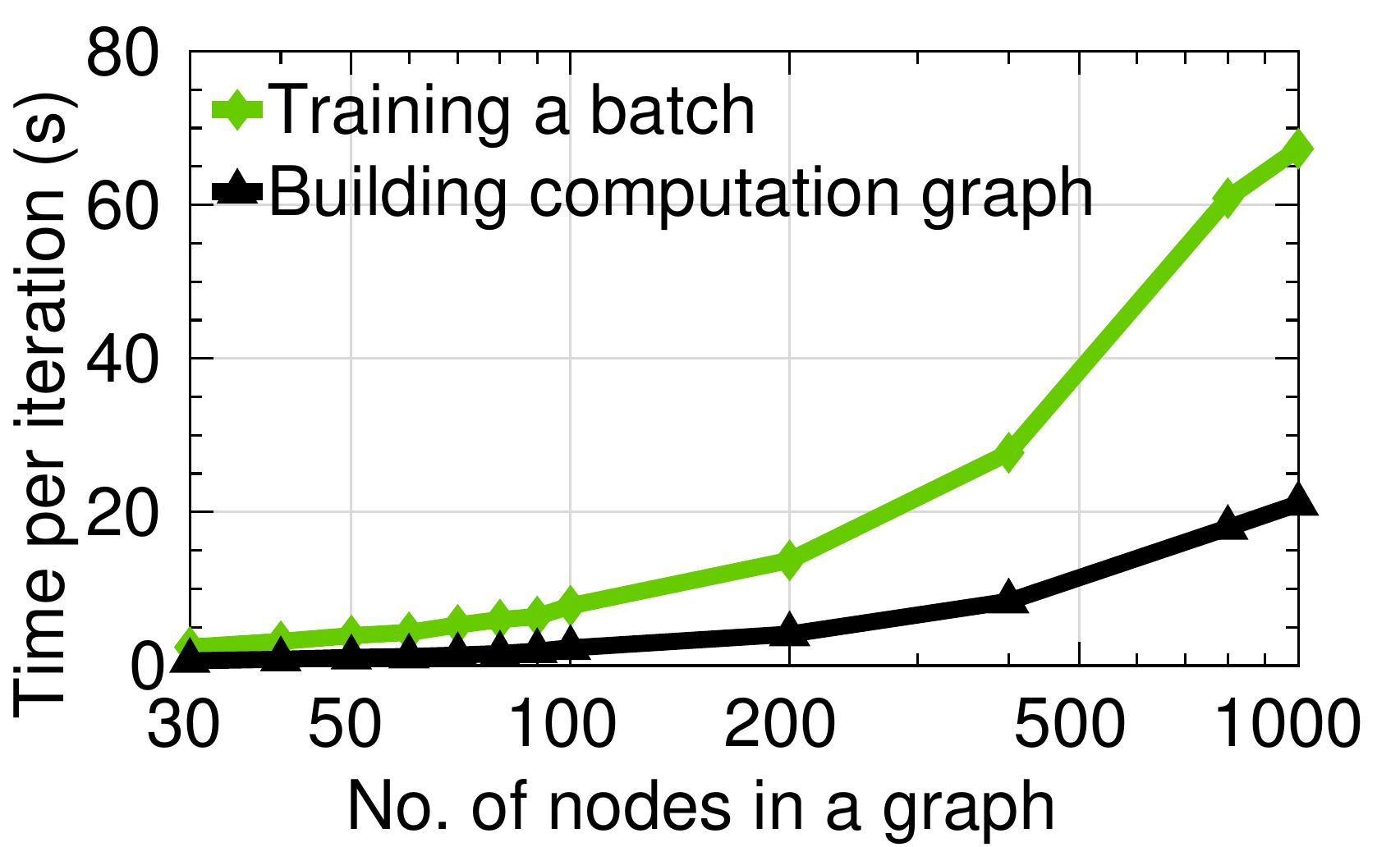}}\hspace*{0.9cm}
	\subfloat[Sampling time vs. \# of nodes]{\includegraphics[width=0.26\textwidth]{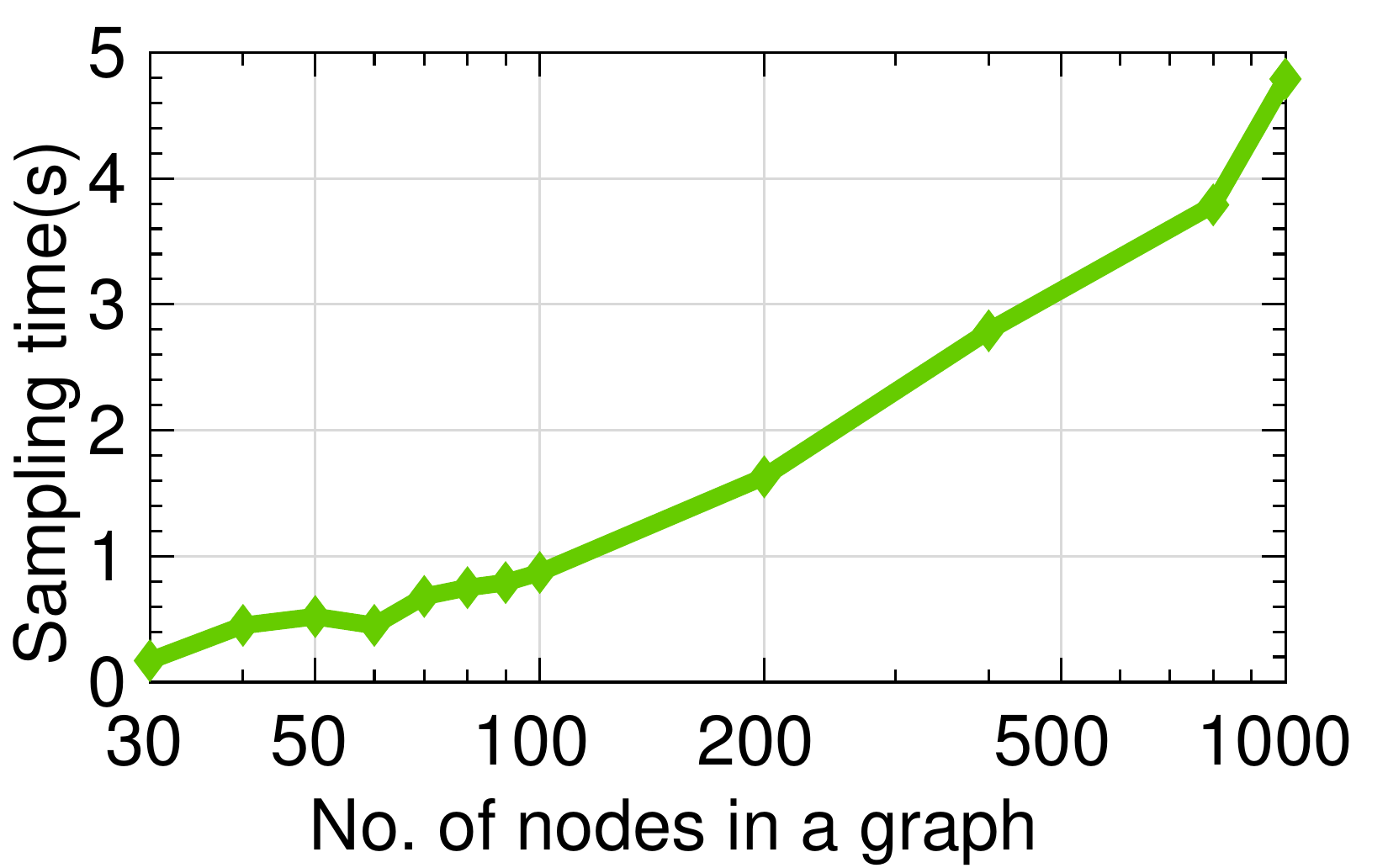}}
	\caption{Scalability of our inference procedure and probabilistic decoder. 
	Panel (a) shows the time per iteration of our variational inference procedure against the size of the graphs in the training set using batches of $10$ graphs with 
	average degree $3$.
	%
%	Due to negative sampling, the times for building the  computational graph as well as training our model are significantly low even for $n=1000$. \
	%
	Panel (b) shows the time our probabilistic decoder takes to sample an entire graph with average degree $3$ against the size of the graph. 
		 %Running time for inference and sampling. The experiments were performed on a machine with OS:Debian 9 CPU:Intel(R) Xeon(R) CPU E5-2667 v4 @ 3.20GHz RAM:0.5TB.
	}
	\vspace{-2mm}
	\label{fig:scalability}
\end{figure}
\subsection{Scalability} \label{app:scalability}
We first compute the running time of our variational inference procedure against the size of the graphs in the training set and then compute the running
time of our probabilistic decoder against the size of the sampled (generated) graphs.
Figure~\ref{fig:scalability} summarizes the results, which show that both in terms of inference and sampling, our model easily scales to $\sim$$1{,}000$ nodes. 
For example, for graphs with $1000$ nodes (average degree $3$), our inference procedure takes $67+20$ seconds to run one iteration of SGD with a 
batch size of $10$ graphs and, for graphs with $50$ nodes, our inference procedure takes less than $10$ seconds per iteration.
Moreover, our probabilistic decoder can sample a graph with $1000$ ($50$) nodes (average degree $3$) in only $5$ ($0.5$) seconds.
%
% \begin{figure}[!t]
% 	\centering
% 	\subfloat[Training time vs. \# of nodes]{\includegraphics[width=0.26\textwidth]{FIG/scalability/scalability_with_label.pdf}}\hspace*{0.9cm}
% 	\subfloat[Sampling time vs. \# of nodes]{\includegraphics[width=0.26\textwidth]{FIG/scalability/sampling_with_label.pdf}}
% 	\caption{Scalability of our inference procedure and probabilistic decoder. 
% 	%
% 	Panel (a) shows the time per iteration of our variational inference procedure against the size of the graphs in the training set using batches of $10$ graphs with 
% 	average degree $3$.
% 	%
% %	Due to negative sampling, the times for building the  computational graph as well as training our model are significantly low even for $n=1000$. \
% 	%
% 	Panel (b) shows the time our probabilistic decoder takes to sample an entire graph with average degree $3$ against the size of the graph. 
% 		 %Running time for inference and sampling. The experiments were performed on a machine with OS:Debian 9 CPU:Intel(R) Xeon(R) CPU E5-2667 v4 @ 3.20GHz RAM:0.5TB.
% 	}
% 	\vspace{-2mm}
% 	\label{fig:scalability}
% \end{figure}

\vspace{-3mm}
\subsection{Property oriented graph generation}
We first train \ourmodel\ over 10,000 \ba~\cite{barabasi1999emergence} graphs with mean number of nodes $\lambda_{n}=50$ and mean number of 
edges $\lambda_l = 95$ and
%\subsection{Loss functions and Metrics} and 
% Next we train \ourmodel with these synthetic graphs. 
then we optimize this trained model using Algorithm~\ref{alg:prop} twice so that it generates graphs with (i) low diameter and (ii) high clustering coefficients, respectively.
% 
% \noindent\emph{--- Property optimization. }
% First we consider two  optimized to generate graphs with (a) \emph{low diameter} and (b)\emph{high clustering coeffcient}. 
To this aim, we set the loss functions $\ell(\Gcal)=\text{Diameter}(\Gcal)$ and $\ell(\Gcal)=1-\text{Clustering-coefficient}(\Gcal)$, respectively. 
%
% 
% 
% For our experiment we used two structural properties (a) diameter (\textit{dia}) and (b) clustering coefficient (\textit{cc}). The distribution of these property values in training dataset is given in Fig ~\ref{fig:train_graphs}.
% 
% \xhdr{Property optimization}
% Here we focus to generate graphs with lower \textit{dia} and higher \textit{cc}. We define the penalty function $\ell(.)$ for diameter optimization for a generated graph $\Gcal$ from $p_\theta^\prime$ as below:
% \begin{align}
% \ell(\Gcal)= dia(\Gcal) \text{ } 
% \end{align}
% and for clustering coefficient optimization as below:
% \begin{align}
% \ell(\Gcal)= 1.0 - cc(\Gcal)
% \end{align}
% Using stochastic gradient descent as described in Sec~\ref{sec:prop} we were able to generate graphs with desired property values.
Figure~\ref{fig:Experiment2} summarizes the results by means of the distributions of diameters and clustering coefficients of the generated graphs against different
values of the parameter $\rho$. The results show that, the smaller the value of $\rho$, the lower (higher) the diameter (clustering coefficient), as one could expect.
%
% We report the change in property distribution with various values of $\rho$ in Figure ~\ref{fig:Experiment2}. There is a smooth transition if property distribution as we decrease the value of $\rho$ for both the cases.  

%To train $p_\theta^\prime$, the property oriented decoder we define the penalty functions given a generated graph $\Gcal$ as below:
%$\ell(\Gcal) = dia(\Gcal)$ where 
%
\begin{figure}[h]
		\hspace*{1cm}\subfloat{\includegraphics[width=0.4\textwidth]{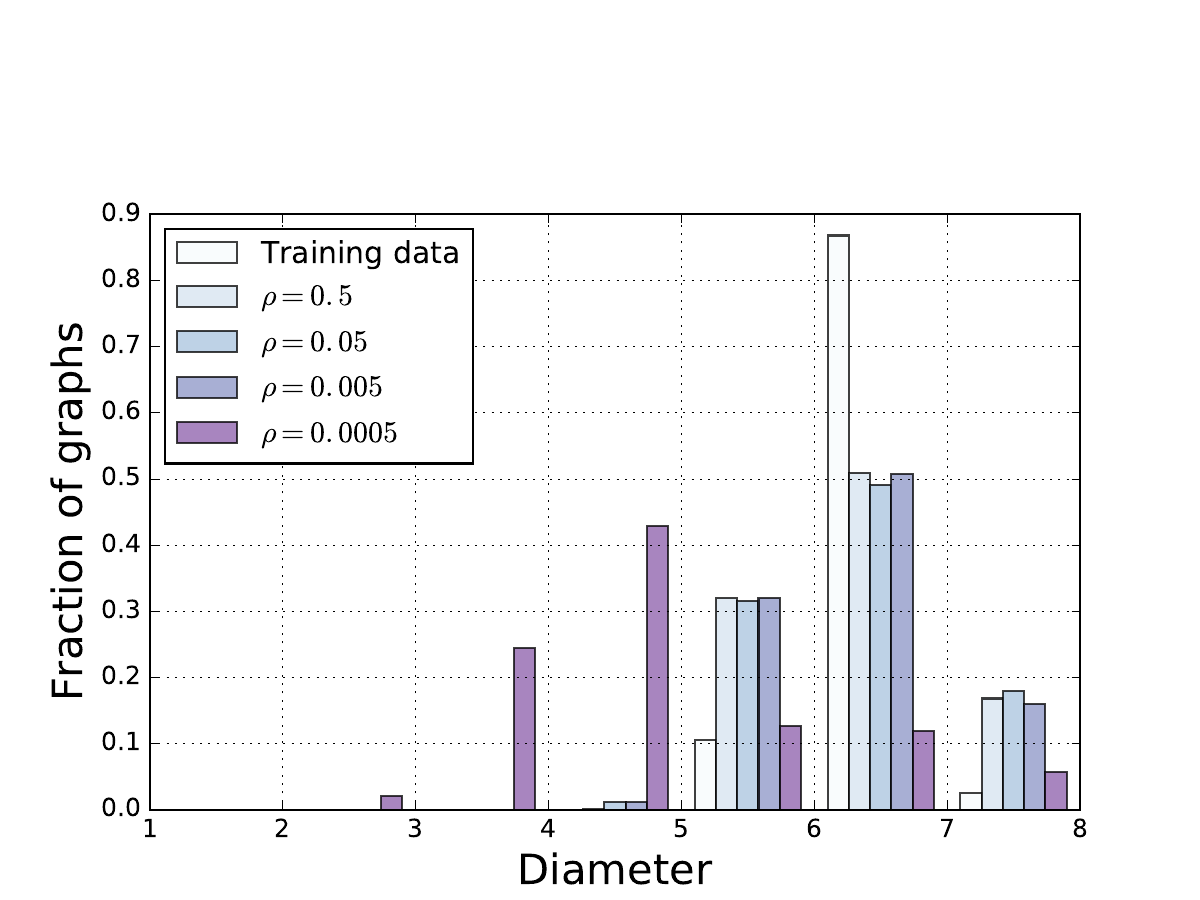}}  \hspace{1cm} 
		\subfloat{\includegraphics[width=0.4\textwidth]{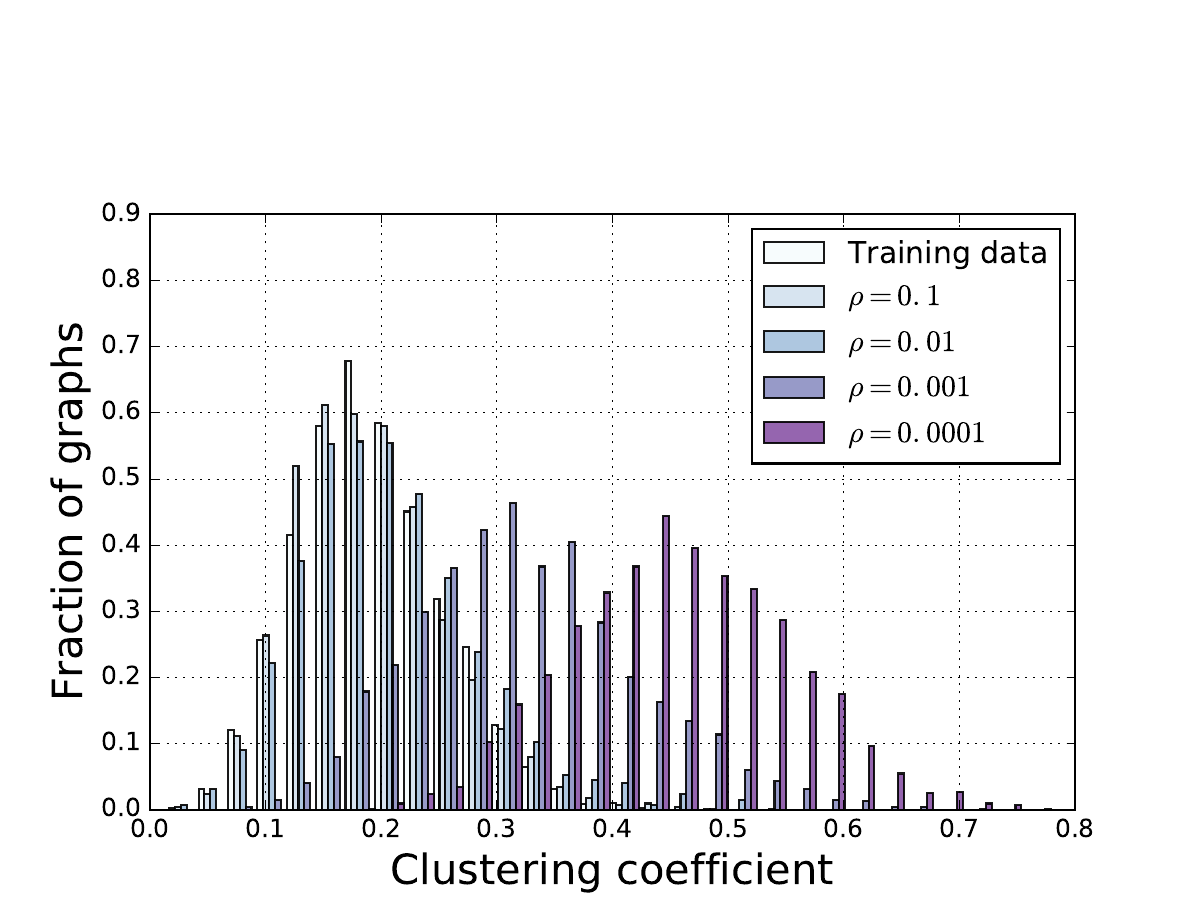}} %\hspace{-1cm} 
\caption{Diameter and clustering coefficient distribution for different values of $\rho$. 
 }
\label{fig:Experiment2}
\end{figure}

Next, we specify an upper bound for the diameter $(\bar{D})$ and then train our property oriented decoder to generate graphs whose diameters
are always smaller than $\bar{D}$. To this aim, we set the loss function 
% To achieve the same we define the penalty function as below:
% 
% \begin{align}
$\ell(\Gcal)=\max \left\{\text{Diameter}(\Gcal) - \bar{D} , 0\right\}$. 
% \end{align}
Here $\Gcal$ is the generated graph and $\bar D$ is the maximum allowed specified diameter value. 
%dia($\Gcal$) denotes the diameter of the graph $\Gcal$.
We measure the quality of the generated graphs in terms of success rate~\cite{you2018graph}, \ie,
\begin{equation}
\text{Success rate}=\frac{ |\{ \Gcal | \text{Diameter}(\Gcal) \le \bar{D}  \}| }{| \{\Gcal\}|}
\end{equation}
Table ~\ref{tab:synthres1} summarizes the results for different $\rho$, which shows that, the lower the value of $\rho$, the higher the success rate.
%
% \newpage
\begin{table}[!!h]
	\small
	\centering
	\begin{tabular}{|c|c|c|}
		\hline
		& $\bar{D}= 4$ & $\bar{D}= 6$  \\ \hline
		$\rho = 1$   & 0.23  &  0.40 \\
		\hline
		$\rho = 10$   & 0.38  &  0.45 \\ \hline
		$\rho = 100$   & 0.62  &  0.75 \\ \hline
		%Diversity Index &  0.50                 &  1.06                                    \\ \hline
		%Log likelihood  &  -410.69 $\pm$ 5.52                     &      -350.54 $\pm$ 6.35             &     -324.32 $\pm$   10.25           \\ \hline
	\end{tabular}
	\caption{Success rate at generating graphs with diameter smaller than $\bar{D}$ for different values of $\rho$.}
	\label{tab:synthres1}
\end{table}

\end{document}